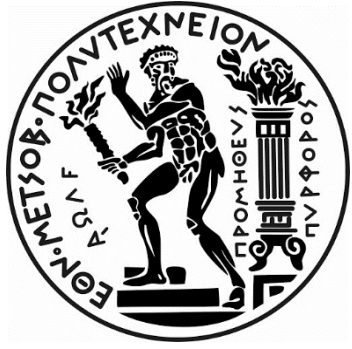

National Technical University of Athens

School of Rural and Surveying Engineering

# Machine learning methods for modelling and analysis of time series signals in geoinformatics

## Doctoral Dissertation

### Maria Kaselimi

Dipl. Rural and Surveying Engineer, NTUA

Supervisor: **NIKOLAOS DOULAMIS**, *Assoc. Professor, NTUA*

# Table of contents

# List of figures

# List of tables

# Chapter 1

# Introduction

## 1.1 Background

A time series (TS) is a sequence taken at successive equally spaced points in time. Time series modeling (TSM) plays a key role in a wide range of real-life problems that have a temporal component [4–6]. Time series analysis (TSA) investigates the mechanism that produces the time series and evaluates its characteristics. The prediction of future time series values is an important research topic in modern time series modelling and analysis and is essential in many emerging research fields such as the weather [7], energy consumption [8], financial indices [9], medical monitoring [10], anomaly detection [11], traffic prediction [12], climate change [13], etc. In these problems, time series modelling often pose significant challenges for the existing techniques both in terms of their structure (i.e. irregular sampling or spatiotemporal structure in climate data) and size [14]. Given that, traditional methods have focused on parametric models informed by domain expertise – such as linear predictive modeling techniques, i.e., autoregressive (AR) models [15], exponential smoothing, or structural time series models [16] – modern machine learning (ML) methods [17] provide a means to learn temporal dynamics in a purely data-driven manner [18]. In recent times, with the increasing data availability and computing power, machine learning has become a vital part of the next generation of time series models [19]. Thus, there is both a great need and an exciting opportunity for the machine learning community to develop theory, models and algorithms specifically for the purpose of processing and analyzing time series data.

Time series data from complex systems capture dynamic behavior and causalities [4]. There are various techniques from time series prediction and monitoring [20]. The nonlinear and non-stationary dynamics of the various applications pose a major challenge for accurate prediction and modeling [4]. Time series emerging from such complex systems exhibit aperiodic patterns even under steady state circumstances. Also, since real-world systems often evolve under transient conditions, the signals obtained therefore tend to exhibit various forms of non-stationarity. This thesis presents the advancements in nonlinear and non-stationary time series models under a machine learning based framework and a comparison of their performances in certain real-world environmental applications and applications in geoinformatics and energy.

Time series have many characteristics that distinguish them from other types of data. Firstly, the time-series data contain high dimensionality and much noise. Dimensionality reduction, wavelet analysis or filtering are Signal processing techniques [21] that can be applied to eliminate some of the noise and reduce the dimensionality [22]. The next characteristic of time-series data, it is not certain that the available information is enough to understand the process. Finally, Time series data is high-dimensional, complex, and has unique characteristics that make time-series challenging to model and analyze. Recent advancements in sensor technologies and the consequent availability of abundant data sources in the form of time series can transform the way that real-world complex systems are monitored and controlled. These advancements lead to complex time series problems.

As already described, handling time series problems is a demanding process, as:

- time series is often noisy with high non-stationarity;

- time series is explicit dependent on the time variable and in addition, are related to multiple external measurable parameters;

- in case of a limited amount of data, it is challenging to provide enough and multi-perspective information to understand the mechanism behind it;

- extrapolation of time series (prediction) is also a challenging procedure;

- blind source separation for time series is a demanding process. Decomposing time series totals to estimate its disaggregates;

Bellow, we summarize the theoretical background for time series modelling and prediction, as well as the fundamentals to identify and extract the factors from which observed time series data was generated in various application domains.

### 1.1.1 Fundamentals on time series modelling and prediction

Given a time series $\{y_1, y_2, ..., y_t\}$, time series prediction is the task of predicting one $\{y_t\}$ or several $\{y_t, y_{t+1}, ... y_{t+n}\}$ future observations in a time series, given a certain number of past observations $\{y_1, ... y_{t-1}\}$. An autoregressive model is a common linear regression approach [23], [24] for time series modeling and prediction. In an autoregression model, the prediction of the variable of interest is by using a linear combination of past values of the variable. The term autoregression indicates that it is a regression of the variable against itself. Thus, an autoregressive model of order $p$ is written as:

$$y_t = \sum_1^{\mathbf{p}} f(y_{t-p}) + e_t \tag{1.1}$$

where $f$ is a linear function and $e_t$ is white noise. We refer to this model as an $AR(\mathbf{p})$ model, an autoregressive model of order $\mathbf{p}$, $\forall \mathbf{p} \in 1, ..., p$.

The limitation of an AR filter is that no external measurable parameters are allowed to be utilized by the model. Thus, Autoregressive Moving Average (ARMA) filters have been also investigated for time series modelling and prediction [25], [26, 27]. The main difference between an AR and ARMA process is that the first (i.e., AR) models a timeseries in terms of its own lags (previous observable measurements), while an ARMA filter includes two parts; an autoregressive (AR) where previous observations affect the output and the moving average (MA) part involves modeling the error term $e_t$ as a linear combination of error terms occurring contemporaneously and at various times in the past. In ARMA models, the notation $ARMA(\mathbf{p},\mathbf{q})$ refers to the model with p autoregressive terms and q moving-average terms. This model contains the $AR(\mathbf{p})$ and $MA(\mathbf{q})$ models,

$$y_t = \sum_1^{\mathbf{p}} f(y_{t-p}) + \sum_1^{\mathbf{q}} g(z_{t-q}) + e_t \tag{1.2}$$

where $\{e_1, ..., e_t\}$ is the white noise error terms.

In cases where external observations trigger the output, the ARMAX model is introduced. The notation $ARMAX(\mathbf{p},\mathbf{q},\mathbf{r})$ refers to the model with $\mathbf{p}$ autoregressive terms, $\mathbf{q}$ moving average terms and $r$ exogenous inputs terms. This model contains the $AR(p)$ and $MA(q)$ models and a linear combination of the last $\mathbf{r}$ terms of a known and external time series $x_1, ..., x_t$. It is given by:

$$y_t = \sum_1^{\mathbf{p}} f(y_{t-p}) + \sum_1^{\mathbf{q}} g(z_{t-q}) + \sum_1^{\mathbf{b}} h(x_{i,t-b}) + e_t \tag{1.3}$$

where $x_t$ is the exogenous variables that affect the model.

In general, time series forecasting models predict future values of a target $y_t$ at time $t$. Each entity represents a logical grouping of temporal information – such as measurements from individual weather stations in climatology, or vital signs from different patients in medicine – and can be observed at the same time. In the simplest case, one-step-ahead forecasting models take the form:

$$y_t = f(y_{t-p:t-1}, \ x_{t-p:t-1}, \ s) \tag{1.4}$$

where $y_t$ is the model prediction, $y_{t-p:t-1} = \{y_{t-p}, ..., y_{t-1}\}$, $\quad x_{t-p:t-1} = \{x_{t-p}, ..., x_{t-1}\}$ are observations of the target and exogenous inputs respectively over a look-back window $p$, and $s$ is static metadata associated with the entity (e.g., sensor location), and $f(\cdot)$ is the prediction function learnt by the model. While we focus on univariate forecasting in this survey (i.e., 1D targets), we note that the same components can be extended to multivariate models without loss of generality. The AR models described above and their variants, model the prediction function $f$ as a linear function. However, more complex, non-stationary, and noisy real time series cannot be expressed by analytical equations with parameters to solve because the dynamics are either too intricate or undiscoverable. Besides, the traditional models, provided with a small number of nonlinear mechanisms, are not able to capture the complicated data. In order to better model complex real time series depended on various external parameters, the highly recommended approach is to exploit robust features that grasp the relevant information from time series. Nevertheless, exploiting domain-specific features for each task is both time-consuming and costly, and requires expertise of the time series. Recently, and mainly due to the advances in Artificial Intelligence (AI) research, many efforts are utilized to develop time series models using non-linear regression architectures based on machine learning algorithms. In this context, a widely used model is the feedforward neural network, consisting of interconnected artificial neurons capable of modelling non-linear input-output relationships [28]. Feedforward neural network examples have been utilized in time series signals modeling for geodetic applications, examples are the works of [29], [30], [31]. Actually, feedforward neural networks are capable of approximating non-linear ARMA relationships and therefore, improving performance in timeseries modeling. Non-linear ARMA filters with recursive capabilities have been also proposed in [32], [33]. Other works in this field apply Radial Basic Function (RBF) models [34], with advance non-linear interpolation capabilities, or Support Vector Machines (SVMs) [35]. The latter are supervised learning paradigms for data classification and regression.

The main limitations of the aforementioned non-linear regression models is that they present convergence instabilities especially when a large number of neurons are employed in the network. In addition, there is no recurrent feedback mechanisms among the artificial neurons. Therefore, such filters fail to approximate temporal dependencies and high abrupt changes in timeseries values with a high precision accuracy.

For this reason, recently deep machine learning have been proposed as an alternative paradigm for regression and classification [36]. Deep learning incorporates multiple hidden neurons and applies advanced learning algorithms, such as input compression and dimensionality data reduction, to handle the computational issues arising when a large number of neurons are considered. Recurrent Neural Networks (RNNs) are a class of networks allowing connections (feedback) between the nodes (neurons) in order to model the temporal behaviours of a timeseries signal [37, 38]. Thus, RNNs are capable of handling the time-dependencies. However, RNNs fail to approximate more complex temporal dependencies, presenting also computational issues in computing the gradient during the learning process especially when a large number of neurons are employed (the so called vanishing gradient problem [39]).

To address these limitations and simultaneously to retain the advantages of deep learning in approximating temporal dependencies, Long-Short Term Memory (LSTM) architectures have been recently proposed [40, 41]. LSTM networks memorize temporal correlations of the signals, providing, therefore better modeling capabilities [42–44]. LSTMs have more trainable parameters (e.g., weights) compared to the traditional RNNs models. Consequently, they give us, on the one hand, more controlability and thus better results, but with the cost, on the other hand, of more complexity and the need of large amount of training (annotated) data.

Although the aforementioned architectures are good modelling approximators of uni-dimensional signals, they mainly focus on the temporal modelling dimension. However, various timeseries signals have also a longitudinal dimension and therefore, different weights should be assigned to the model for different regions. This so-called *spatial variability* can not be adequately modelled through the traditional LSTM deep learning architectures. Recently, more and more approaches leveraging convolutional architectures have been proposed for sequence modeling tasks. Long- and short-term time series network (LSTNet) uses a combination of a convolutional neural network (CNN) and an RNN to perform multivariate time series prediction. The intention of this approach is to leverage the strengths of both the convolutional layer (to extract short-term local dependency patterns among the multidimensional variables) and the recurrent layer (to discover long-term patterns for time series trends). In the evaluation of real-world data with complex mixtures of repetitive patterns, LSTNet achieved significant performance improvements over that of several state-of-the-art baseline methods, as, e.g., autoregression models and gated recurrent unit (GRU) networks.

The problem described in (1.4) is the direct problem of computing the future $y_t$ values given the previous observations of the target $y$ and exogenous inputs $x_i$; the inverse problem is to compute the inputs $x_i$ given the $y_t$ values.

### 1.1.2 The blind source separation problem formulation

A critical point in the analysis of time series data, is the development of data-driven methods that allow the different sources of the signals to be discerned and characterized in various domains. Blind source separation (BSS) is a class of inverse problems, in which one wishes to separate a set of linearly mixed signals, having only the mixtures in hand and information neither about the sources, nor about the mixing coefficient. BSS is a popular multivariate approach for decomposing multivariate data into uncorrelated components which are useful for dimension reduction, and intended for an easier interpretation or easier modeling of the data. BSS is about decomposing the time series signal $\{y_1, y_2, ..., y_t\}$ into its $x_m$ additive sub-components, where $x_m$ is a timeseries $\{x_{m,1}, x_{m,2}, ..., x_{m,t}\}$ and $m = 1, ..., M$. Thus,

$$\begin{bmatrix} y_1 \\ y_2 \\ \vdots \\ y_t \end{bmatrix} = W \begin{bmatrix} x_{1,1} & x_{2,1} & ... & x_{m,1} & ... & x_{M,1} \\ x_{1,2} & x_{2,2} & ... & x_{m,2} & ... & x_{M,1} \\ \vdots & \vdots & & \vdots & & \vdots \\ x_{1,t} & x_{2,t} & ... & x_{m,t} & ... & x_{M,t} \end{bmatrix}, \forall m \in 1, ..., M \tag{1.5}$$

Many inverse problems are considered to be ill-posed, because the application of $W^{-1}$ to the noisy data $y$ produces a result that is deemed inadequate or because it is not possible to define an operator $W^{-1}$. Being ill-posed does not mean that a problem cannot be solved. However, it means that additional information needs to be incorporated into the inversion.

## 1.2 Research Objectives

Deep learning neural networks learn complex mappings between inputs and outputs and support multiple inputs and outputs. Deep neural networks have successfully been applied to address time series problems. They have proved to be an effective solution given their capacity to automatically learn the temporal dependencies present in time series [45]. However, selecting the most convenient type of deep neural network and its parametrization is a complex task that requires considerable expertise. Therefore, there is a need for deeper studies on the suitability of all existing architectures for different tasks for time series modeling. Hence, practical aspects, such as the setting of values for hyper-parameters and the choice of the most suitable frameworks, for the successful application of deep learning to time series are also provided and discussed.

Given its complexity, time series modeling is an area of paramount importance in various research fields. Time series models needs to take into account several issues such as trends and seasonal variations of the series, the correlation between observed values that are close to time as well as their dependence from various external parameters. In this dissertation, proposed methods for dealing with the above mentioned aspects are discussed.

Another problem that is discussed here is the problem called blind source separation (BSS) [46]. In a large number of cases, the signal received by a sensor (antenna, microphone, etc.) is the sum (mixture) of elementary contributions that we can call sources. For instance, the signal received by an antenna is a superimposition of signals emitted by all the sources which are in its receptive field. Generally, sources as well as mixtures are unknown. In this case, without any knowledge on the sources (except independence assumption), this problem is called blind separation of sources. The separation of independent sources from an array of sensors is a classical but difficult problem in signal processing and it is a problem discussed also here.

In this dissertation is provided a comparative analysis that evaluates the performance of several deep learning (DL) architectures on a large number of time series datasets of different nature and for different applications. Two main fruitful research fields are discussed here:

- Ionospheric Total Electron Content (TEC) modeling which is an important issue in many real-time GNSS applications. Reliable and fast knowledge about ionospheric variations becomes increasingly important. GNSS users of single-frequency receivers and satellite navigation systems need accurate corrections to remove signal degradation caused by the ionosphere. Ionospheric modeling using signal-processing techniques is the subject of discussion in the present contribution.

- Energy disaggregation which is an important issue for energy efficiency and energy consumption awareness. Reliable and fast knowledge about residential energy consumption at appliance level becomes increasingly important nowadays and it is an important mitigation measure to prevent energy wastage. Energy disaggregation or Non-intrusive load monitoring (NILM) is a single channel blind source separation problem where the task is to estimate the consumption of each electrical appliance given the total energy consumption.

### 1.2.1 The challenge of TEC modeling using GNSS data

The variability of the ionospheric parameters limits the efficiency of communications, radars and navigation systems. Total Electron Content (TEC), defined as the integral of the electron density over a signal path, is often used to describe ionosphere variability [47], [48]. TEC values are necessary in order to correct ionospheric refraction [49] and are crucial for satellite based navigation systems, in order to guarantee the high performance of satellite systems in positioning.

The transmitted signals from Global Navigation Satellite Systems (GNSS) are directly affected by the ionospheric variations, causing delays [50]. These delays depend on the signal frequency and the electron density along the transmission path. Hence, ionospheric variability introduces an additional error source in GNSS positioning [51]. The use of multiple navigation signals of distinct center frequency transmitted from the same GNSS satellite allows direct estimation of these ionospheric delays. Exploiting the fact that different signal frequencies are affected differently by the ionosphere, an appropriate processing strategy of multiple-frequency GNSS signals, eliminates the ionospheric error [51]. Contrary to multi-frequency GNSS receivers, real-time (RT) single-frequency (SF) positioning with a low-cost receiver has received increasing attention in recent years due to its large amount of possible applications. However, in this case, one major challenge is the effective mitigation of these ionospheric delays [52]. RT-SF-SPP (Standard Point Positioning)/PPP (Precise Point Positioning) techniques use ionospheric vertical TEC (VTEC) products released by the International GNSS Service (IGS) real-time service [53], to eliminate the ionospheric error and apply corrections to the model as external parameters. The Center for Orbit Determination in Europe (CODE) provides Global Ionospheric Maps (GIM) in a grid of $(2.5° × 5.0°)$, in 2-h temporal resolution, leading to a spatio-temporal sparse model. This means that the corrections applied improve the accuracy of the provided solution for positioning. However, these ionospheric VTEC products fail to remove the total amount of noise caused by the ionosphere, due to their global coverage and their inability to be accurately adapted at regional (and more so at local) levels.

The objective of this dissertation is to create regional TEC models using deep learning methods and surpass problems arising using the traditional methods for TEC modeling. Thus, stations or roving users near the reference area of these regional deep learning models, could benefit from these model instead of a global one.

In particular, the following research objectives that stem from the aim of this dissertation have been defined:

- to improve the existing regional TEC models accuracy, and to investigate the model's response in cases of high ionosphere activity and during irregular conditions;

- to investigate the importance of the various external parameters (e.g. related solar and geomagnetic activity) for improving the accuracy in TEC modeling;

- to propose meaningful adaptations for CNNs models to be applied for time series TEC analysis and compare their performance with the traditional ML methods for time series modeling;

- to introduce reliable regional observation data to characterize regional anomalies that affect the ionosphere at local level (*spatial*), and additionally to estimate TEC changes between epochs (*temporal*), targeting to spatio-temporal VTEC prediction models

- to present a comprehensive analysis for different years under different solar and geomagnetic activity and to test the models for different days and at various latitudes.

## 1.2.2 The energy disaggregation challenge

The residential total power consumption is measured using smart meter devices, thus the consumers be aware of their total (aggregate) power consumption. However, as far as energy efficiency is concerned, energy consumption awareness at appliance level is essential. One way to measure consumption per appliance is through the usage of smart plugs, a solution which is economically unaffordable. For this reason, usually Non-Intrusive Load Monitoring (NILM) or energy (power) disaggregation methods are applied. Separating the household aggregated energy consumption signal into its additive sub-components (the power signal from individual appliances), is called energy (power) disaggregation. Energy disaggregation can be considered as an efficient and cost effective framework to reduce energy consumption [54]. NILM is a research field receiving increasing attention among the signal processing technical community [55], due to its great significance as a means towards attainment of global energy conservation goals, as well as its interest as a challenging scientific and engineering problem. NILM resembles the signal source separation problem and poses several challenges given that it is an ill-posed problem. The following specific issues were addressed in detail:

- to exploit in what extend the entrance of the various external parameters (e.g. voltage, current) help in improving the prediction accuracy in NILM

- to implement optimization techniques for hyper-parameters tuning to improve models performance and increase the reliability of machine learning algorithms;

- explore NILM techniques at a larger scale and across different households. Examine the proposed models applicability in different households and transferability across different datasets;

- to examine the importance of the dynamic context of an appliance, as well as changing contextual conditions in NILM performance;

- to investigate the ability of generative adversarial networks to create robust appliance power patterns for energy disaggregation in the presence of noise;

## 1.3   Contribution and Originality

### 1.3.1   Contribution in Signal Processing and Time series analysis

The overall contribution of this dissertation lies in the following aims that were defined:

- to propose automated approaches for time series modeling - including automatic parametric model selection;

- to propose models that enable complex and larger data to be processed and analysed;

- to propose a sequence-to-sequence deep learning framework for multi-step-ahead time series modeling, addressing the dynamic, spatial-temporal and nonlinear characteristics of multivariate time series data ;

- to propose meaningful adaptations for deep learning algorithms initially proposed for two-dimensional data to one-dimensional time series data;

- to propose robust structures for time series modeling in cases where we have a limited amount of data and/or corrupted data with noise;

- to propose deep learning methods for solving inverse ill-posed time series problems.

### 1.3.2   Contributing to GNSS positioning and TEC modeling

In addition, the following research objectives related to GNSS positioning and TEC modeling have been defined:

- to propose an LSTM-based model for accurate timeseries TEC estimates from GNSS data, surpassing problems related to ionosphere irregular behavior at regional level;

- to propose robust deep learning models that exploit various external parameters (e.g. related solar and geomagnetic activity) to improve the prediction and the accuracy in TEC modeling;

- to propose temporal sequence-to-sequence CNNs models to be applied for time series TEC analysis and compare their performance with the traditional ML methods for time series modeling;

- to scrutinize the effectiveness and efficiency of a convolutional enriched recurrent neural network for spatio-temporal VTEC prediction

- to present a comprehensive analysis for different years under different solar and geomagnetic activity. The models are tested for different days and at various latitudes to see model's response in cases of high ionosphere activity.

### 1.3.3   Contributing to the domain of Environmental Awareness and Energy disaggregation

Energy disaggregation poses several challenges not only as it is an ill-posed problem, but also, due to unsteady appliance signatures, abnormal behaviour that is usually detected in appliances operation and the existence of noise in the aggregated signal. To this day, various approaches have been proposed to solve the NILM problem, as presented in Chapter 6. Some of the most successful ones exploit deep learning neural network structures for modelling an energy disaggregation problem (e.g. recurrent neural network structures [56]). Nevertheless, there are barriers and limitations that until recently, have not been properly addressed. In this connection, a modular and a self-trained energy disaggregation model is proposed, with the capability of updating its behavior whenever a significant change of contextual conditions is detected. Since it is difficult to define the general polyparametric notion of context, and consequently, decide when a model update is required, the proposed techniques have not been applied successfully across different households and datasets [57]. Thus, it is difficult to create algorithms with a good *generalization ability*. In addition, noisy aggregate energy consumption measurements significantly deteriorate the performance of NILM methods. Usually, the detected appliances have unsteady signatures or present abnormal behavior and additionally, the existence of noisy as well as inadequate datasets deteriorates overall models' performance [58]. In that respect, the originality and novelty of this dissertation is summarized in the following objectives which were addressed:

- to exploit deep learning neural network structures for energy disaggregation that include various external parameters (e.g. voltage, current) to improve the prediction and the accuracy in NILM

- to propose a Bayesian-optimized LSTM-based structure with hyper-parameters tuning to improve the performance of the NILM alogrithm;

- to propose a modular and a self-trained energy disaggregation model with the capability of updating its behavior whenever a significant change of contextual conditions is detected;

- to propose a context aware NILM technique that could be applied at a large scale and transferred across different households;

- to propose an inverse GAN-based architecture for robust to noise energy disaggregation;

## 1.4 Structure of the Dissertation

Chapter 2 introduces the theoretical background of the TEC modeling. A review in the existing literature is provided and the the problem formulation follows. In addition, a discussion on the role of GNSS observables in TEC modeling is provided.

Chapter 3 proposes a deep learning-based approach for ionospheric modeling. The method exploits the advantages of Long Short-Term Memory (LSTM) Recurrent Neural Networks (RNN) for timeseries modeling and predicts the total electron content per satellite from a specific station. The evaluation of the proposed recurrent method for the prediction of vertical total electron content (VTEC) values is compared against the traditional Autoregressive (AR) and the Autoregressive Moving Average (ARMA) methods.

Chapter 4 proposes a model suitable for predicting the ionosphere delay at different locations of receiver stations using a temporal 1D convolutional neural network (CNN) model. In this chapter, a variant of CNN, called temporal convolutional networks (TCNs), is discussed for modeling non-linear and complex time series data, competing directly with RNNs in terms of accuracy.

In Chapter 5 a deep learning model is proposed, to efficiently predict TEC values and to replace the GIM derived data which inherently have a global character, with equal or better in accuracy regional ones. The proposed model is suitable for predicting the ionosphere delay at different locations of receiver stations. The model is tested during different periods of time, under different solar and geomagnetic conditions and for stations in various latitudes, providing robust estimations of the ionospheric activity at regional level. The model is a hybrid structure comprising of a convolutional layers and recurrent layers.

Chapter 6 introduces the problem of energy disaggregation and provides a brief literature review.

Chapter 7 proposes a CNN-based (Convolutional Neural Network) architecture with inputs and outputs formed as data sequences taking into consideration an appliance's previous states for better estimation of its current state. Furthermore, the proposed model endows CNN models with a recurrent property in order to better capture energy signal interdependencies. Using a multi-channel CNN architecture fed with additional variables related to power consumption (current, reactive and apparent power), additionally to active power, overall performance, robustness to noise and convergence times are improved.

Chapter 8 introduces a non-causal adaptive context-aware bidirectional deep learning model for energy disaggregation. The proposed model, CoBiLSTM, harnesses the representational power of deep recurrent Long Short-Term Memory (LSTM) neural net-works, while fitting two basic properties of NILM problem which state of the art methods do not appropriately account for: non-causality and adaptivity to contextual factors (e.g. seasonality). A Bayesian-optimized framework is introduced to select the best configuration of the proposed regression model, driven by a self-training adaptive mechanism. Furthermore, the proposed model is structured in a modular way to address multi-dimensionality issues that arise when the number of appliances increases. Experimental results indicate the proposed method's superiority compared to the current state of the art.

Chapter 9 proposes a EnerGAN++, a model based on Generative Adversarial Networks (GAN) for robust energy disaggregation. EnerGAN++ unifies the autoencoder (AE) and GAN architectures into a single framework, in which the autoencoder achieves a non-linear power signal source separation. EnerGAN++ is trained adversarially using a novel discriminator, to enhance robustness to noise. The proposed architecture of the discriminator leverages the ability of Convolutional Neural Networks (CNN) in rapid processing and optimal feature extraction, among with the need to infer the data temporal character and time dependence. Experimental results indicate the proposed method's superiority compared to the current state of the art.

Finally, in Chapter 10, a summary and the contribution of this dissertation are presented, the conclusions and accomplishments derived by the research conducted during the course of this dissertation are discussed and ideas for future research are outlined.

# Chapter 2

# Ionospheric TEC Modeling Based on GNSS Observations

The ionosphere is typically defined as that part of the earth's upper atmosphere with sufficient concentration of free electrons affecting the propagation of electromagnetic waves. The electrons density depends on the time of the day and the Sun's activity, the atmospheric density profile, the geographic location, as well as the magnitude and orientation of the Earth's magnetic field [59]. Total Electron Content (TEC), defined as the integral of the electron density over a signal path, is often used to describe ionosphere variability [47], [48]. TEC values are often reported in multiples of the so called TEC unit, defined as $TECU = 10^{16}el/m^2 \simeq 1.66 \times 10^{-8}molm^{-2}$, which corresponds to 0.162 m on the GPS $L1$ frequency. Since radio wave signals pass through the electrons of the ionosphere, the signal velocity and the ray path change [60].

Consequently, the transmitted signals from Global Navigation Satellite Systems (GNSS) are directly affected by the ionospheric variations, causing delays [50]. These delays depend on the signal frequency and the electron density along the transmission path. Hence, ionospheric variability introduces an additional error source in GNSS positioning [51]. The use of multiple navigation signals of distinct center frequency transmitted from the same GNSS satellite allows direct estimation of these ionospheric delays. Exploiting the fact that different signal frequencies are affected differently by the ionosphere, an appropriate processing strategy of multiple-frequency GNSS signals, eliminates the ionospheric error [51]. Contrary to multi-frequency GNSS receivers, real-time (RT) single-frequency (SF) positioning with a low-cost receiver has received increasing attention in recent years due to its large amount of possible applications. However, in this case, one major challenge is the effective mitigation of these ionospheric delays [52]. RT-SF-SPP (Standard Point Positioning)/PPP (Precise Point Positioning) techniques use ionospheric vertical TEC (VTEC) products released by the International GNSS Service (IGS) real-time service [53], to eliminate the ionospheric error and apply corrections to the model as external parameters. However, these ionospheric VTEC products have global coverage.

## 2.1 Review and analysis of the literature

Many researchers investigate the ionospheric variation, applying regional models. Traditional methods for TEC modeling include parameter estimations, e.g. autoregressive (AR), autoregressive moving average (ARMA), autoregressive integrated moving average (ARIMA) [52]. An Autoregressive model is a common linear regression approach [23], [24]. The limitation of an AR filter is that no external measurable parameters are allowed to be utilized by the model. Thus, Autoregressive Moving Average (ARMA) filters have been also investigated for time series modelling [25], [26]. The main difference between an AR and ARMA process is that the first (i.e., AR) models a timeseries in terms of its own lags (previous observable measurements), while an ARMA filter includes two parts; an autoregressive (AR) where previous observations affect the output and the moving average (MA) part where external observations trigger the output. Wang et al. [52] compare the performance of various models including ARMA model and ARIMA models to predict total electron content (TEC) in various latitudes. Ratnam et al. [26] propose a new multivariate ionospheric TEC forecasting model based on linear time series model in combination with ARMA model. Mandrikova et al. [23] suggest a technique of ionospheric parameter modelling and analysis based on combining the wavelet transform with ARIMA models. Widely used linear approaches for time series forecasting are autoregressive models and its different variations. Even though these models are often proposed in the literature for modeling ionospheric TEC values, these models fail to capture nonlinear patterns and abrupt changes in electron density values due to the linear assumptions made. To this end, various nonlinear approaches for time series TEC modeling have been proposed.

Machine learning techniques can successfully deal with short-term forecasting [61]. Due to the recent widespread advancements in machine learning, many significant research efforts are utilized to develop the suitable TEC prediction methods including shallow methods, such as Support Vector Machine (SVM) [35], the nonlinear radial basic function (RBF) neural network [49, 34].

Recently, and mainly due to the advances in Artificial Intelligence (AI) research, many efforts are utilized to develop time series models using non-linear regression architectures based on machine learning algorithms. In this context, a widely used model is the feedforward neural network, consisting of interconnected artificial neurons capable of modelling non-linear input-output relationships [28]. Feedforward neural network examples have been utilized in time series signals modeling for geodetic applications, examples are the works of [29–31]. Actually, feedforward neural networks are capable of approximating non-linear ARMA relationships and therefore, improving performance in TEC modeling. Tulunay et al. [62] propose a data-driven neural network model. Mallika et al. [63] introduce a novel ionospheric forecasting algorithm based on the fusion of principal component analysis and artificial neural networks (PCA-NN) methods to forecast the ionospheric TEC values. The work of [64] present a new ionospheric prediction model, named the singular spectrum analysis-artificial neural network (SSA-ANN) as a preprocessing tool for the ionosphere total electron content (TEC) prediction based on the ANN approach. Wang et al. [65] propose an artificial neural network (ANN)-based Australian TEC model is developed to predict the TEC values.

The main limitations of the aforementioned non-linear regression models is that they present convergence instabilities especially when a large number of neurons are employed in the network.

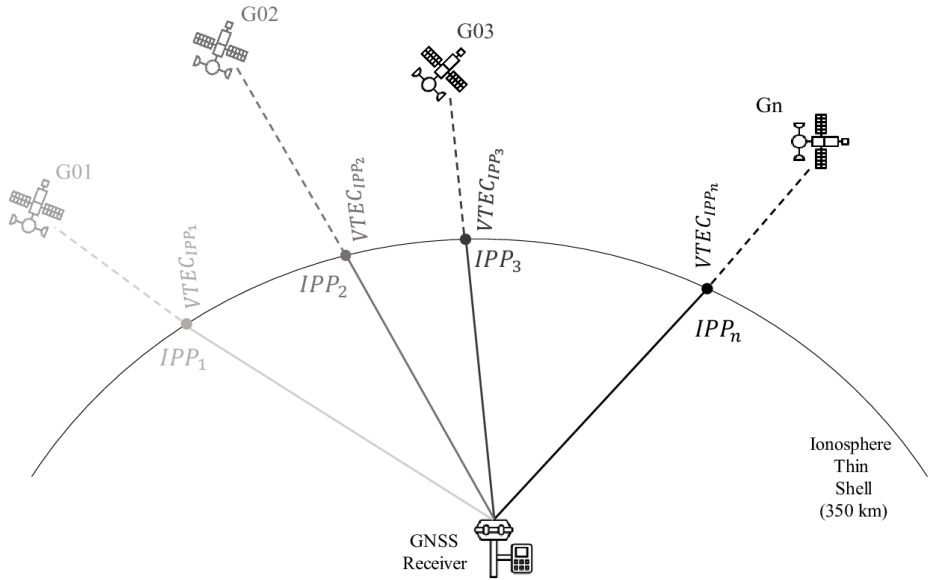

Fig. 2.1 VTEC variations per individual PRN, at a specific time epoch. IPP points are the intersections of GNSS signal in the ionosphere assuming the theory of single layer model (SLM). Based on the assumption that adjacent points share the same ionosphere conditions, our model investigates GNSS method's efficiency in providing TEC estimations.

In addition, there is no recurrent feedback mechanisms among the artificial neurons. Therefore, such filters fail to approximate temporal dependencies and high abrupt changes in timeseries values with a high precision accuracy.

For this reason, recently deep machine learning have been proposed as an alternative paradigm for regression and classification [36]. Deep learning incorporates multiple hidden neurons and applies advanced learning algorithms, such as input compression and dimensionality data reduction, to handle the computational issues arising when a large number of neurons are considered. Recurrent Neural Networks (RNNs) are a class of networks allowing connections (feedback) between the nodes (neurons) in order to model the temporal behaviours of a timeseries signal [37]. Thus, RNNs are capable of handling the time-dependencies. RNNs can sufficient handle TEC observations and are able to be adapted in order to learn the temporal dependencies from context [31] and monitor the irregular ionospheric structure. However, RNNs fail to approximate more complex temporal dependencies, presenting also computational issues in computing the gradient during the learning process especially when a large number of neurons are employed (the so called vanishing gradient problem [39]). To address these limitations and simultaneously to retain the advantages of deep learning in approximating temporal dependencies, Long-Short Term Memory (LSTM) architectures have been recently proposed [41]. LSTM networks memorize temporal correlations of the signals, providing, therefore better modeling capabilities [42], [43], [44]. LSTMs can model complex sequences with various features, as in [66], [67], [42], where solar radio flux at 10.7 cm and magnetic activity indices are taken into consideration to provide more accurate results. However, the above mentioned studies focus on TEC prediction based on estimates at station level using data derived from TEC models, such as GIM maps or GPS TEC analysis software [68] and they do not predict ionosphere delays separately from every visible GNSS satellite, as in our approach.

Tang et al. [69] analyzed and compared the performances of TEC prediction models of ARIMA, LSTM and seq2seq under the condition of a magnetic storm, and the result shows that overall prediction performance of the LSTM is the best during a magnetic storm. Sun et al. [42] analyzed and compared the accuracy of TEC prediction for MLP and LSTM models, and the result shows that the prediction precision of LSTM is better than for MLP under quiet and stormy geomagnetic conditions. Srivani et al. [68] investigated the performance of ionospheric delay prediction for NN, LSTM, and IRI models. Their research found that LSTM has best accuracy in the ionospheric delay prediction. These studies all prove that the performance of LSTM in TEC prediction is better than for existing empirical models [70]. Ruwali et al. [71] provides the application of deep learning models, long short-term memory (LSTM), gated recurrent unit (GRU), and a hybrid model that consists of LSTM combined with convolution neural network (CNN) to forecast the ionospheric delays for GPS signals.

## 2.2   Non-linear Ionosphere TEC modeling problem formulation

Our goal is to construct robust regional models of the ionosphere variability, applied in single-frequency PPP methods as external ionosphere correction information. A typical model assumes that the ionosphere is a thin shell above the Earth, located near the mean altitude of maximum TEC (approximately 350 km). The intersection between a signal's line of sight and this shell is called the Ionospheric Pierce Point (IPP). At different epochs, at every IPP point, each individual satellite provides a different VTEC measurement (Figure 2.1). In our model, we use as input the IPP points coordinates, noted by $x_\phi$ and $x_\lambda$. In addition, the information of daily time hour $x_{dt}$ has been incorporated into the model to boost its performance. Let us denote as $vtec^s(t)$ the value of the VTEC at a time instance $t$ for a visible satellite $s$. Then, we have that

$$vtec^s(t) = f(\mathbf{x}(t)) + e(t) \qquad (2.1)$$

$$vtec^s(t) = f(x_\phi(t),\, x_\lambda(t),\, x_{SSN}(t),\, x_{DST}(t),\, x_{F10.7}(t),\, x_{Ap}(t),\, x_{Kp}(t),\, x_{dt}(t)) + e(t) \qquad (2.2)$$

In Eq. (2.2) $f(\cdot)$ refers to the nonlinear relationship between the VTEC values and the inputs of solar and geomagnetic indices. It is clear that this non-linear relationship is actually unknown. In Chapters 3, 4, 5 we estimate the model parameters (weights) used to approximate $f(\cdot)$. These weights are estimated through the application of a supervised learning methodology [36]. In Eq. (2.2), variables $x_{SSN}$, and $x_{F10.7}$ refer to the input solar indices and $x_{DST}$, $x_{K_p}$, $x_{A_p}$ are the geomagnetic indices used as input variables of the proposed non-linear model.

In the data-driven supervised approach, we feed our proposed networks with pairs of data inputs-outputs, during training. Thus:

$$\mathbf{x}_{tr} = \begin{bmatrix} x_\phi(t) & x_\lambda(t) & x_{dt}(t) & x_{SSN}(t) & x_{F10.7}(t) & x_{DST}(t) & x_{K_p}(t) & x_{A_p}(t) \end{bmatrix}^T \quad and \quad \mathbf{vtec}_{tr} = vtec(t) \tag{2.3}$$

after training has been completed, we test the model providing the $(\mathbf{x})_{test}$ values and expecting the outcome $(vtec)_{test}$.

## 2.3   The role of GNSS observations in TEC modeling using Deep Learning

GNSS observables are used here in order to estimate the ground truth values of vertical TEC (VTEC) $vtec_{tr}$ that are used during the training. The workflow in order to compute the $vtec_{tr}$ values that are necessary for training of the deep learning algorithms is as follows: we first apply dual-frequency undifferenced and unconstrained PPP model to estimate Slant TEC (STEC) values. These values are then separated from satellite and receiver DCBs (Differential Code Biases). Then, having pure STEC values we convert them to Vertical TEC (VTEC) ones. These VTEC ground truth (labelled) values are combined with solar and geomagnetic indicators to construct the TEC model using a supervised learning framework [36], as explained in Section 2.2. For this reason, in the following, we describe the methodology adopted for VTEC values estimation.

The code $P$ and phase $\phi$ observations in a given frequency band $f_i$ between a receiver $r$ and a GNSS satellite $s$, are written as [72]:

$$P^s_{(fi)_r} = \rho^s_r + c \cdot (dt_r + dt^s) + d_{(fi)_r} - d^s_{fi} + T^s + I^s_{(fi)} + \epsilon^s_{P_{(fi)}} \tag{2.4}$$

$$\phi^s_{(fi)_r} = \rho^s_r + c \cdot (dt_r + dt^s) + \delta_{(fi)_r} - \delta^s_{fi} + T^s - I^s_{(fi)} + \lambda N^s_{(fi)} + \epsilon^s_{\phi_{(fi)}} \tag{2.5}$$

where $P^s_{(fi)_r}$ and $\phi^s_{(fi)_r}$ denote pseudorange and carrier phase observables, respectively; $\rho^s_r$ is the geometric distance between the receiver to the satellite; $dt_r$ and $dt^s$ are the receiver and satellite clock offsets, respectively; $d_{(fi)_r}$ is the frequency-dependent receiver uncalibrated code delay (UCD) while $d^s_{fi}$ is the frequency dependent satellite UCD; $T^s$ is troposphere delay; $I^s_{(fi)}$ is the line of sight (LOS) ionospheric delay on the frequency $fi$; $\delta_{(fi)_r}$ and $\delta^s_{fi}$ are the frequency-dependent receiver and satellite uncalibrated phase delay, respectively; $N^s_{(fi)}$ is the phase ambiguity; $\epsilon^s_{P_{(fi)}}$ and $\epsilon^s_{\phi_{(fi)}}$ are the sum of measurement noise and multi-path error for pseudorange and carrier phase observations.

In the case of dual-frequency GPS observations and assuming the frequencies $f_1$ and $f_2$ noted as "1" and "2", respectively, the equation (1) is written as:

$$P^G_{1_r} = \rho^G_r + c \cdot (dt_r + dt^G) + d_{1_r} - d^G_1 + T^G + I^G_1 + \epsilon^G_{P_1} \tag{2.6}$$

$$P^G_{2_r} = \rho^G_r + c \cdot (dt_r + dt^G) + d_{2_r} - d^G_2 + T^G + I^G_2 + \epsilon^G_{P_2} \tag{2.7}$$

The code biases are commonly referred as differential code biases (DCBs): $DCB = DCB_{P1/P2} = d_{f_1} - d_{f_2}$ and also, given that $\gamma_2 = f_1^2/f_2^2$, we have:

$$d_{f_1} = d_{IF} + 1/(1-\gamma_2) \cdot DCB \ \ and \ d_{f_2} = d_{IF} + \gamma_2/(1-\gamma_2) \cdot DCB \tag{2.8}$$

and then,

$$d_{f_{1_r}} - d_{f_1}^G = d_{(IF)_r} - d_{IF}^G + + \frac{1}{(1-\gamma_2)} \cdot (DCB_r - DCB^s) \tag{2.9}$$

$$d_{2_r} - d_2^G = d_{(IF)_r} - d_{IF}^G + + \frac{\gamma_2}{(1-\gamma_2)} \cdot (DCB_r - DCB^s) \tag{2.10}$$

The term $\tilde{I}_1$ is computed as:

$$\tilde{I}_1 = I_1 - \frac{1}{(1-\gamma_2)}DCB^s + \frac{1}{(1-\gamma_2)}DCB_r \tag{2.11}$$

The Uncombined PPP (UPPP) model computes the ionosphere delay as unknown parameter, in contrast to the traditional ionosphere-free (IF) model which combines multiple frequency observations to eliminate the ionospheric error. Our model is ionosphere-constrained dual-frequency PPP model that estimates the STEC values. These values are then separated from satellite and receiver DCBs (Differential Code Biases), using the products of IGS service [73]. Finally, the VTEC values, called with the variable *vtec*, are estimated as

$$vtec = \frac{1}{MF_I} \cdot (\tilde{I}_1 + \frac{1}{(1-\gamma_2)}DCB^s - \frac{1}{(1-\gamma_2)}DCB_r) \tag{2.12}$$

The above equation evaluates the ground truth VTEC values. Then, the Slant TEC, denoted with the variable *stec*, is converted into *vtec* through the mapping function $MF_I$ [74]:

$$MF_I = \frac{stec}{vtec} = \frac{1}{(1 - (\frac{R_e}{R_e + h_s}cos\theta)^2)^{1/2}} \tag{2.13}$$

where $R_e$ is the mean Earth's radius; $\theta$ is the satellite's elevation angle; $h$ is the height of the ionospheric layer and usually has been taken about $350 \, km$.

The GNSS pre-processing is implemented using the GAMP [72] GNSS processing software. Further details about the experimental setup are provided in the Table 5.1.

# Chapter 3

# Applying an LSTM architecture for TEC modeling

In this chapter we propose a deep learning scheme in order to accurately estimate TEC values. Recurrent Neural Networks (RNNs) are employed as the formalism for solving the ionosphere variations modeling problem at hand. RNN is a class of neural network that allows previous outputs to be used as inputs, has hidden states and also, uses weights that are shared across time. Extending RNN's ability to accurately model temporal dependencies in timeseries, Long Short-Term Memory networks (LSTMs) are proposed here. LSTM is a special variant of the typical RNN structure that deals with the vanishing gradient problem encountered by traditional RNNs. Adopting LSTM method for ionosphere variations prediction is a suitable approach, as this is a time-sequence prediction problem. Recently, encouraging results have been obtained by employing LSTM-based configurations in domains such as language modelling, speech recognition and action recognition [75–77]. As already mentioned, ionosphere modeling is a problem of high *complexity*, as TEC values are *space− and time−* varying. Our model exploits LSTM [76] model's non-linearity, which is learned from the data. On the contrary, traditional time series prediction models, such as AR and ARMA [53], are particularly simple structures which are essentially linear update models plus some noise thrown in.

Our model handles TEC values at a micro-scale level and examines how consistent the extracted TEC values per observed satellite are, and whether the remaining biases affect them. Therefore, our analysis examines the ability of a GNSS data-based method to provide accurate estimates of ionosphere variations in the likely presence of unaccounted noisy conditions. Then, an ensemble solution at station-level is extracted with proper combination of the information about TEC variations that provided from the analysis of each satellite results.

In this work, we create regional TEC models with autoregressive character, using deep learning methods. Our intention, as a next step, is to extend the applicability of this approach towards the implementation of this model in near-real time PPP-RTK processing in order to use it as part of the integer ambiguity resolution (IAR)–enabled PPP owing to the use of predicted ionospheric delays in addition to other corrections. Until now, CODE provides GIM maps in a grid of $(2.5° × 5.0°)$, in 2-h temporal resolution, leading to a spatio-temporal sparse model. This means that the corrections applied improve the accuracy of the provided solution for positioning, but fail to remove the total amount of noise caused by the ionosphere. Stations or roving users near the reference area in which regional LSTM models have been created, could use corrections from the regional model and not a global one.

The contribution of this study is to explore the potential of LSTM in time series analysis for TEC modeling and prediction. Compared with traditional time series prediction models, LSTM can obtain better prediction results, by taking advantage of the following features of the method:

- **Sequential modeling**: An important contribution is that our non-linear model is based on the so called *sequence-to-sequence* (Seq2seq) modeling framework; a family of machine learning algorithms to transform one sequence into another sequence. Seq2seq methods have been originated by Google engine for text/speech language translation [78], but they have recently been adopted in solving complex timeseries problems in power energy domain [54]. The Seq2seq structure permits the prediction of the ionosphere values for all satellites in view at every epoch [79]. Given that there are multiple input and output time steps, this problem is referred to as *'many-to-many'* sequence prediction problem. The Seq2seq modeling approach means that we are able to model not only the electron density values at the next epoch, but a sequence of successive epochs.

- **Non-linear modeling of temporal dependencies**: LSTMs with their internal memory, remember previous information that reflects the past behavior of the ionosphere and find patterns across time for accurate prediction of the next guess–estimates, making them ideal for time series prediction.

The outline of the chapter is as follows. Section 3.2 describes the proposed LSTM model. In Section 3.3, experimental results and analyses are provided, together with various comparisons, as for instance: (i) between traditional AR and ARMA methods, as well as, (ii) with well known TEC models. Conclusions are given in the Section 3.4.

## 3.1   LSTM Regression Model for Ionospheric Correction

### 3.1.1   Unidirectional Recurrent Neural Network for VTEC estimation

Our proposed deep learning model is based on timeseries modelling. The inputs during training are not independent but they follow a time dependent order. In contrast to other deep learning schemes, mostly applied in classification problems, in which the common procedure is to randomize the entire input data, here we form our data in a time-dependent sequence to train our network. Thus, the proposed network is specially designed for time-series modelling and infers successfully the time-related information.

Assuming L hidden neurons and one linear output layer, the estimate $vtec^s$ every epoch, is given by [33]

$$vtec^s(t) = \mathbf{h}^T(t) \cdot \mathbf{v} \tag{3.1}$$

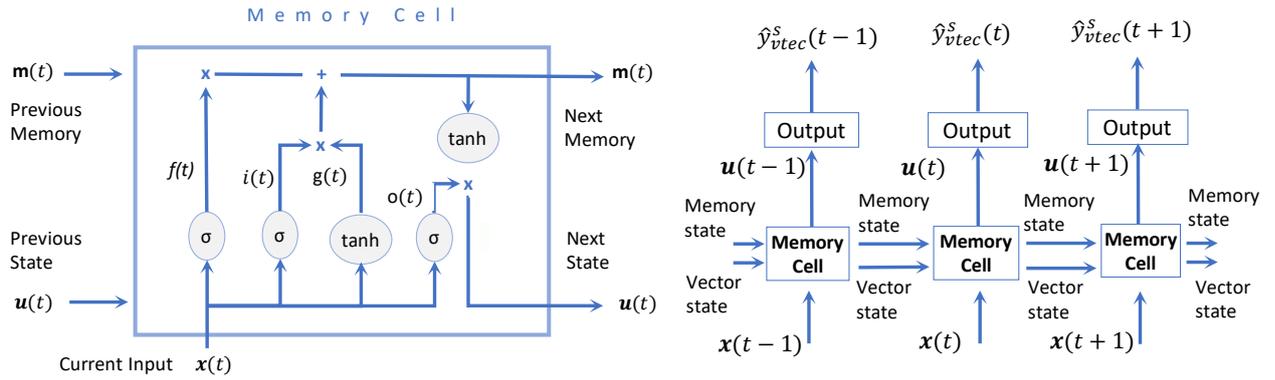

Fig. 3.1 The memory cell of an LSTM network. It contains three different components; (i) the forget gate f(t), (ii) the input gate i(t) and the input node g(t) and (iii) the output node o(t) (**right**). Unidirectional LSTM regression model used for TEC estimation (**left**).

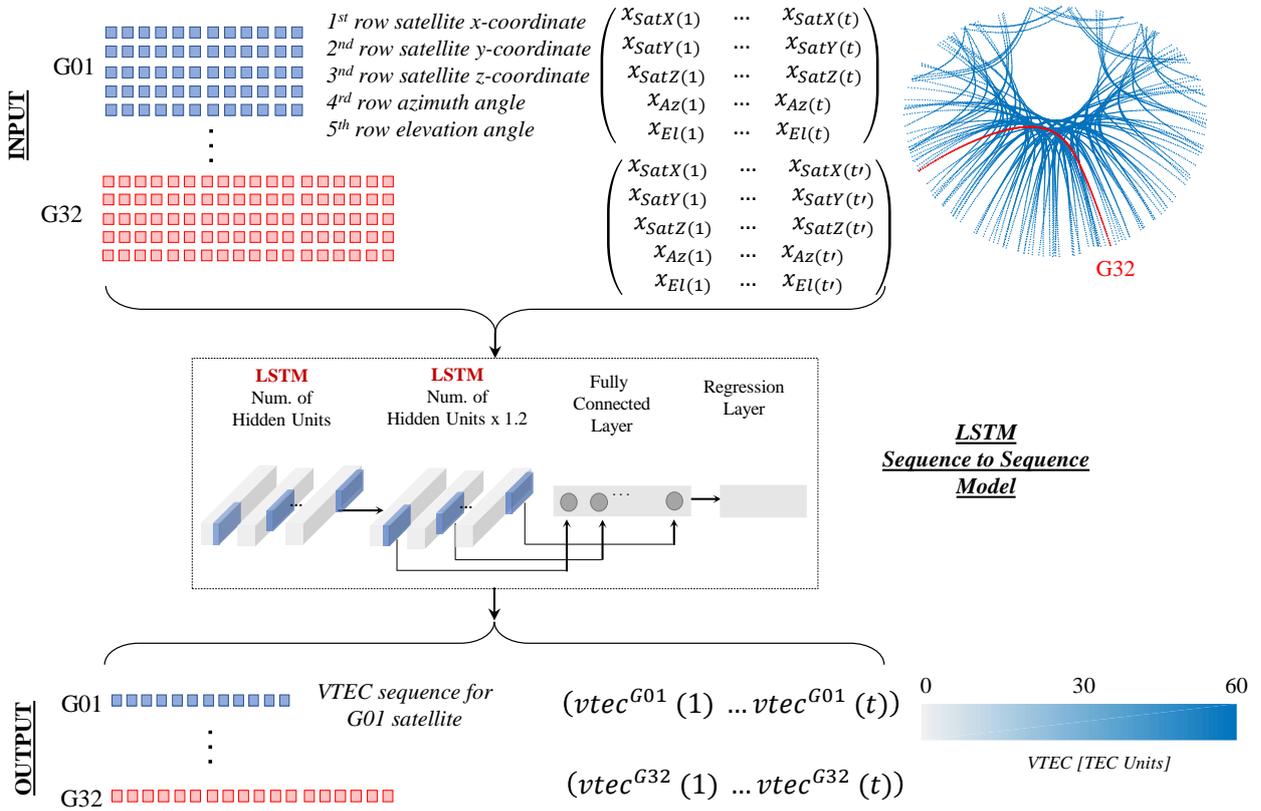

Fig. 3.2 This proposed LSTM architecture for TEC sequence ($vtec^s(1)... \ vtec^s(t)$) prediction. The deep learning structure consists of two stacked LSTM layers. The proposed LSTM is a sequence to sequence regression model. The output is TEC sequences from GNSS observations at a given station which are of varying length (e.g., $t$, $t'$) and the input has as features the satellite position vector [$x_{SatX}$ $x_{SatY}$ $x_{SatY}$], the azimuth ($x_{Az}$) and the elevation angles $x_{El}$ of the satellite direction from the receiver site. For example satellite G32 with a trajectory as illustrated in the figure with red line, is visible from station bor1 for about 5 h, thus the input sequence for G32 satellite has length 600 values, for the measurements sampling rate at 30 s.

$$\mathbf{h}(t) = \begin{bmatrix} h_1(t) \\ \vdots \\ h_L(t) \end{bmatrix} = \begin{bmatrix} \sigma(\mathbf{w}_1^T \cdot \mathbf{x}(t)) \\ \vdots \\ \sigma(\mathbf{w}_L^T \cdot \mathbf{x}(t)) \end{bmatrix} \tag{3.2}$$

$\sigma(\cdot)$ that refers to the sigmoid function. The weights $w_i, i = 1, \ldots, L$, connect the input $\mathbf{x}(t)$ with the i-th hidden neuron. Similarly, $\mathbf{v}$ are the weights that connect the hidden neurons with the output neuron. The term $h(t)$ gathers the outputs of all L hidden neurons $h_i$ with values ranging between 0 and 1.

Since the ionosphere delay follows a causal relationship, the value of a state depends on its previous values, forming a short-term recurrent neural network structure. Therefore,

$$u_i(t) = \sigma(\mathbf{w}_i^T \cdot \mathbf{x}(t) + h_i{}^T \mathbf{u}(t-1)) \tag{3.3}$$

### 3.1.2 Unidirectional Long-Term Recurrent Neural Network for VTEC estimation

Long Short-term Memory (LSTM) [76] is an RNN architecture designed to be better at storing and accessing information than standard RNNs.

In most RNNs (see Section 3.1.1), the hidden layer function $h$ is an elementwise application of a sigmoid function. However, it has been proved that the LSTM architecture [], which uses purpose-built memory cells to store information, is better at finding and exploiting long range dependencies in the data. LSTM has recently given state-of-the-art results in a variety of sequence processing tasks, including speech and handwriting recognition [80].

An LSTM network is a variation of the classic RNN model, enriched with slowly changing weights to address the vanishing gradient problem that RNN suffer from. LSTM networks have internal mechanisms called gates that can regulate the flow of information, maintain the worth-remembering pieces of information and forget the unnecessary ones. LSTMs are of similar structure with the recurrent regression model, but each node in the hidden layer is replaced by a memory cell, instead of a single neuron [39]. It processes the available data passing on information as it propagates forward.

The memory cell contains three different components: (i) the forget gate $f(t)$, (ii) the input gate $i(t)$, (iii) the cell candidate $g(t)$ and (iii) the output gate $o(t)$. For each component, a non-linear relation to the inner product between the input vectors and respective weights is applied during the training process. In some of the components the sigmoid function $\sigma(\cdot)$ is applied, while in others the hyperbolic tangent function $\tanh(\cdot)$ is used. Forget gate $f(t)$ keeps unnecessary information out of memory cell, thus separating the worth-remembering information from the useless one [81]. Input gate $i(t)$ regulates whether the information is relevant enough to be applied in future steps for the accurate estimation of TEC values. Cell candidate $g(t)$ activates appropriately the respective state (true or false output from the tanh activation). Output gate $o(t)$ decides if the response of the current memory cell is "significant enough" to contribute to the next cell.

$$\begin{bmatrix} i(t) \\ f(t) \\ o(t) \\ g(t) \end{bmatrix} = \begin{bmatrix} \sigma \\ \sigma \\ \sigma \\ \tanh \end{bmatrix} W \begin{bmatrix} x(t) \\ h(t-1) \end{bmatrix} + \begin{bmatrix} b_i \\ b_f \\ b_o \\ b_g \end{bmatrix} \tag{3.4}$$

The cell state at time step t is given by

$$c(t) = f(t) \odot c(t-1) + i(t) \odot g(t) \tag{3.5}$$

where $\odot$ denotes the Hadamard product (element-wise multiplication of vectors).

The hidden state at time step $t$ is given by

$$h(t) = o(t) \odot \sigma(c(t)) \tag{3.6}$$

where $\sigma$ denotes the state activation function.

It should be clarified that the nomenclature "Long term" used for LSTM networks, is a deep learning related terminology and is associated with the model's ability to learn from previous historic past values, and should not be confused with ionosphere's long term temporal variations that implies an analysis of many years. In our approach we are trying to model daily variations of the ionosphere and not long term ones.

### 3.1.3   The Proposed LSTM Regression Model for TEC Prediction

Figure 3.2 shows the adopted strategy and the basic structure of the proposed LSTM model. The network configuration setup consists of: (i) the input layer, (ii) two stacked unidirectional LSTM layers and (iii) the regression layer. In our case, we have a sequence-to-sequence model, in which data are organized as sequences of input and output pairs. This implies that LSTM model infers the relative time information during training, as it is not trained for every pair (input features, output) at every epoch t individually, but is trained to recognize a whole sequence/set of these pairs during the day, for a specific satellite. Thus, the input layer receives the current data in a sequence-like form, of a time window with varying duration. The network includes two bidirectional LSTM layers: the first layer consists of 60 filters with a kernel size 1 × 5, while the second layer consists of 72 filters of the same kernel size. The regression layer consists of one fully connected hidden layer and the final output regression layer. The output is a sequence of VTEC predicted values for each satellite, with size that covers the whole duration where the satellite is visible from the receiver.

In order to evaluate the ionospheric delay effects on GPS positioning, days between 10 and 13 December 2018 (day of year (DOY), from 344 to 347), where 80% of them has selected for training and the remaining 20% as validation set. Furthermore, 10 days have been selected as test set (14 December and 19 to 27 December 2018).

Table 3.1 Approximate location of six selected stations of IGS network and the countryt of their origin.

| Stations | Latitude° | Longitude° | Country |
|----------|-----------|------------|---------|
| bor1 | 52.276958352 | 17.073459989 | Poland |
| ganp | 49.034715044 | 20.322939096 | Slovakia |
| graz | 47.067131595 | 15.493484133 | Austria |
| leij | 51.353981980 | 12.374101226 | Germany |
| pots | 52.379299109 | 13.066095112 | Germany |
| wtzz | 49.144213977 | 12.878907414 | Germany |

## 3.2   PPP Processing Strategy

This section describes the pre-processing strategy to extract the necessary inputs and outputs for our LSTM model (see Chapter 2), using observational data from a small group of permanent GNSS stations of International GNSS Service (IGS) network. The time resolution is selected to be 30s. Table 4.1 shows the geographical distribution of the six selected IGS stations [82], which are spread across central Europe and are located in close proximity to each other.

We used the GAMP software [72], a secondary development software based on RTKLIB [83] for multi-constellation and multi-frequency Precise positioning. GAMP allows the use of undifferenced and uncombined observations in dual-frequency PPP processing to extract STEC values. Using the uncombined PPP (UPPP) model allows estimating the ionospheric effects as unknown parameters, without the need to impose ionospheric-free constraints imposed by the traditional PPP model which amplifies the measurement noise when dual frequency observations are combined to cancel the ionospheric effects. The necessary RINEX observation and navigation files, along with precise orbit and clock information, IGS ANTEX (igs14.atx) and SINEX files, as well as ocean tide loading coefficients and DCBs are entered into the GAMP software for static PPP processing. Wuhan University's satellite orbits and clock offsets were provided as input for GAMP software. Differential code biases (DCBs), are essential in many navigation and non-navigation applications (such as ionospheric analysis). As new signals are currently provided by the modernized GNSS systems, the need for a comprehensive multi-GNSS DCB product arises. DCB products, as part of the IGS Multi-GNSS Experiment (MGEX), are provided by the Chinese Academy of Sciences (CAS) in Wuhan (Table 8.2).

## 3.3   Results and Discussion

### 3.3.1   Performance Comparison among Different Methods

We compare the proposed LSTM method against traditional stochastic models for timeseries modeling and forecasting, such as the autoregressive AR model and the autoregressive moving average ARMA model. Our LSTM model has been trained and deployed using Python with Tensorflow and Keras libraries. We trained the model using the adaptive moment estimation optimization algorithm (adam) with a learning rate of $10^{-4}$. Model weights and coefficients are updated using a mini-batch size of 28 samples at each training iteration. The maximum number of epochs for training is selected to be 600. Deep learning performance is improved through data balance and normalization of the data used as input to the LSTM network to the range $[0, 1]$. We have individual networks, one per station. AR and ARMA models were also implemented in Python using the stats models library.

Table 3.2 Static PPP experimental setup.

| Options | Settings |
|---|---|
| Constellation | GPS |
| Positioning mode | static PPP |
| Frequencies | L1, L2 |
| Sampling rate | 30 s |
| Elevation mask | $7°$ |
| Differential code bias (DCB) | Correct using MGEX DCB products for PPP |
| Tropospheric zenith wet delay | initial model + estimated (random walk process) |
| Receiver and Satellite antenna | corrected with igs14.atx |
| Phase wind-up | Corrected |
| Sagnac effect, relativistic effect | Corrected with IGS absolute |
| Station reference coordinates | IGS SINEX solutions |

$$\overline{mae}_r = \frac{1}{S} \sum_{s_i=1}^{S} MAE_r^{s_i} = \frac{1}{S} \sum_{s_i=1}^{S} \left( \frac{\sum_{t=1}^{T} \left| vtec_{r,\,t}^{s_i} - \widehat{vtec}_{r,\,t}^{s_i} \right|}{T} \right) \tag{3.7}$$

$$\overline{min}_r = \frac{1}{S} \sum_{s_i=1}^{S} \left( \min_{\forall t \in T} \left( \left| vtec_{r,\,t}^{s_i} - \widehat{vtec}_{r,\,t}^{s_i} \right| \right) \right) \tag{3.8}$$

$$\overline{max}_r = \frac{1}{S} \sum_{s_i=1}^{S} \left( \max_{\forall t \in T} \left( \left| vtec_{r,\,t}^{s_i} - \widehat{vtec}_{r,\,t}^{s_i} \right| \right) \right) \tag{3.9}$$

$$\overline{rmse}_r = \frac{1}{S} \sum_{s_i=1}^{S} RMSE_r^{s_i} = \frac{1}{S} \sum_{s_i=1}^{S} \left( \sqrt{\frac{1}{T-1} \sum_{t=1}^{T} (vtec_{r,\,t}^{s_i} - \widehat{vtec}_{r,\,t}^{s_i})^2} \right) \tag{3.10}$$

Table 4.2 shows performance metrics between LSTM model, as well as ARMA and AR models. For each satellite $s_i$ in every station $r$, a sequence of VTEC values is produced using the TEC model, for a given time period $T$. For each individual satellite, we compute the absolute difference between the ground truth $\widehat{vtec}_{r,\,t}^{s_i}$ value, as obtained from the processing with the GAMP software, and the estimated from our model $vtec_{r,\,t}^{s_i}$ value, thus we have the difference $\left| vtec_{r,\,t}^{s_i} - \widehat{vtec}_{r,\,t}^{s_i} \right|$, every time/epoch.

For every station $r$, we have also computed two additional metrics for every individual PRN ($s_i$ -satellite): the mean absolute error $MAE_r^{s_i}$ (i.e., the average sum of all absolute errors) and the root mean squared error $RMSE_r^{s_i}$. However, Table 4.2 summarizes an overall accuracy for each station, using the following performance metrics: (i) $\overline{mae}_r$ is the average value of the combined $MAE_r^{s_i}$ for all $s_i$ satellite estimations per station, (ii) $\overline{min}_r$ is the average of all $s_i$ minimum difference values per station, (iii) in accordance to $\overline{min}_r$ metric, $\overline{max}_r$ is the average of all $s_i$ maximum differences values per station, and (iv) $\overline{rmse}_r$ is the average of $s_i$ root mean squared error of all PRNs (satellites) per station. LSTM, ARMA ans AR are causal models, thus they are appropriate for ionospheric time series modeling. However, our proposed unidirectional LSTM model performs best (Table 4.2) mainly due to its capability to effectively identify structure and pattern of TEC data, such as non-linearity and complexity in time series prediction and infer TEC behavior based on previous states.

Table 3.3 Performance metrics (mean, minimum (Min), maximum (Max) and root mean squared error (rmse)) for VTEC (TECU) prediction for six selected stations (bor1, ganp, graz, leij, pots and wtzz) among the proposed solution (LSTM) and two other commonly used methods (ARMA, AR).

| Receiver r Metric | **bor1** $\overline{mae}_r$ | $\overline{min}_r$ | $\overline{max}_r$ | $\overline{rmse}_r$ | **ganp** $\overline{mae}_r$ | $\overline{min}_r$ | $\overline{max}_r$ | $\overline{rmse}_r$ | **graz** $\overline{mae}_r$ | $\overline{min}_r$ | $\overline{max}_r$ | $\overline{rmse}_r$ |
|---|---|---|---|---|---|---|---|---|---|---|---|---|
| LSTM | **0.98** | 0.04 | **3.05** | **1.18** | **1.24** | **0.02** | **2.88** | **1.45** | **1.09** | 0.03 | **2.76** | **1.26** |
| ARMA | 1.41 | **0.01** | 3.76 | 1.76 | 1.63 | 0.02 | 3.94 | 2.03 | 1.70 | 0.04 | 3.85 | 2.03 |
| AR | 1.91 | 0.04 | 4.40 | 2.29 | 1.85 | 0.02 | 4.28 | 2.25 | 1.77 | **0.02** | 3.98 | 2.13 |

| Receiver r Metric | **leij** $\overline{mae}_r$ | $\overline{min}_r$ | $\overline{max}_r$ | $\overline{rmse}_r$ | **pots** $\overline{mae}_r$ | $\overline{min}_r$ | $\overline{max}_r$ | $\overline{rmse}_r$ | **wtzz** $\overline{mae}_r$ | $\overline{min}_r$ | $\overline{max}_r$ | $\overline{rmse}_r$ |
|---|---|---|---|---|---|---|---|---|---|---|---|---|
| LSTM | **0.93** | 0.02 | **2.66** | **1.14** | **1.19** | **0.01** | **3.52** | **1.42** | **1.03** | 0.04 | **2.68** | **1.23** |
| ARMA | 1.63 | **0.01** | 3.96 | 1.98 | 1.71 | 0.01 | 4.12 | 2.07 | 1.54 | **0.01** | 3.78 | 1.91 |
| AR | 1.87 | 0.01 | 4.22 | 2.22 | 1.94 | 0.01 | 4.33 | 2.30 | 1.79 | 0.01 | 4.10 | 2.15 |

Table 3.4 Computational time (in s) per method (LSTM, ARMA and AR) for three selected stations (bor1, graz and wtzz).

| | **Elapsed Time (s)** | | |
|---|---|---|---|
| **Stations** | **LSTM** | **ARMA** | **AR** |
| bor1 | 719 | 55 | 1 |
| graz | 559 | 58 | 3 |
| wtzz | 600 | 53 | 1 |

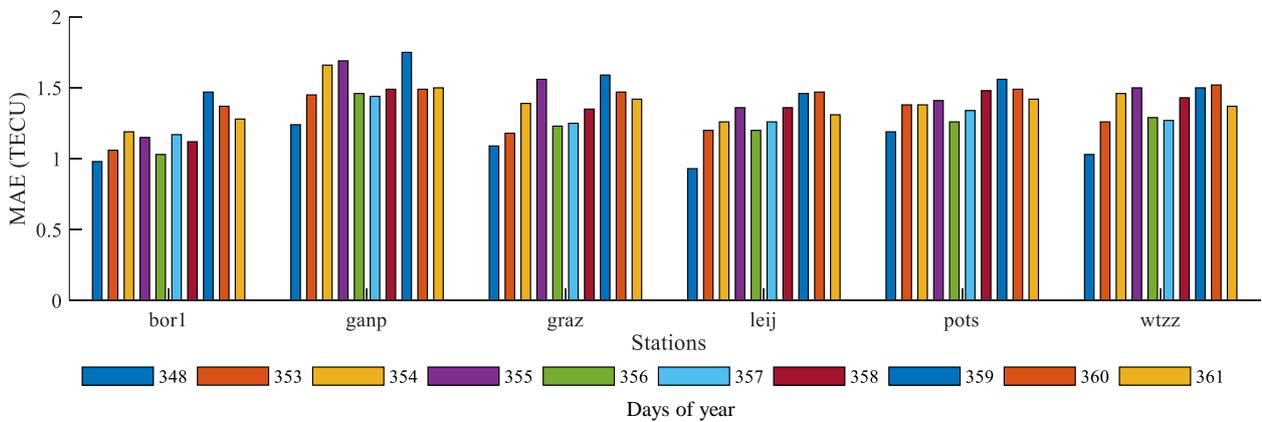

Fig. 3.3 MAE error per station for 10 different days.

LSTM model is undoubtedly, more complicated and difficult to train and in most cases does not exceed the performance of a simple ARMA model. Table 3.4 shows the computational time required for model training by each method. Evidently, in terms of computational load, the AR model requires the minimum time (1 s) for training, while LSTM is the most time-consuming model (>600 s), but this is to be anticipated mainly due to its complexity and its large number of trainable parameters. However, by inspecting the various statistical metrics, it can also be seen that the LSTM model achieves the best performance in terms of its agreement with the conventional GNSS-derived VTEC data.

### 3.3.2   Comparison among Different Stations and Time Periods

Figure 3.3 illustrates the $\overline{mae}_r$ metric for a testing period of 10 days, for the 6 different stations. MAE values ranging from 0.9 to 1.5 TECU. The 'ganp' station shows the worse values with low accuracy, while 'bor1' station achieves the highest accuracy among the other stations. Regarding the test period, the results are quite similar, however we observe a slightly increasing trend in MAE error, which is to be expected, as the predictions are carried out further away from the training period of the LSTM model.

### 3.3.3   Comparison between Different Satellites

In order to gain a further insight into the LSTM's model performance, we evaluated the VTEC prediction performance for each satellite. Figure 3.4 shows the estimated VTEC values for each station, per satellite on December 14, 2018. As noted, VTEC values show significant variation during the day, depending on the electron density in the ionosphere. The ionospheric delay changes slowly through a daily cycle, exhibiting its minimum values (2–4 TECU) between midnight and early morning, and reaching its peak (6–8 TECU) during the daylight hours, in the mid-latitudes.

As a next step, we evaluated the LSTM model's performance using the root mean squared error (RMSE) metric of VTEC values for each visible satellite, in a bar graph format. In general, RMSE TEC values are ranging between 0.5 and 2 TECU for each reference station (Figure 3.5). For station 'bor1', the highest RMSE values are observed for G07, G26 and G11 satellites in view. For station 'ganp', G18 and G11 show the highest error values. Station 'graz' exhibits significant variation between the satellites' RMSE values, with the satellites G15, G13 and G09 showing the maximum RMSE TEC values. Station 'leij' RMSE VTEC values reach the maximum for satellites G18 and G05. Satellite G18 has also the maximum RMSE for the other two remaining stations 'pots' and 'wtzz'.

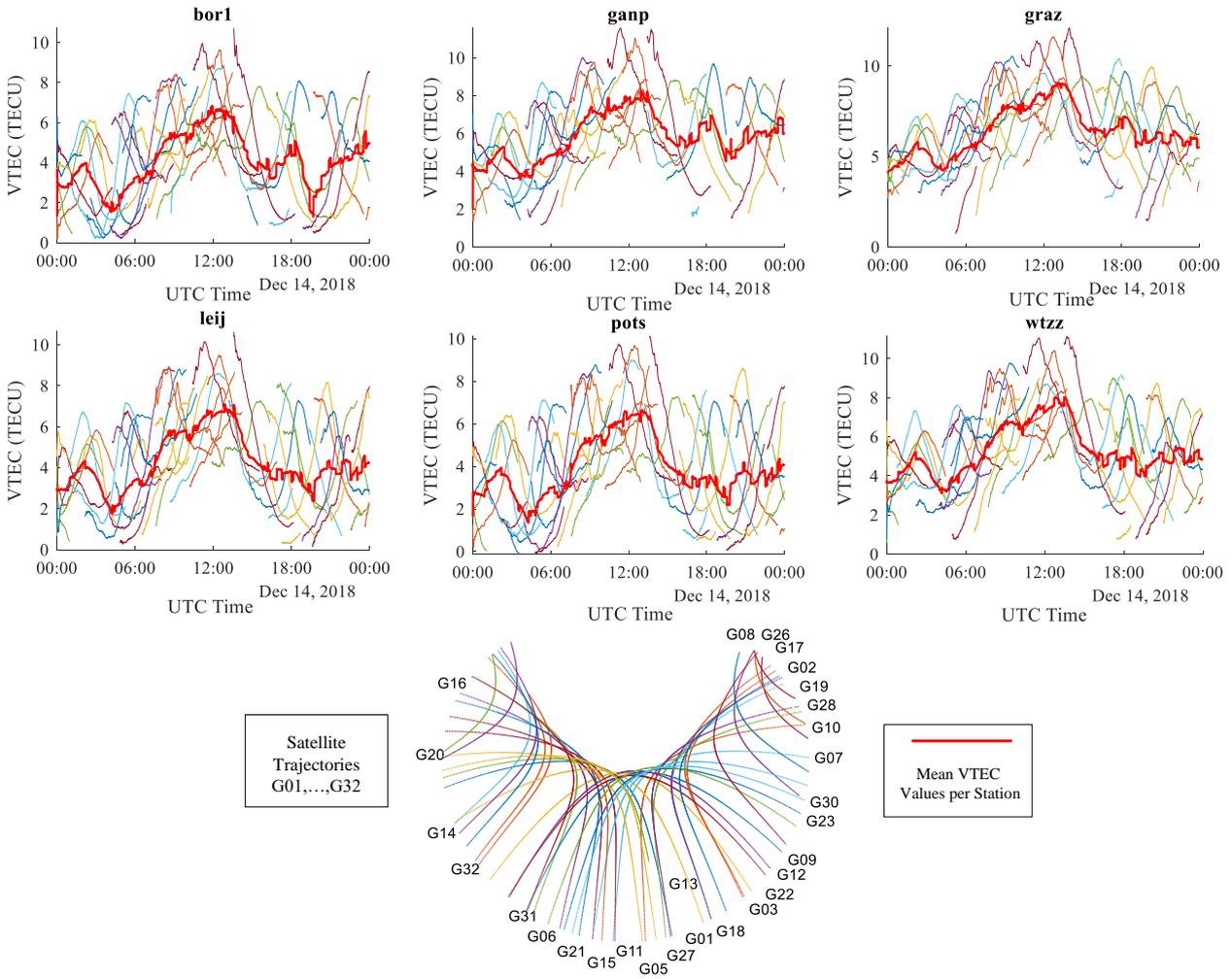

Fig. 3.4 VTEC (TECU) predictions on 14 December 2019 for six stations (bor1, ganp, graz, leij, pots, wtzz) per satellite ID (G1,..., G32 for GPS satellite), and mean VTEC value (with red line).

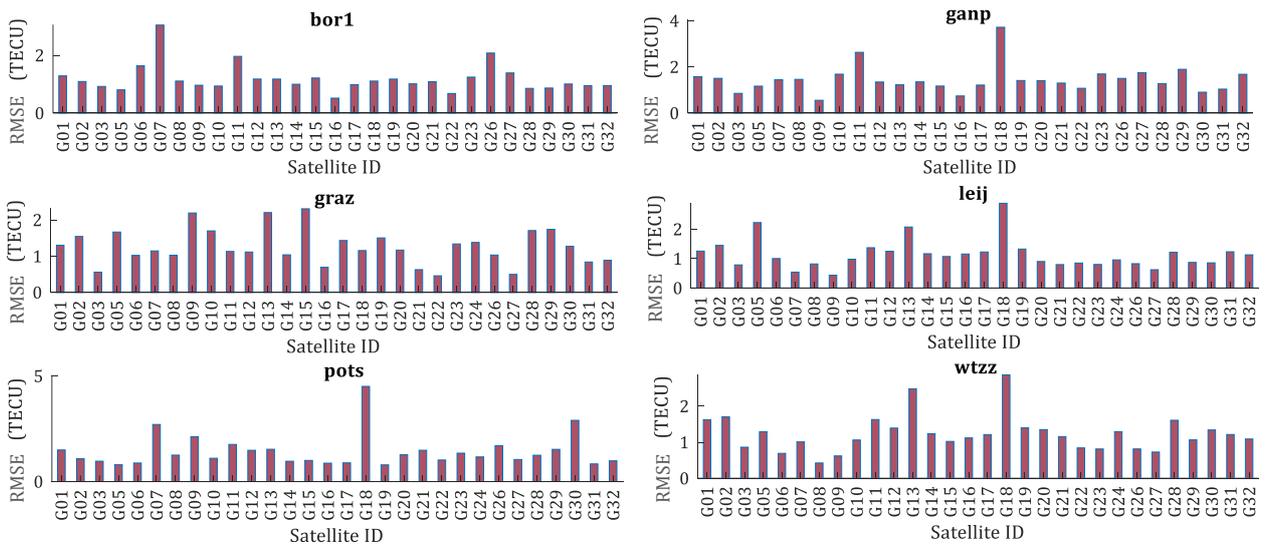

Fig. 3.5 Root Mean Squared Error (RMSE) metric of the VTEC(TECU) predictions using LSTM method per station (bor1, grap, graz, leij, pots, wtzz) and per satellite ID (G1,.., G32 for GPS constellation).

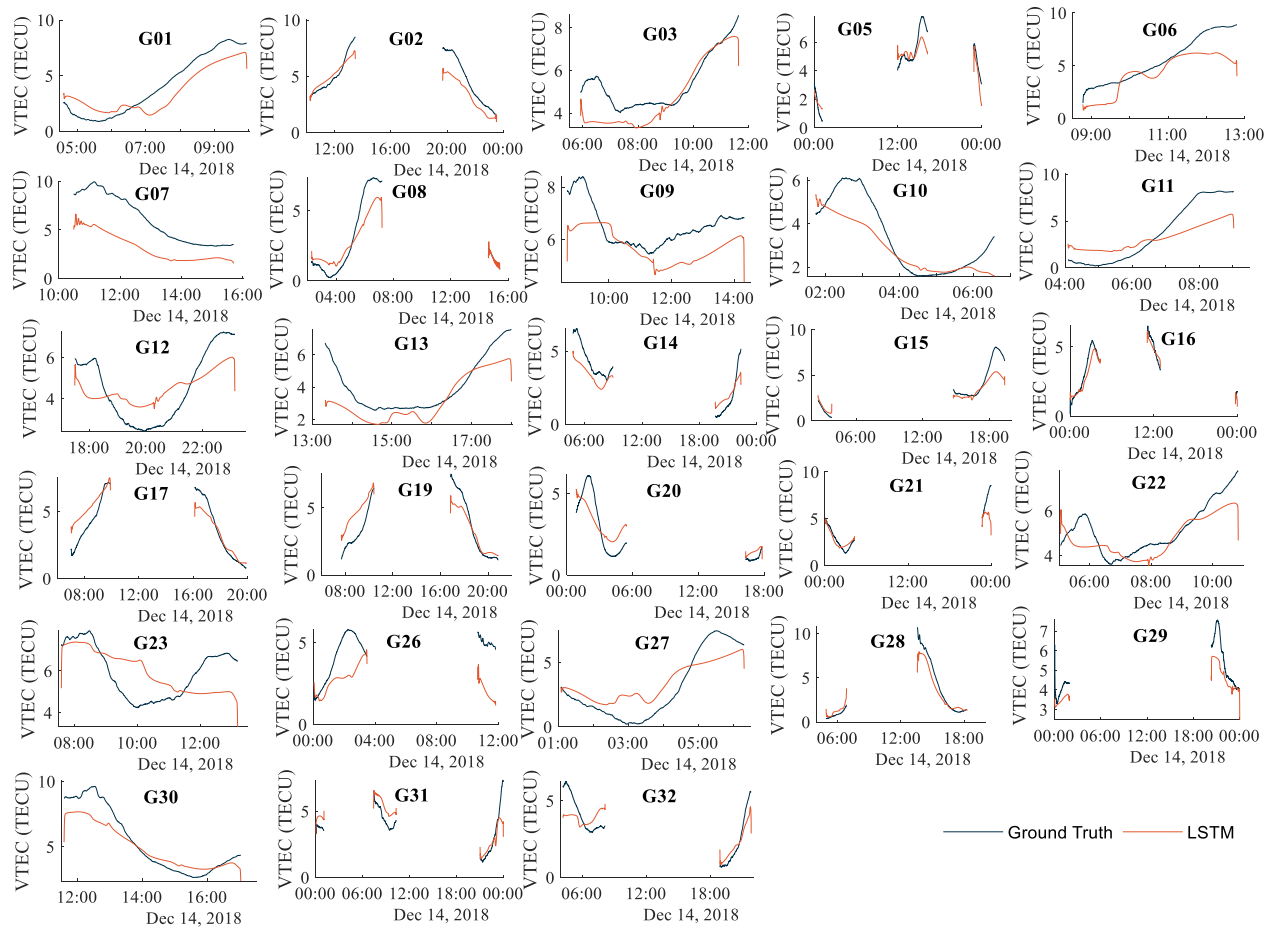

Fig. 3.6 Vertical TEC of each GPS satellite between the GPS ground truth values and our LSTM method's results on December 14, 2018, for 'bor1' station.

In particular, Figure 3.6 shows the predicted LSTM-aided vertical TEC values for each one of the GPS satellites (when they are visible from 'bor1' station), along with the vertical TEC values being derived from GAMP software (ground truth data). As noted, LSTM model can adequately predict VTEC's model performance for every satellite separately. The satellites G07, G11 and G26 have the worst performance, as expected, based on results from Figure 3.5. On the contrary, the satellites G05, G16 and G22 have the best performance in VTEC prediction.

### 3.3.4   Comparison among Different Ionosphere Models

To further analyze the LSTM model's performance, Figure 3.7 shows the mean VTEC values at every station, as obtained from NeQuick, IRI2001, IRI01-cor and GIM TEC estimates compared to the GPS-based TEC and LSTM TEC predictions, during the day. The VTEC maxima is appeared for all stations between 8:00 and 12:00 A.M. In most cases, LSTM predicted values are similar to GPS TEC derived values, however, it is noted that our model underestimates VTEC values, showing lower values than those of GPS TEC. GIM-aided TEC values are also close to LSTM derived TEC values. NeQuick, IRI2001 and IRI01-cor values show greater variability during the day with higher maximum and lower minimum than those of GPS TEC and GIM values.

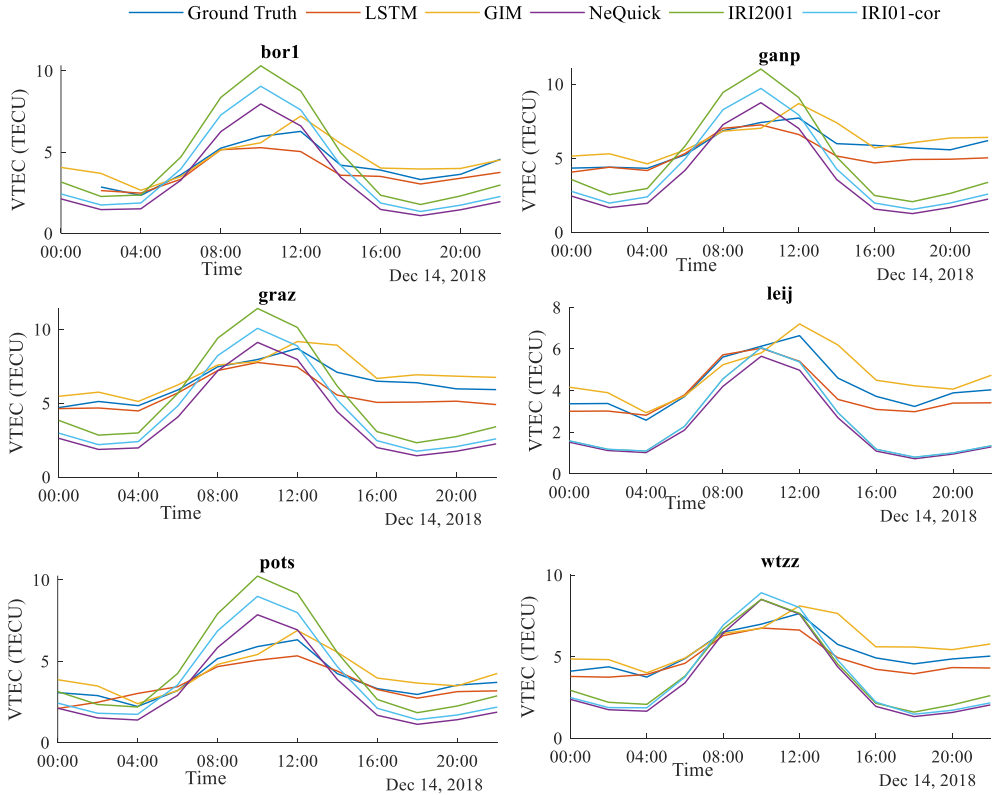

Fig. 3.7 Comparison between VTEC (TECU) LSTM predictions (PRD), estimated GPS ionospheric VTEC values (GRD) and the VTEC models (GIM, NeQuick, IRI2001 and IRI01-cor).

Figure 3.8 shows the mean difference values having as base GPS TEC estimates. As noted, in most cases LSTM predictions are closer to GPS TEC estimates, as they have small difference, except for station 'graz', where GIM model's estimates seem to have a slightly better performance.

## 3.4   Conclusions

We propose a supervised, unidirectional regression LSTM model for efficient prediction of vertical TEC values. Typical and widely used AR and ARMA models are linear models, suitable for TEC prediction, as ionosphere exhibits periodic changes. Thus, the AR and ARMA models' autoregressive character offers an advantage in VTEC prediction. However, the additional inclusion of selected features (such as the satellite vector, and the satellite elevation and azimuth) in the supervised LSTM algorithm, as well as, the fact that LSTM networks are equipped with memory cell and consequently they are able to preserve the worth-remembering information content in the input STEC data, gives superiority and an extra boost to our proposed method. Non-linearity and long-term prediction are additional advantages of our proposed LSTM method. As next steps, we intend to direct our studies into investigating the performance of this LSTM model under typical non-quiet conditions, including data from periods with intense ionospheric activity (e.g., periods with ionosphere anomalies and disturbances), and explore the possibility of using, with appropriate expansions, our LSTM model as a tool capable of adapting its performance to detect/catch extreme ionosphere conditions. To this end, alternative recurrent neural network methods, such as bidirectional LSTMs, that introduce non-causality, should be explored or/and check the performance of advanced Convolutional Neural Network (CNN) -based methods and how they can be optimally adapted in timeseries prediction problems for further benefit of the variant PPP techniques. In addition, we intend to explore models' behavior in datasets consisting of multiple GNSS constellations (such as GLONAS, Galileo, etc.).

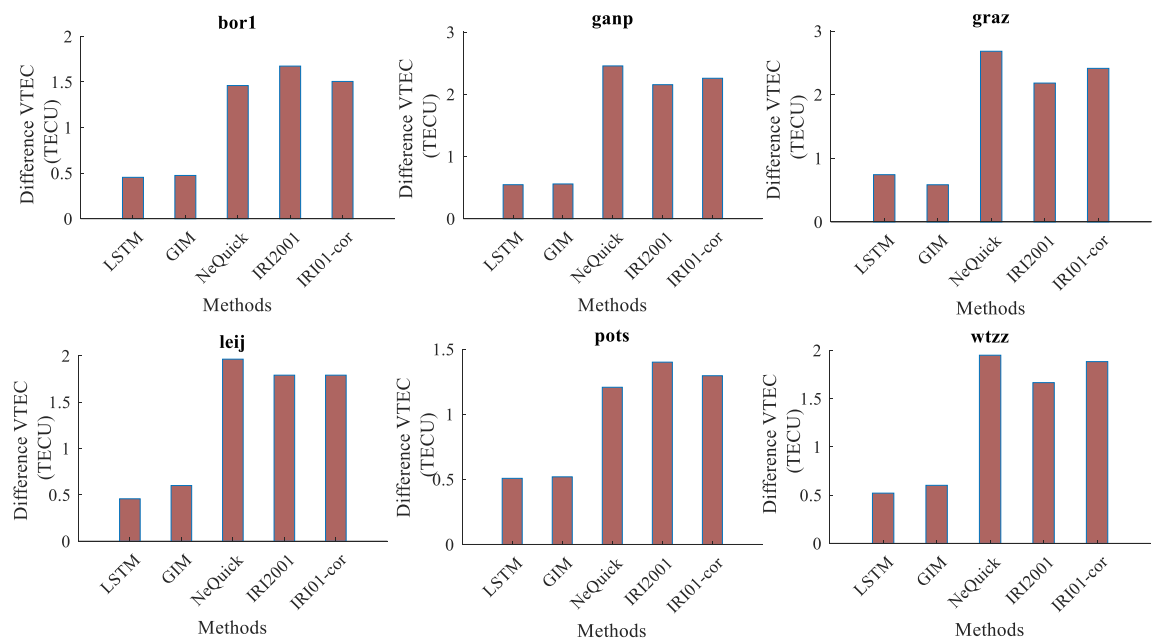

Fig. 3.8 Mean difference between the estimated GPS VTEC values and the VTEC (TECU) LSTM predictions (LSTM), GIM VTEC model values, NeQuick VTEC model values, IRI2001 VTEC model values and IRI01-cor VTEC model values.

# Chapter 4

# The adoption of a CNN architecture for TEC modeling and the adaptations needed for CNNs to solve a time series problem

Since AlexNet [84] won the ImageNet competition in 2012, deep CNNs have seen a lot of successful applications in many different domains [85] such as reaching human level performance in image recognition problems [86] as well as different natural language processing tasks [78, 87]. Motivated by the success of these CNN architectures in these various domains, researchers have started adopting them for time series analysis [88]. A convolution can be seen as applying and sliding a filter over the time series. Unlike images, the filters exhibit only one dimension (time) instead of two dimensions (width and height). The filter can also be seen as a generic non-linear transformation of a time series.

Furthermore, sequence-to-sequence (seq2seq) models [78] are implemented also here, to successfully catch the sequential character of TEC values and thus predict them for each visible satellite at a given station. In seq2seq models, the input is a time sequence of data, and the output is also a time sequence (see Chapter 3). Adopting this structure help us to estimate ionosphere TEC values at regional level.

The work in this chapter is further attempts:

- to propose a model with the ability of easily expanding the inputs variables,

- to derive potentially suitable models which can eventually be used to apply ionosphere corrections for single station and single frequency techniques,

- to propose an efficient model for constructing dynamic changing, regional TEC models.

## 4.1   GNSS AND IONOSPHERIC VARIABILITY

The ionosphere is that part of the earth's upper atmosphere (extending roughly between 70 and 600 km) with sufficient concentration of free electrons to affect the propagation of electromagnetic waves. Its existence is primarily the result of the absorption of solar ultraviolet radiation in that part of the atmosphere which in turn reacts to produce free electrons and ions. Total Electron Content (TEC) is often used to describe ionospheric variability and is space and time varying. it is widely known that ionosphere exhibits significant variations with:

- latitude and longitude: the most disturbed region is the aurora zone (between 60° to 70° N geomagnetic latitude) followed by the polar zone( >70° N), while irregularities at equatorial ionosphere follow,

- local time: during the sunny hours of the day, the ionospheric condition variations are higher than those in night-time period,

- solar cycle and geomagnetic activity.

The GNSS signal travels through the ionosphere and due to the severe spatio-temporal changes of the electron density, significant disruptions on the traveling GNSS radio wave are caused. As a consequence of ionosphere's dispersive nature, is that for different carrier wave frequencies, different delays are caused. This fact provides one of the greatest advantages of a dual-frequency receiver over the single-frequency receivers: appropriate mathematical combinations among the different frequencies can eliminate ionosphere delay error. Hence, the use of multiple navigation signals of distinct center frequency transmitted from the same GNSS satellite allows direct observation and removal of the great majority of the ionospheric delay. It is worth mentioned that, the severity of the ionosphere's effect on a GNSS signal depends on the amount of time that signal spends traveling through it. A signal originating from a satellite near the observer's horizon (low satellite elevation) must pass through a larger amount of the ionosphere to reach the receiver than does a signal from a satellite near the observer's zenith (high satellite elevation). Thus, the longer the signal is in the ionosphere, the greater the ionosphere's effect on it.

## 4.2   THE PROPOSED CNN FOR TEC MODELLING

### 4.2.1   Seq2seq temporal 1D CNN regression model for TEC prediction

At first, we model the unknown function $f(\cdot)$ through a feed-forward neural network. Assuming L hidden neurons and a linear output layer, the estimate $vtec^s$ every time, is given by [33]:

$$vtec^s(t) = \mathbf{u}^T(t) \cdot \mathbf{v} \tag{4.1}$$

$$\mathbf{u}(t) = \begin{bmatrix} u_1(t) \\ \vdots \\ u_L(t) \end{bmatrix} = \begin{bmatrix} \tanh(\mathbf{w}_1^T \cdot \mathbf{x}(t)) \\ \vdots \\ \tanh(\mathbf{w}_L^T \cdot \mathbf{x}(t)) \end{bmatrix} \tag{4.2}$$

Table 4.1 The stations' information (names, location (longitude and latitude) and country) for six selected sites of the IGS network.

| Site | Latitude° | Longitude° | Country |
|------|-----------|------------|---------|
| bor1 | 52.27695 | 17.07345 | Poland |
| ganp | 49.03471 | 20.32293 | Slovakia |
| graz | 47.06713 | 15.49348 | Austria |
| leij | 51.35398 | 12.37410 | Germany |
| pots | 52.37929 | 13.06609 | Germany |
| wtzz | 49.14421 | 12.87890 | Germany |

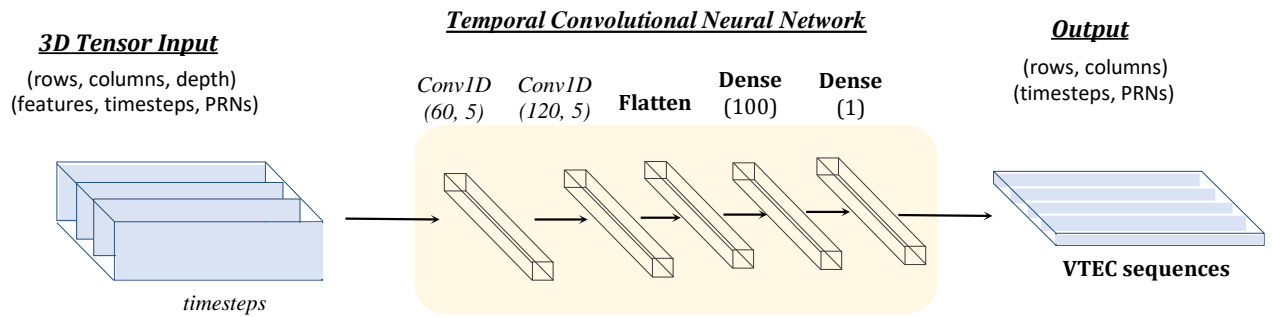

Fig. 4.1 Temporal CNN network architecture.

Table 4.2 A comparative table that summarizes the performance achieved by the CONV1D network in predicting the VTEC (TECU) values over the site stations (bor1, ganp, graz, leij, pots and wtzz).

| Site | bor1 | | | | ganp | | | | graz | | | |
|------|------|------|------|------|------|------|------|------|------|------|------|------|
| Metric | $\overline{mae}_r$ | $\overline{min}_r$ | $\overline{max}_r$ | $\overline{rmse}_r$ | $\overline{mae}_r$ | $\overline{min}_r$ | $\overline{max}_r$ | $\overline{rmse}_r$ | $\overline{mae}_r$ | $\overline{min}_r$ | $\overline{max}_r$ | $\overline{rmse}_r$ |
| CONV1D | **0.91** | 0.02 | **2.41** | **1.09** | **0.93** | 0.06 | **2.21** | **1.08** | **0.88** | 0.01 | **2.25** | **1.03** |
| LSTM | 0.98 | 0.04 | 3.05 | 1.18 | 1.24 | 0.02 | 2.88 | 1.45 | 1.09 | 0.03 | 2.76 | 1.26 |
| ARMA | 1.41 | **0.01** | 3.76 | 1.76 | 1.63 | 0.02 | 3.94 | 2.03 | 1.70 | 0.04 | 3.85 | 2.03 |
| AR | 1.91 | 0.04 | 4.40 | 2.29 | 1.85 | 0.02 | 4.28 | 2.25 | 1.77 | 0.02 | 3.98 | 2.13 |
| Site | leij | | | | pots | | | | wtzz | | | |
| Metric | $\overline{mae}_r$ | $\overline{min}_r$ | $\overline{max}_r$ | $\overline{rmse}_r$ | $\overline{mae}_r$ | $\overline{min}_r$ | $\overline{max}_r$ | $\overline{rmse}_r$ | $\overline{mae}_r$ | $\overline{min}_r$ | $\overline{max}_r$ | $\overline{rmse}_r$ |
| CONV1D | **0.71** | 0.02 | **2.07** | **0.88** | **0.84** | 0.02 | **2.39** | **1.03** | **0.83** | 0.03 | **2.25** | **0.99** |
| LSTM | 0.93 | 0.02 | 2.66 | 1.14 | 1.19 | 0.01 | 3.52 | 1.42 | 1.03 | 0.04 | 2.68 | 1.23 |
| ARMA | 1.63 | **0.01** | 3.96 | 1.98 | 1.71 | 0.01 | 4.12 | 2.07 | 1.54 | **0.01** | 3.78 | 1.91 |
| AR | 1.87 | 0.01 | 4.22 | 2.22 | 1.94 | 0.01 | 4.33 | 2.30 | 1.79 | 0.01 | 4.10 | 2.15 |

tanh($\cdot$) is the hyperbolic tangent, $x(t)$ is the input vector (see Equation 11), $w_i, i = 1, \ldots, L$, are the weights connect the input vector $\mathbf{x}(t)$, with the i-th hidden neuron. The term $u(t)$ gathers the outputs of all L hidden neurons $u_i$ with values ranging between -1 and 1. Similarly, $\mathbf{v}$ -vector (see Equation 12) encloses weights that connect the hidden neurons with the output neuron.

Convolutional Neural Network (CNN) is a more complex variant than a simple feed forward neural network, has its origins in the field of computer vision, but nowadays, due to its popularity, is spread in a wide range of applications. Its name is derived from the type of hidden layers that it consists of. Typically, the hidden layers of a CNN consist of convolutional layers, pooling layers, fully connected layers, and normalization layers. In case of one-dimensional convolutional neural networks, the major advantage is the model's low computational complexity since the only operation with a significant cost is a sequence of 1D convolutions which are weighted sums of two 1D arrays. For a given layer $l$, each (hidden or output) unit $z_k^l$ in such a network, computes a function given by:

$$z_k^l = f(b_k^l + \sum_{j=1}^{l-1} (w_{kj}^{l-1}, s_j^{l-1}))$$ (4.3)

where $x_k^l$ is defined as the output/hidden unit, $b_k^l$ is defined as the bias of the $k^{th}$ neuron at layer $l-1$, $s_j^{l-1}$ is the output of the $j^{th}$ neuron at layer $l-1$, $w_{kj}^{l-1}$ is the kernel from the $j^{th}$ neuron at layer $l-1$ to the $k^{th}$ neuron at layer $l$.

### 4.2.2   The CNN network configuration for TEC modelling

As illustrated in Figure 4.1, our proposed network configuration consists of two convolutional layers. The first convolutional layer has 60 filters of height 5 and the second one 120 filters of the same size. Then, a layer that flattens the output of convolutional neural network layer follows. Dense layer assigns a linear operation in which the input of the previous layer is connected to the output by a weight and is followed by a non-linear activation function. In our case, the activation function is the hyperbolic tangent function. For the training, the number of epochs is selected 600. The optimizer is the Adam and the loss function is the mean squared error.

## 4.3   EXPERIMENTAL RESULTS

### 4.3.1   Dataset

The experimental setup consists of a selected small group of permanent GNSS stations of the global network of International GNSS Service (IGS). The time granularity of the data is 30s. Table 4.1 shows the position of the selected ground receiver IGS stations [82], across central Europe in close proximity to each other.

The GAMP software [72], a secondary development software based on RTKLIB [83] has been used for precise point positioning. We have used the uncombined PPP (UPPP) ionosphere constraint model to estimate the slant TEC values as unknown parameters. The observation and navigation files, the precise orbit and clock information, the antenna phase centre corrections for both receivers network and satellites, as well as ocean tide loading coefficients and the differential code biases are processed using the GAMP software in static PPP mode.

### 4.3.2 Performance Evaluation

Table 4.2 provides comparative results for the proposed CONV1D method and (i) the recurrent LSTM (Long Short-term Memory) network [54], (ii) the AR model (Autoregressive) and (iii) the ARMA model (Autoregressive moving Average) model. AR and ARMA models are traditional stocha-stic models used for timeseries modeling, while recurrent neural networks allow previous outputs to be used as inputs for the next step, enhancing their recurrent character and their ability to successfully deal with sequential data. For each individual PRN, we compute the absolute difference between the ground truth $\widehat{vtec}_{r,\,t}^{s_i}$ value, as computed using the GAMP software with PPP processing, and the respective VTEC values as being estimated from our Conv1d model $vtec_{r,\,t}^{s_i}$. Thus, every epoch $t$, the absolute difference is $\left| vtec_{r,\,t}^{s_i} - \widehat{vtec}_{r,\,t}^{s_i} \right|$. The metrics for comparison where selected to be:

(i) the mean absolute error $\overline{mae}_r$, which is the average value of the mean absolute errors $MAE_r^{s_i}$ per individual PRN $s_i$,

$$\overline{mae}_r = \frac{1}{S} \sum_{s_i=1}^{S} MAE_r^{s_i} = \frac{1}{S} \sum_{s_i=1}^{S} \left( \frac{\sum_{t=1}^{T} \left| vtec_{r,\,t}^{s_i} - \widehat{vtec}_{r,\,t}^{s_i} \right|}{T} \right) \tag{4.4}$$

(ii) $\overline{min}_r$ is the average of all $s_i$ minimum difference values per station,

$$\overline{min}_r = \frac{1}{S} \sum_{s_i=1}^{S} \left( \min_{\forall t \in T} \left( \left| vtec_{r,\,t}^{s_i} - \widehat{vtec}_{r,\,t}^{s_i} \right| \right) \right) \tag{4.5}$$

(iii) in accordance to $\overline{min}_r$ metric, $\overline{max}_r$ is the average of all $s_i$ maximum differences values per station, and

$$\overline{max}_r = \frac{1}{S} \sum_{s_i=1}^{S} \left( \max_{\forall t \in T} \left( \left| vtec_{r,\,t}^{s_i} - \widehat{vtec}_{r,\,t}^{s_i} \right| \right) \right) \tag{4.6}$$

(iv) $\overline{rmse}_r$ is the average of $s_i$ root mean squared error $RMSE_r^{s_i}$ of all PRNs (satellites) per station.

$$\overline{rmse}_r = \frac{1}{S} \sum_{s_i=1}^{S} RMSE_r^{s_i} = \frac{1}{S} \sum_{s_i=1}^{S} \left( \sqrt{\frac{1}{T-1} \sum_{t=1}^{T} (vtec_{r,\,t}^{s_i} - \widehat{vtec}_{r,\,t}^{s_i})^2} \right) \tag{4.7}$$

The $\overline{mae}_r$ error ranges between 0.71 and 0.91 for our proposed CONV1D method, which is better than $1\,TECU$. As regards the recurrent LSTM method, the $\overline{mae}_r$ error is 0.93 to $1.24\,TECU$, while for the autoregressive methods, the respective mean absolute error is greater than $1.5\,TECU$. The $\overline{rmse}_r$ values for CONV1D are between 0.88 and 1.09, while in the other methods used for comparison, the results shown values greater than $1\,TECU$. As derived from the results, CONV1D extracts the temporal information and models successfully the TEC sequential problem.

Figure 4.2 shows the estimated TEC timeseries for each individual PRN for bor1 station for a single day of observations. Satellites with PRNs G01, G08, G11, G17 and G27 have the best performance, while G16 and G23 show the worst one. Overall, as noticed, the predicted values are close to the estimated ones and the CONV1D network can successfully model TEC values per PRN.

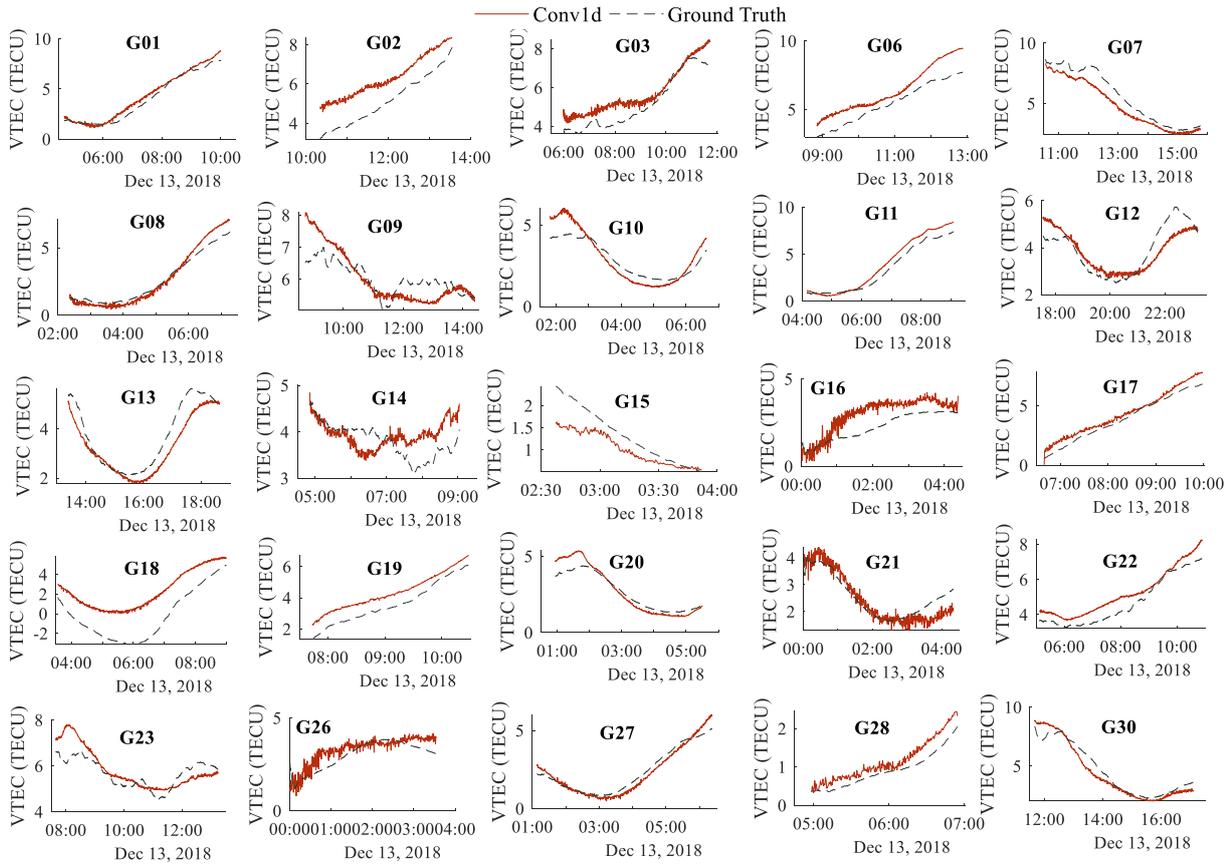

Fig. 4.2 Conv1d versus ground truth per satellite PRN for bor1 station.

Figure 4.3 shows the MAE results for two different test sets. Here, we notice that the proposed results with the CONV1D method achieve slightly better performance than the traditional recurrent networks used for timeseries modelling.

Figure 4.4 illustrates the mean VTEC values at every site, as obtained from NeQuick, IRI2001 and GIM TEC estimates compared to the ground truth values (GRD) and CONV1D TEC predictions, during the day. The maximum VTEC values appeared the time duration between 8:00 and 12:00 A.M. In most cases, the CONV1D values are similar to those of GPS TEC derived values. This is as expected, because the CONV1D model has been trained using as training set these values. However, it is noted that our model underestimates VTEC values, showing lower values than those of GPS TEC. GIM-aided TEC values are also close to CONV1D derived TEC values, which is also expected as the processing PPP strategy uses GIM TEC values for input (ionosphere constrained model). Finally, NeQuick and IRI2001 values show wider variability during the day with higher maximum and lower minimum values, compared those of GIM and GPS TEC methods.

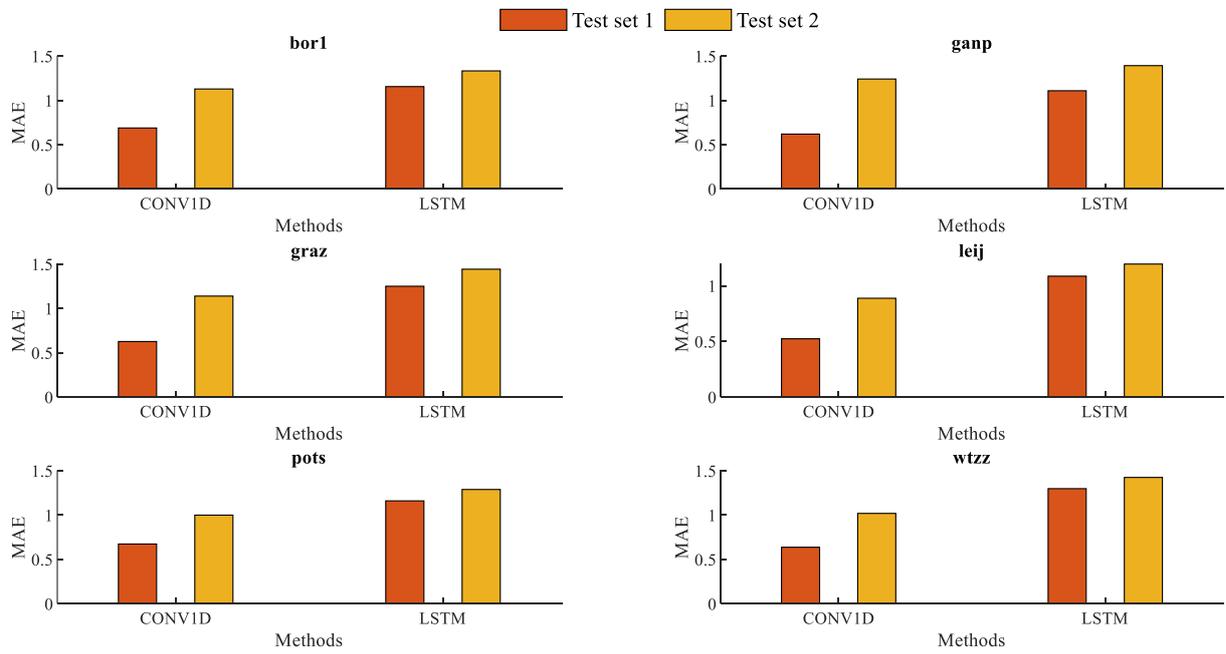

Fig. 4.3 Conv1d versus ground truth per satellite PRN.

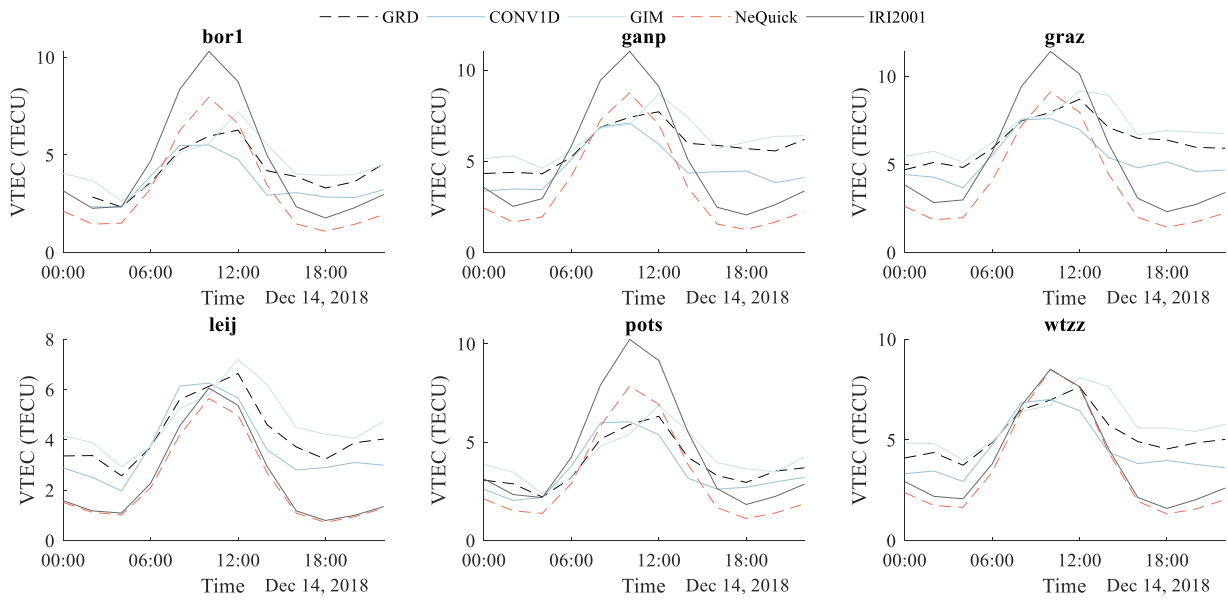

Fig. 4.4 Conv1d versus ground truth per satellite PRN.

## 4.4   CONCLUSION

With regard to the adoption of a suitable CNN architecture, a non-linear autoregressive network based on convolutions for TEC modelling is proposed. Temporal convolutional neural networks can successfully model sequential data. In contrast to the recurrent neural networks that traditionally have been used for timeseries modelling, our CONV1D method except for higher accuracy, present lower complexity since the only operation with a significant cost is a sequence of 1D convolutions which are weighted sums of two 1D arrays. The aim of the proposed model, is to define the characteristics of the GNSS ionospheric delays, especially at mid-latitudes, using multiple frequency observations and data from multiple PRNs. Here, we aim at models that apply accurate ionosphere TEC corrections, to increase the accuracy and convergence time for single station and single frequency techniques. propose an efficient model for constructing dynamic changing, regional TEC models. The experimental results indicate that mean absolute error ranges between 0.71 and 0.91 for the CONV1D method, which means that the error is better than $1\,TECU$.

# Chapter 5

# The implementation of a hybrid Deep learning scheme (CNN+GRU) for TEC modeling

Here we introduce a new *a spatio-temporal* hybrid deep learning paradigm for TEC modelling. The proposed method combines two deep learning structures; *a Convolutional Neural Network* and a *Gated Recurrent Unit* [89]. These two structures are combined together, forming a common trainable model.

TEC is characterized by high complexity and it is *space− and time−* varying. In order to capture the spatio-temporal behaviour of TEC, in our proposed model, measurements of every satellite visible from the observing location (ground station) are taken into account. This means that the model's input data form a spatial-temporal 3D tensor. The tensor consists of several timeseries data corresponding to different single satellite visible from the station over a time window period. Earth's ionosphere shows marked variations with latitude, longitude, universal time, season, solar and geomagnetic activity. Thus, our proposed model feeds all these variables as inputs to the model and relates them under a non-linear relationship.

The CNN captures the spatial variability of the TEC values assigning different learnable weights to the following GRU structure, taking into account all the visible timeseries data from a ground station. The main operational unit of a CNN architecture is the convolutional kernels, units able to model with high efficiency spatial signal properties under complex non-linear relationships [36]. In other words, the purpose of the CNN is to appropriately adjust the model weights to better capture the particularities of each station.

On the other hand, the GRU units model the temporal variability of the ionosphere variability. GRUs are similar deep learning structures with LSTM, requiring, however, fewer parameters. Therefore, they retain the temporal dependent capabilities of the LSTM, but, they advance in terms of complexity (and therefore convergence) as well as the need of smaller training (annotated) data sets. For this reason, GRUs units have been selected for modeling the temporal ionospheric variability. In other words, our model leverages the capability of the CNNs to optimally approximate the spatial particularities of each station, based on a set of weights, along with GRUs structures able to learn the temporal dependencies of the TEC values.

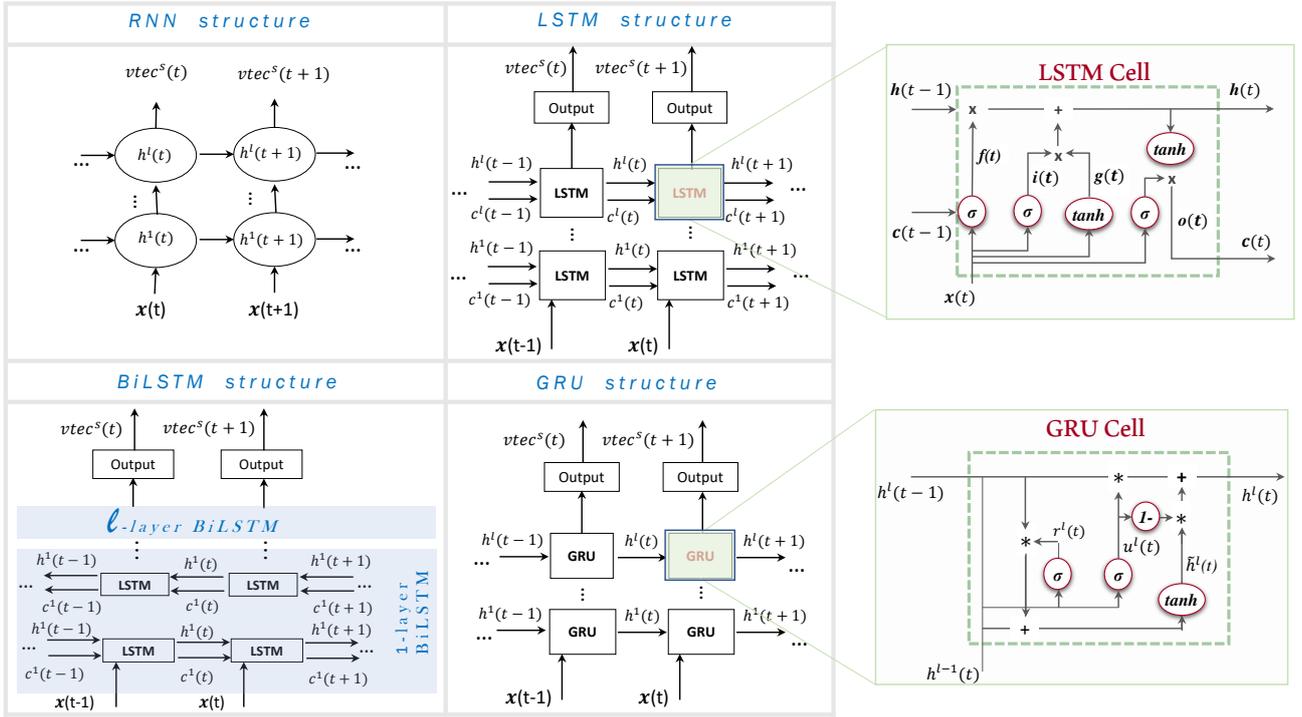

Fig. 5.1 Stacked RNN-based structures for TEC prediction.

The chapter is structured as follows: Section 5.1 describes the proposed spatio-temporal deep learning TEC model. Section 5.2 describes the proposed model architecture that combines convolutional and recurrent layers together in an optimal structure. In Section 5.3, an extensive experimental evaluation of the discussed methods is provided, while Section 5.4 closes the chapter with a summary of findings.

## 5.1 Sequence-to-Sequence Spatio-temporal AI for TEC modeling

As we have previously stated, a spatio-temporal deep learning model is used, here, to model the ionospheric variability. The proposed deep learning model consists of a convolutional layer (see Section 5.1.1) and stacked GRU recurrent layers (see Section 5.1.2).

### 5.1.1 Spatial Variability Modeling: The Convolutional Neural Network Layer

The purpose of the convolutional layer is to encode the spatial variability of the input signals by assigning different weights to the model depending on the station latitude. In other words, the output of the convolutional layer is to transform the input signals into a vector representation more suitable for TEC modeling. The weights of the convolutional layer are learnable during the training phase and each input contributes in a different way to the model, for different ground stations. Let us denote as $\mathcal{I}_{input}$ the input sequence of data which are feeding to the convolutional layer. Then, the CNN transforms these inputs $\mathcal{I}_{input}$ to an encoded vector $f_{\mathcal{M}}$ more suitable for TEC modeling.

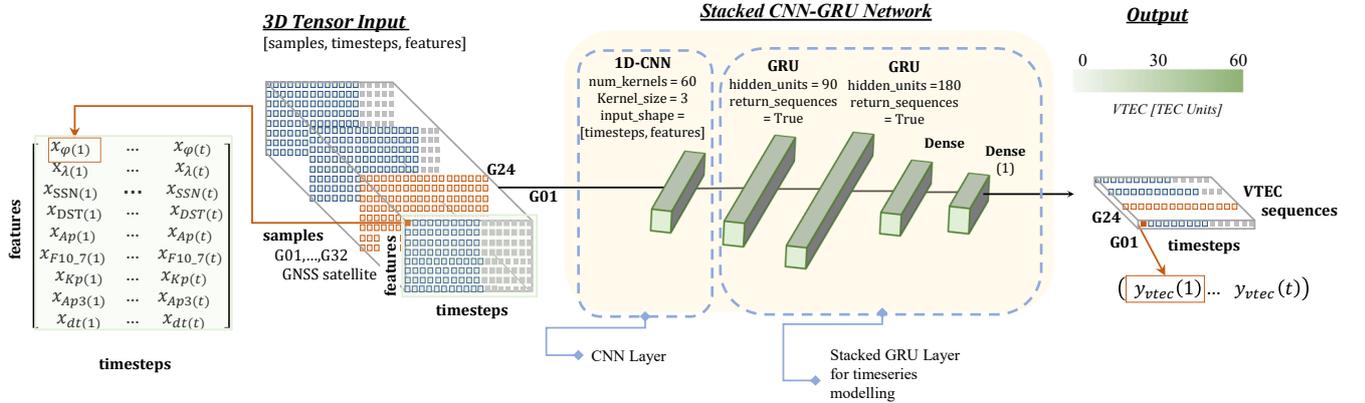

Fig. 5.2 The proposed model CNN-GRU architecture, consisting of two parts; the Convolutional Neural Network (CNN) for spatial variability modeling and the Gated Recurrent Unit (GRU) for temporal variability modeling.

Table 5.1 Static PPP experimental setup.

| Options | Settings |
|---|---|
| Constellation | GPS |
| Positioning mode | static PPP |
| Frequencies | L1, L2 |
| Sampling rate | 1 min |
| Elevation mask | 7° |
| Differential code bias (DCB) | MGEX DCB products for PPP |
| Receiver and Satellite antenna | corrected with igs14.atx |
| Satellite orbits and clock offsets | Wuhan University products |
| Phase wind-up | corrected |
| Sagnac effect, relativistic effect | corrected with IGS absolute |
| Station reference coordinates | IGS SINEX solutions |

Table 5.2 The selected sites from the IGS network.

| Site | Latitude (°) | Longitude (°) | Height (m) | Country |
|---|---|---|---|---|
| graz | 47.067 | 15.493 | 538.3 | Austria |
| iqal | 63.755 | -68.510 | 91.7 | Canada |
| mal2 | -2.996 | 40.194 | 0.0 | Uganda |
| qaq1 | 60.715 | -46.047 | 110.4 | Greenland |
| ramo | 30.597 | 34.763 | 893.1 | Israel |
| tixi | 71.634 | 128.866 | 46.9 | Russia |

$$f_M \sim Conv_{\mathcal{M}}\left(\mathcal{I}_{input}\right)$$
$$\text{with}$$
$$\mathcal{I}_{input}(t) = [x_\phi \ x_\lambda \ x_{ssn} \ x_{Dst} \ x_{Ap} \ x_{F10_7} \ x_{Kp} \ x_{dt}]^T$$
(5.1)

Then, the encoded vector $f_M$ feeds the GRU structure. The convolutional layer architecture and its position with respect to the whole deep learning model is shown in Fig. 5.2.

### 5.1.2    Temporal Variability Modeling: The Gated Recurrent Unit Layer

As we have previously stated, GRU are simpler forms of LSTM models having less gates than the LSTM memory cell [90]. In addition, the LSTM networks are deep learning extensions of RNNs model, better modeling the non-linear attributes of a time series signal [43]. Therefore, before describing the GRU unit used for modeling the temporal variability of TEC, we discuss the structure of the Recurrent Neural Networks (RNNs) and their stacked version and Long-Short Term Memory (LSTM) networks

A stacked recurrent model is an extension to the traditional one consisting of several recurrent layers one stacked over the other [91]. This is presented in Fig. 5.1. A stacked model has two main types of operation; *the in-depth* and *the temporal operation*. The in-depth operation implies that the response of one layer is propagated as input to the next layer. Instead, the temporal operation assumes that inputs at previous time instances trigger the current unit. The stacked approach adopted here, is applied for all recurrent structures, that is, the RNN, the LSTM and its bi-directional mode, and the GRU.

### Stacked Recurrent Neural Networks

Recurrent models are powerful tools for timeseries modeling. The main operational unit of an RNN is an artificial neuron, approximating a non-linear operation of an inner product of the network weights (parameters) and output responses of other neurons or model input vectors. The difference of an RNN with a traditional neural network model is that, in an RNN, each neuron is also triggered from the response of other neurons at previous time instances, allowing modeling of temporal dependencies. These neurons are also called hidden states since they are located between the input vector and the output of the model. In particular, the response of an artificial neuron is given by

$$h^l(t) = tanh(\mathbf{W}_l^T h^l(t-1) + \mathbf{U}_l^T h^{l-1}(t)), \ l > 1$$
(5.2)

where $tanh(\cdot)$ is the hyperbolic tangent function, referring to the non-linear operational unit of a single neuron of the RNN. Variables $\mathbf{W}$ and $\mathbf{U}$ are the learnable weight parameters of the model. The $h^l(t)$ is the response of a neuron at $l$-th level of the network at a time instance $t$. It should be mentioned the $h^0(t) \equiv f_M$ coincides with the output response of the convolutional layer, that is the vector $f_M$ of Eq. (9.9), instead of the previous hidden state $h^{l-1}(t)$.

Once the top-level hidden state is computed, the estimate of the output $\widehat{vtec}^s$ is obtained using (see Fig. 5.1)

Table 5.3 Performance metrics (mae, mse, min and max) for VTEC modeling for six selected stations (graz, iqal, mal2, qaq1, ramo, tixi) among the RNN -based methods as well as two other commonly used methods (ARMA, AR) for 9 October 2018. mae and mse metrics are the average values of the individual PRNs metrics in every station, whereas min and max are the minimum and maximum mae values observed for the individual PRNs. The best values are presented in bold.

| | graz | | | | iqal | | | | mal2 | | | |
|---|---|---|---|---|---|---|---|---|---|---|---|---|
| | **mae** | **mse** | **min** | **max** | **mae** | **mse** | **min** | **max** | **mae** | **mse** | **min** | **max** |
| proposed | **0.714** | **0.903** | 0.290 | 1.961 | 0.971 | **1.591** | **0.390** | **1.722** | 1.814 | 5.878 | **0.472** | **3.301** |
| BILSTM [42] | 0.824 | 1.083 | **0.276** | 1.968 | 1.096 | 2.118 | 0.457 | 2.998 | 2.175 | 7.814 | 0.487 | 3.986 |
| LSTM *[68], [43]* | 0.836 | 1.304 | 0.421 | **1.745** | **0.940** | 1.594 | 0.523 | 2.411 | 2.166 | 8.874 | 0.598 | 4.633 |
| RNN *[69]* | 0.965 | 1.531 | 0.295 | 1.772 | 1.023 | 1.687 | 0.446 | 1.913 | **1.758** | **5.867** | 0.702 | 3.587 |
| ARMA *[92]* | 2.660 | 13.278 | 1.482 | 3.917 | 2.419 | 9.773 | 1.269 | 3.279 | 6.767 | 70.930 | 3.513 | 14.071 |
| AR *[92]* | 2.772 | 14.358 | 1.484 | 4.391 | 2.501 | 10.733 | 1.626 | 3.254 | 4.713 | 26.280 | 3.241 | 12.592 |

| | qaq1 | | | | ramo | | | | tixi | | | |
|---|---|---|---|---|---|---|---|---|---|---|---|---|
| | **mae** | **mse** | **min** | **max** | **mae** | **mse** | **min** | **max** | **mae** | **mse** | **min** | **max** |
| proposed | **0.779** | **1.072** | **0.221** | 1.979 | **1.093** | **2.066** | **0.339** | **2.506** | **0.891** | **1.335** | 0.442 | **1.605** |
| BILSTM [42] | 1.123 | 1.839 | 0.317 | 2.200 | 1.393 | 3.317 | 0.358 | 3.889 | 0.912 | 1.431 | 0.494 | 2.049 |
| LSTM *[68], [43]* | 1.022 | 1.775 | 0.225 | 3.441 | 1.247 | 2.807 | 0.392 | 3.256 | 0.913 | 1.423 | 0.480 | 1.904 |
| RNN *[69]* | 0.889 | 1.213 | 0.235 | **1.496** | 1.461 | 3.414 | 0.541 | 4.266 | 1.141 | 2.043 | **0.423** | 2.673 |
| ARMA *[92]* | 2.942 | 13.571 | 2.078 | 4.070 | 5.914 | 56.487 | 4.032 | 7.588 | 1.919 | 7.242 | 1.392 | 2.877 |
| AR *[92]* | 2.995 | 14.612 | 1.883 | 4.100 | 5.671 | 56.737 | 3.195 | 8.384 | 2.150 | 8.641 | 1.021 | 2.637 |

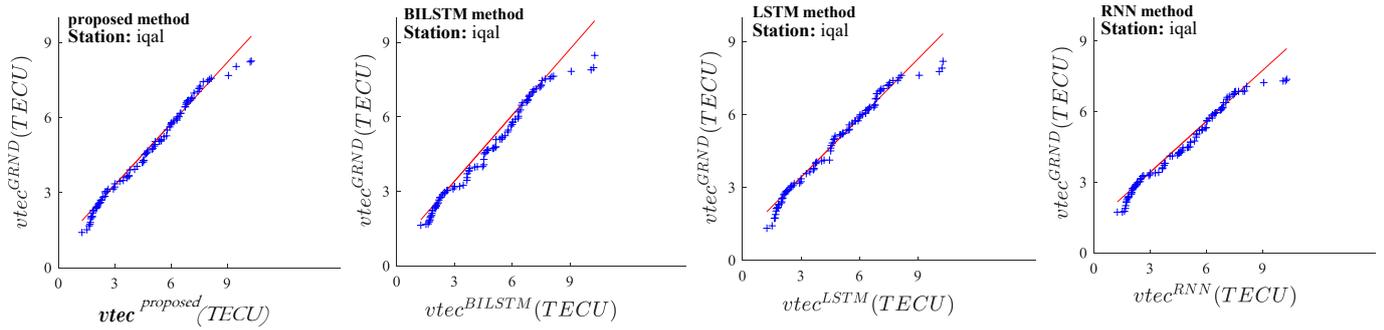

Fig. 5.3 Q-Q plots for the 'iqal' station, between the ground truth and the other deep learning techniques used for comparison.

$$\widehat{vtec}^{\,s}(t) = tanh(\mathbf{V}^T h^l(t)) \tag{5.3}$$

where $\mathbf{V}$ is a matrix corresponding to the the output learnable parameters (weights) of the model.

**The LSTM Structure**

LSTM is a special kind of the traditional RNN structure, where each node in the hidden layer is replaced by a more complex structure, called memory cell, instead of RNN's single neuron [40]. The core structure of a single memory cell is presented in Fig. 1. The memory cell contains three different components: (i) the forget gate $f(t)$, (ii) the input gate $i(t)$, (iii) the cell candidate $g(t)$ and (iii) the output gate $o(t)$. For each component, a non-linear relation to the inner product between the input vectors and respective weights is applied during the training process. In some of the components the sigmoid function $\sigma(\cdot)$ is applied, while in others the hyperbolic tangent function $\tanh(\cdot)$ is used. Forget gate $f(t)$ keeps unnecessary information out of memory cell, thus separating the worth-remembering information from the useless one [81]. Input gate $i(t)$ regulates whether the information is relevant enough to be applied in future steps for the accurate estimation of TEC values. Cell candidate $g(t)$ activates appropriately the respective state (true or false output from the tanh activation). Output gate $o(t)$ decides if the response of the current memory cell is "significant enough" to contribute to the next cell.

Regarding stacked LSTMs, the additional LSTM layers can recombine the learned representation from prior layers and create new representations at high levels of abstraction. Stacked LSTMs or Deep LSTMs were introduced by [76] in their application of LSTMs to speech recognition, beating a benchmark on a challenging standard problem.

$$\begin{bmatrix} i^l(t) \\ f^l(t) \\ o^l(t) \\ g^l(t) \end{bmatrix} = \begin{bmatrix} \sigma \\ \sigma \\ \sigma \\ \tanh \end{bmatrix} W_l \begin{bmatrix} h^l(t-1) \\ h^{l-1}(t) \end{bmatrix} + \begin{bmatrix} b_i^l \\ b_f^l \\ b_o^l \\ b_g^l \end{bmatrix} \tag{5.4}$$

when $l=1$, the state is computed using the encoded vector $f_M$ (see Section 5.1.1) instead of $h^{l-1}(t)$.

The cell state at time step t is given by

$$c^l(t) = f^l(t) \odot c^l(t-1) + i^l(t) \odot g^l(t) \tag{5.5}$$

where $\odot$ denotes the Hadamard product (element-wise multiplication of vectors).

The hidden state at time step $t$ is given by

$$h^l(t) = o^l(t) \odot \sigma(c^l(t)) \tag{5.6}$$

where $\sigma$ denotes the state activation function.

**The Bidirectional LSTM**

The stacked bidirectional LSTM has a similar structure with the LSTM model with the difference that it allows bi-directional data processing. Therefore, the operational units of a bi-directional LSTM model is defined by two parts; the *forward* and the *backward* part. As far as the forward part is concerned we have the following equations

$$\overrightarrow{c}^l(t) = \overrightarrow{f}^l(t) \odot \overrightarrow{c}^l(t+1) + \overrightarrow{i}^l(t) \odot \overrightarrow{g}^l(t) \tag{5.7}$$

$$\overrightarrow{h}^l(t) = \overrightarrow{o}^l(t) \odot \sigma(\overrightarrow{c}^l(t)) \tag{5.8}$$

and backward:

$$\overleftarrow{c}^l(t) = \overleftarrow{f}^l(t) \odot \overleftarrow{c}^l(t-1) + \overleftarrow{i}^l(t) \odot \overleftarrow{g}^l(t) \tag{5.9}$$

$$\overleftarrow{h}^l(t) = \overleftarrow{o}^l(t) \odot \sigma(\overleftarrow{c}^l(t)) \tag{5.10}$$

**The stacked GRU architecture**

The GRU is simpler form of the LSTM unit, having two control gates; the *reset* gate $r^l(t)$ and the *update* gate $u^l(t)$ (see Fig. 1). The reset gate $r^l(t)$ is responsible for determining how much of information to forget. The update gate $u^l(t)$ is responsible for determining the worth-remembering information of the previous states that should be forwarded to the next state. Therefore, the gates $r^l(t)$ and $u^l(t)$ are related with the hidden states $h^l(t)$ and $h^l(t-1)$ as follows

$$\begin{bmatrix} u^l(t) \\ r^l(t) \end{bmatrix} = \begin{bmatrix} \sigma \\ \sigma \end{bmatrix} \mathbf{W}_l \begin{bmatrix} h^l(t-1) \\ h^{l-1}(t) \end{bmatrix} + \begin{bmatrix} b_u^l \\ b_r^l \end{bmatrix} \tag{5.11}$$

In Eq. (9.6) $\sigma$ is the sigmoid function and $b_u^1$, $b_r^1$ are the respective biases of each component for the GRU cell. Variables $\mathbf{W}$ and $\mathbf{U}$ are the transition matrices of the $l$-th GRU.

In stacked GRUs configuration a recursive approach is considered as regards the operation of each GRU cell. A new memory state, denoted as $\tilde{h}^l(t)$, acts as the consolidation of the hidden state of the previous layer $h^{l-1}(t)$ and the previous hidden state $h^l(t-1)$ of the current layer. The consolidated hidden state $\tilde{h}^l(t)$ is given by

$$\tilde{h}^l(t) = \tanh\left(r^l(t)\mathbf{U}h^l(t-1) + \mathbf{W}h^{l-1}(t)\right) \tag{5.12}$$

In Eq. (9.7) function $tanh(\cdot)$ refers to the hyperbolic tangent relationship. Eq. (9.7) means that the consolidated state is related with the output of the hidden state $h^l(t-1)$ at the time instance $t-1$ and the output of the previous hidden layer $h^{l-1}(t)$ at the time instance $t$. Using the values of the consolidated state $\tilde{h}^l(t)$ and the values of the update gate $u^l(t)$, the value of the hidden state of the $l$-th GRU element is estimated

$$h^l(t) = \left(1 - u^l(t)\right)\tilde{h}^l(t) + u^l(t)h^l(t-1) \tag{5.13}$$

In more detail, the GRU related above mentioned operations are illustrated in Fig. 1.

## 5.2 The Proposed CNN-GRU Architecture for VTEC Modeling

### 5.2.1 Implementation Details

Fig.2 illustrates the CNN-GRU model structure. The network configuration setup consists of: (i) the input layer, (ii) a one-dimensional convolutional layer, (iii) two stacked recurrent GRU layers and (iii) two dense layers (Fig. 2). The GRU structure accepts a 3D tensor input. The number of kernels of the convolutional layer is 60 with size 5. The first GRU layer consists of 90 filters with size 1x5, while the second layer consists of 180 filters with the same size. The output is the sequence of VTEC values.

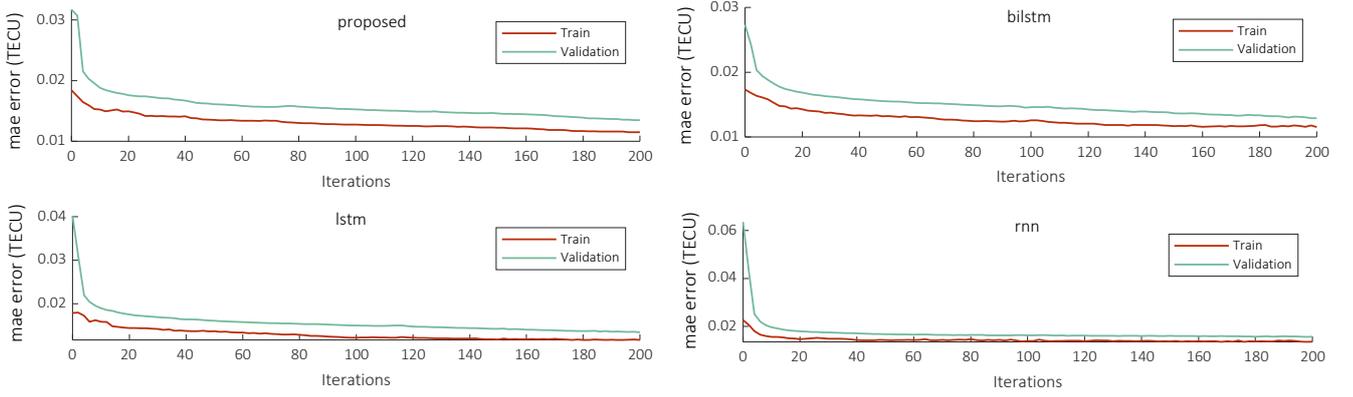

Fig. 5.4 Loss curve per RNN-based method (GRU, BILSTM, LSTM and RNN) for training and validation set for ramo station.

## 5.2.2   Evaluation Metrics

Here, we present the metrics used for model evaluation and comparisons with other liner and non-linear approximators.

For each satellite $s_i$, visible in ground station, a sequence of VTEC values is produced using the proposed CNN-GRU deep learning model. Then, we compute the absolute difference between the ground truth $vtec^{s_i}(t)$ value, as obtained from the GAMP software, and the estimated from our model $\widehat{vtec}_r^{s_i}(t)$, that is $dvtec^{s_i}(t) = |vtec^{s_i}(t) - \widehat{vtec}^{s_i}(t)|$, for a time instance $t$. Based on the values of $dvtec^{s_i}(t)$ the following metrics are considered:

$$mae = \frac{1}{S} \sum_{s_i=1}^{S} mae^{s_i} = \frac{1}{S} \sum_{s_i=1}^{S} \left( \frac{\sum_{t=1}^{T} dvtec^{s_i}(t)}{T} \right) \tag{5.14}$$

$$mse = \frac{1}{S} \sum_{s_i=1}^{S} mse^{s_i} = \frac{1}{S} \sum_{s_i=1}^{S} \left( \frac{\sum_{t=1}^{T} [dvtec^{s_i}(t)]^2}{T-1} \right) \tag{5.15}$$

$$min = \frac{1}{S} \sum_{s_i=1}^{S} \min_{\forall t \in T} dvtec^{s_i}(t) \tag{5.16}$$

$$max = \frac{1}{S} \sum_{s_i=1}^{S} \max_{\forall t \in T} dvtec^{s_i}(t) \tag{5.17}$$

In the aforementioned equations, i) $mae$ refers to mean absolute error for all satellite $s_i$ per ground station, ii) $mse$ to the respective mean square error and iii) $min$ ($max$) to the average minimum (maximum) of $dvtec^{s_i}(t)$ for all visible satellites. Variable $T$ denotes the time period over which the evaluation takes place.

Another evaluation metric used here, is the percentiles values of (50th, 68th, 95th) of the $dvtec(t)$ error distribution per ground station. The error $dvtec(t) = |vtec(t) - \widehat{vtec}(t)|$ is defined as the absolute different between the estimated VTEC value and the ground truth for a station. An 95th percentile quantity means the value that 95% of the VTEC errors are contained, after having sorted all errors in ascending order. Reflecting the error by percentiles is better than by simple mean absolute error.

Table 5.4 The 50, 65 and 98% percentile scores, for the evaluation of the different RNN-based networks with the proposed one.

| | percentile error (TECU) | | | | | |
|---|---|---|---|---|---|---|
| | graz | | | iqal | | |
| | 50% | 65% | 98% | 50% | 65% | 98% |
| proposed | 0.90 | 1.34 | 2.70 | 0.45 | 0.73 | 1.81 |
| BILSTM | 1.03 | 1.55 | 2.69 | 0.48 | 0.74 | 1.65 |
| LSTM | 1.12 | 1.45 | 2.91 | 0.44 | 0.73 | 1.82 |
| RNN | 1.00 | 1.55 | 2.86 | 0.59 | 0.95 | 2.61 |
| | mal2 | | | qaq1 | | |
| | 50% | 65% | 98% | 50% | 65% | 98% |
| proposed | 1.62 | 2.48 | 3.88 | 0.87 | 1.50 | 2.42 |
| BILSTM | 1.54 | 2.25 | 4.66 | 1.17 | 1.56 | 2.35 |
| LSTM | 1.93 | 2.52 | 4.69 | 1.00 | 1.43 | 2.50 |
| RNN | 1.21 | 1.66 | 4.38 | 0.85 | 1.28 | 3.09 |
| | ramo | | | tixi | | |
| | 50% | 65% | 98% | 50% | 65% | 98% |
| proposed | 0.87 | 1.26 | 3.04 | 0.67 | 0.97 | 1.88 |
| BILSTM | 1.09 | 1.67 | 3.56 | 0.77 | 1.13 | 2.47 |
| LSTM | 0.91 | 1.50 | 3.57 | 0.68 | 1.18 | 2.53 |
| RNN | 0.88 | 1.45 | 3.34 | 0.86 | 1.28 | 2.62 |

Table 5.5 Performance metric mae values for the training and validation sets and for the six stations among the RNN -based methods (GRU, BILSTM, LSTM and RNN).

| | graz | | iqal | | mal2 | | qaq1 | | ramo | | tixi | |
|---|---|---|---|---|---|---|---|---|---|---|---|---|
| | mae | | mae | | mae | | mae | | mae | | mae | |
| | train | val | train | val. | train | val. | train | val | train | val. | train | val. |
| proposed | 0.0167 | 0.0177 | 0.0206 | 0.0213 | 0.0164 | 0.0166 | 0.0239 | 0.0250 | 0.0115 | 0.0135 | 0.0298 | 0.0240 |
| BILSTM | 0.0168 | 0.0190 | 0.0213 | 0.0240 | 0.0185 | 0.0206 | 0.0252 | 0.0270 | 0.0115 | 0.0129 | 0.0299 | 0.0317 |
| LSTM | 0.0175 | 0.0189 | 0.0217 | 0.0243 | 0.0191 | 0.0201 | 0.0254 | 0.0268 | 0.0118 | 0.0135 | 0.0301 | 0.0303 |
| RNN | 0.0184 | 0.0214 | 0.0235 | 0.0295 | 0.0204 | 0.0241 | 0.0258 | 0.0281 | 0.0137 | 0.0157 | 0.0319 | 0.0364 |

## 5.3    Experimental Evaluation

The experimental set up covers various aspects of the ionosphere phenomenon, in order to accurately evaluate the performance of our proposed model. Thus, in our experimental evaluation, we have included stations installed in various latitudes and longitudes at worldwide level (see Table 5.2). Also, we have included data from different years ranging from 2014 to 2018 to create robust training, validation and test sets. We have chosen different years under different solar and geomagnetic activity and also, we have tested different months, in order to evaluate our algorithm under different weather and seasonal conditions.

We highlight that our proposed model is not a "pure" GRU model but a hybrid model comprising of convolutional and stacked recurrent GRU layers; *the proposed spatio-temporal deep learning paradigm*. We experimentally validate the superiority of the proposed CNN-GRU model for TEC modeling in comparison to other deep learning networks (LSTM [68], [43], BILSTM [42] and RNN [69] models), as well as state of the art techniques for TEC modeling (AR [92] and ARMA [92] models). Also the proposed method is compared with the conventional GIM and IRI2016 global ionosphere models.

### 5.3.1    Data Pre-processing and Experiment Setup

This section describes the pre-processing approach used to extract the recurrent model inputs and outputs, based on observables from the global IGS (International GNSS Service) network of permanent GNSS stations. The necessary observation, navigation, precise orbit and clock, IGS ANTEX (igs14.atx), IGS SINEX, ocean tide loading coefficients, DCBs are fed into the GAMP software for static PPP processing (see Table 5.1). As described in Section 2.3, the GNSS observations are pre-processed to estimate the VTEC values. GNSS observables for the years 2014 to 2018 were processed with 1 min data granularity, for the selected ground stations (see Table 5.2).

As far as the deep learning models are concerned, they have been trained and deployed using the Python Tensorflow and Keras libraries. The proposed CNN-GRU as well as the compared deep learning models have been trained using the adaptive moment estimation optimization algorithm (ADAM) [93] with a learning rate of $10^{-4}$. The model weights are updated using a mini-batch size of 28 samples at each training iteration. We set to 200 the maximum number of epochs for training, that is, the maximum number of training cycles of the network. In our experiments, we use a dataset of the years 2014 to 2018 (a five year dataset). This dataset is divided into three subsets; *the Training, the Validation and the Test Set*. The training set is used for the estimation of the model parameters (weights) during the learning process. The validation set assesses the performance of the model during the learning process on a set different than the one used for training to avoid overfitting. Finally, the test set evaluates the model on data that have not used during training. In our experiments, the training set consists of 70% of the total available data, while both the validation and the test sets each consists of the remaining 15% of the total data.

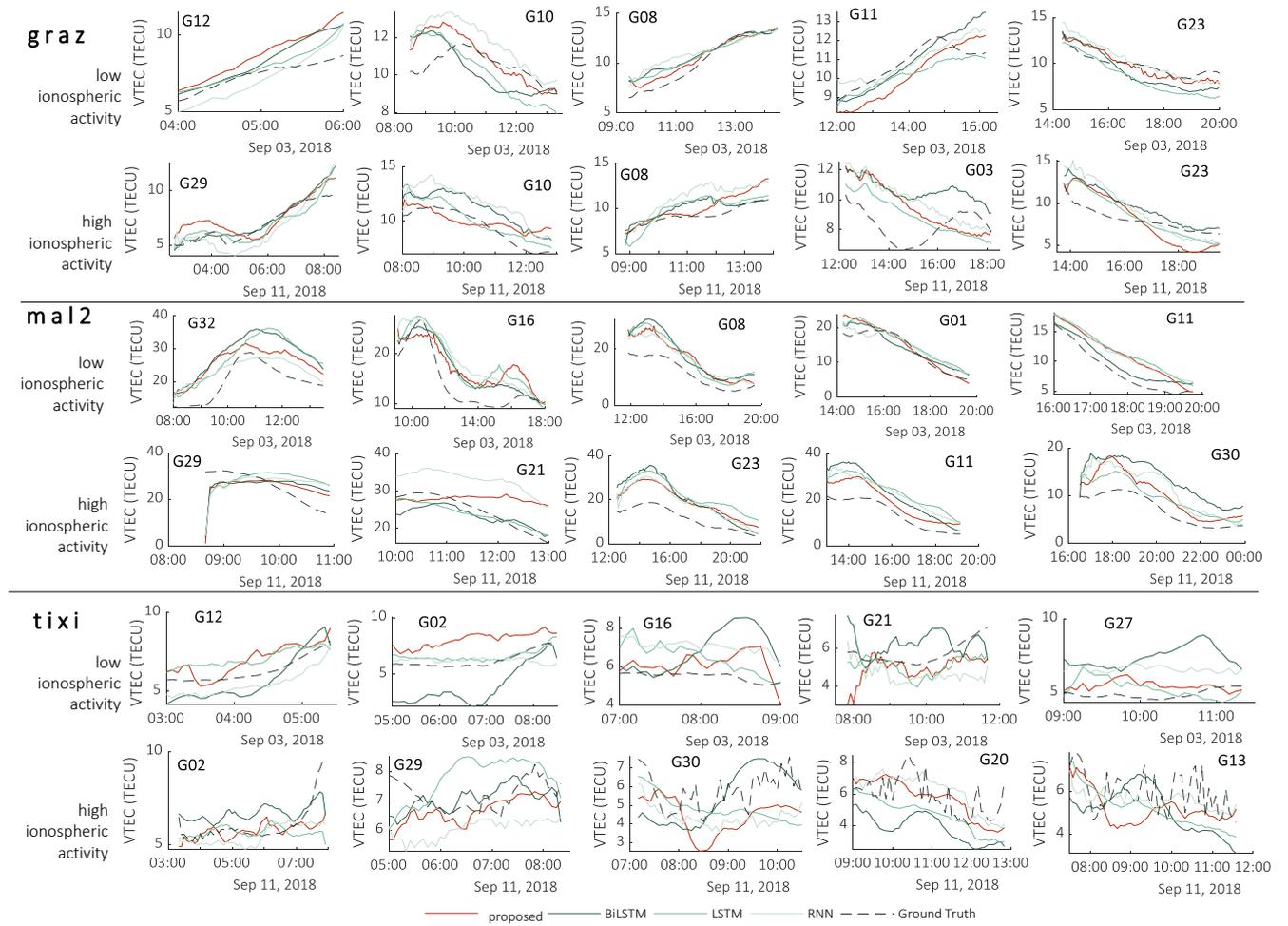

Fig. 5.5 Vertical TEC for individual PRNs (unique pseudo-random noise that each GPS satellite transmits) between the GPS ground truth values and GRU, BILSTM, LSTM and RNN methods in various hours two different days: a day of low ionospheric activity (3/9) and a day of high ionospheric activity (11/9). The figure illustrates VTEC values estimated per satellite from a low-latitude station('mal2') and a mid-latitude station ('graz') and a high-latitude station ('tixi').

Both, training and validation sets cover the period between the years 2014 to mid-2018. Data from the second half of 2018 are used as test data.

The AR and ARMA models [92] are implemented in Python using the statsmodels library. The order of the autoregressive part (AR) for both AR and ARMA is selected to be 60, while the order of the moving average (MA) part equals 5 in case of the ARMA model.

### 5.3.2   Performance Evaluation and Comparison

In this section, we compare the various learning architectures to model the temporal dynamics of the ionospheric total electron content (TEC). In particular, we implement and evaluate the traditional RNN, the unidirectional LSTM, the bidirectional LSTM (BILSTM) and the proposed CNN-GRU deep learning model (see Section 5.1). We also compare our model performance against linear regressors of AR and ARMA. The comparison is carried out in terms of performance accuracy and computational efficiency.

Table 5.6 Comparison among different RNN -based methods architecture (GRU, BILSTM, LSTM and RNN) characteristics and parameters (trainable parameters and processing time).

|  | Methods | | | |
|---|---|---|---|---|
|  | proposed | BILSTM | LSTM | RNN |
| Trainable Parameters (num.) | 112493 | 379485 | 149453 | 38573 |
| Epochs | 200 | 200 | 200 | 200 |
| Training Time (min) | 67 | 280 | 104 | 23 |

Table 5.3 shows the performance evaluation metrics, as described in Section 5.2.2, of (i) mean absolute error (mae) in $TEC$ units ($TECU$), (ii) mean squared error (mse) in $TECU^2$ and (iii) minimum (min) and maximum (max) TEC values again in $TECU$. The results in Table 5.3 have been calculated for the six stations (see Table 5.2) and a randomly selected testing period of a whole day (the 9th of October 2018). In this table, we have also depicted the performance of the four compared non-linear architectures (i.e., the proposed CNN-GRU, the BILSTM, the LSTM and the RNN) along with the traditional AR and ARMA models. As is observed, the proposed CNN-GRU model has the best performance, in most cases, with a mae value between the interval $[0.7 - 1.8]$. On the other hand, the worst performance is for the AR and ARMA linear models with a mae value between $[2.2 - 5.7]$. Regarding the compared deep learning models, the mae values are as follows: RNN lies in $[0.9 - 1.8]$, LSTM in $[0.8 - 2.2]$, BILSTM in $[0.8 - 2.2]$. The same conclusions are drawn for the other evaluation metrics.

Table 5.5 illustrates the average mae error in TECU over the six stations for the training and validation sets. The resuts have been obtained for all the non-linear models. In most cases, the proposed CNN-GRU model has better performance than the other methods. In particular, from Table 5.5, CNN-GRU achieves an average mae of 1.98% for training set and 1.97% for the validation set. The other non-linear models have an average mae of: BILSTM 2.05% (training set) and 2.25% (validation set), LSTM 2.09% (training set) and 2.23% (validation set), RNN 2.23% (training set) and 2.59% (validation set). This means that the proposed CNN-GRU model has an improvement of 11.07% and 23.9% with respect to the conventional RNN architecture in the training and validation set respectively. In addition, the proposed CNN-GRU model achieves an improvement of 3,69% and 12.65% with respect to the BILSTM model for the training and validation set. It should be mentioned that the improvement in the validation set is more significant than the one of the training set, since the latter refers to data that they have not be used during training.

Fig. 5.3 illustrates the Quantile-Quantile (Q-Q) plots for the 'iqal' station, between the ground truth and the evaluated deep learning architectures. In case of a perfect performance, the points of the Q-Q plot will lie on the line $y = x$ (the red line in Fig. 5.3). The closer the data points are to the red line, the better is the model accuracy. Thus, the proposed CNN-GRU method has better performance compared the other ones, since the blue points are closer to the red line. We also observe that higher $vtec$ values are more challenging to be accurately modelled. However, again, the proposed CNN-GRU model achieves better performance for these high $vtec$ values compared to the other models.

Table 5.4 illustrates the 50%, 65% and 98% percentile scores for the proposed and the compared non-linear models. The proposed model of the 'graz', 'iqal' and 'tixi' ground stations have the best performance, whereas the model of the 'mal2' ground station present the worst one. Again, the proposed CNN-GRU model has better performance against the compared architectures. Based on the values of Table 5.4, we have concluded that the proposed CNN-GRU model, in case of 98% percentile, achieves an improvement of $16,77\%$ with respect to the RNN architecture and of $12.71\%$ and $9.49\%$ with respect to the LSTM and BILSTM model respectively. This means that the proposed CNN-GRU modeling error are not as large as in case of the compared models.

Processing time in minutes and the number of the required trainable parameters per method are listed in Table 5.6. As is observed, the most efficient is the RNN architecture which requires 23min (for 200 training epochs), while the heaviest is the BILSTM model needing 280 min (for the same 200 epochs). The proposed CNN-GRU architecture requires 67 min for its training (again for 200 epochs).

The above mentioned processing time refer to the training phase of the models. It should be mentioned that training is carried out once. After model training is completed and the model parameters (weights) have been estimated, the time needed for modelling the TEC values is negligible. In particular, our trained CNN-GRU model requires 0.96 sec to model the TEC values of over a period of a half year (test dataset-the second half of 2018).

Fig. 5.4 illustrates the training and validation learning curves, as computed for four evaluated models, for the 'ramo' station. As is observed, all models convergences to an acceptable mae error after 200 training epochs.

### 5.3.3  Performance Evaluation for Stations at Different Latitudes and Days of Different Ionosphere Activity

Fig. 5.5 illustrates the VTEC values versus time for the four evaluated non-linear models (CNN-GRU, LSTM, BILSTM, and RNN) at three different ground stations. In this figure, we also depict the ground truth data as a dashed line. The satellites selected in this figure are GNSS visible from the sites: 'graz' (mid-latitude), 'mal2' (near the equator), 'tixi' (aurora region). Also, the satellites depicted in this figure have been selected in various time intervals during the day from each site, focusing more where the highest values of the ionospheric variability are observed. For each station/site, two days are presented in this figure: (i) September the 3rd, which is a day with low ionospheric activity and (ii) September the 11th, in which high ionosphere activity is observed. For these two days ,the values of Dst and F10.7 parameters are illustrated in the up-left diagram of Fig. 5.6. As observed from this figure, the 11th of September is a day of increased solar activity. On the contrary, the 3rd of September is a day with normal solar and geomagnetic conditions. As expected the model performance on 3/9 is slightly better in contrast to predictions during the of 11/9 in which abnormal ionosphere variability is observed [see Fig. 5.5]. However, the model catches the trend of the line in both cases. Also, the model for the 'graz' mid-latitude station, shows better performance than the respective model of 'tixi' station. Overall, what is immediately noticeable in this figure is the CNN-GRU model ability to adequately estimate VTEC values for every satellite separately.

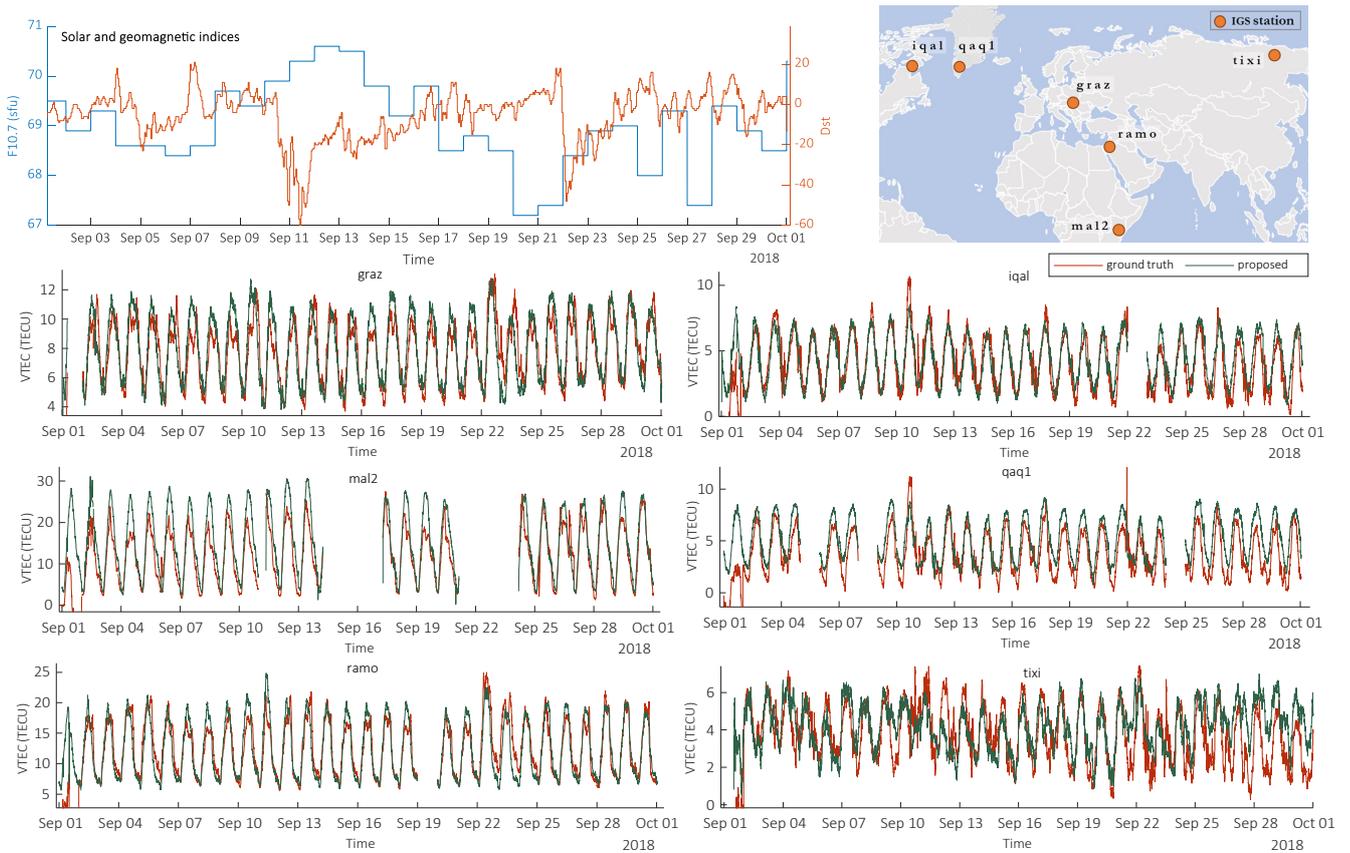

Fig. 5.6 [Up-left]: the diagram shows the daily variations of the solar F10.7 index and the Dst-index. [Up-right]: the map shows the location of each station. [Bottom]: the diagrams illustrate the Vertical TEC values per station on September 2018.

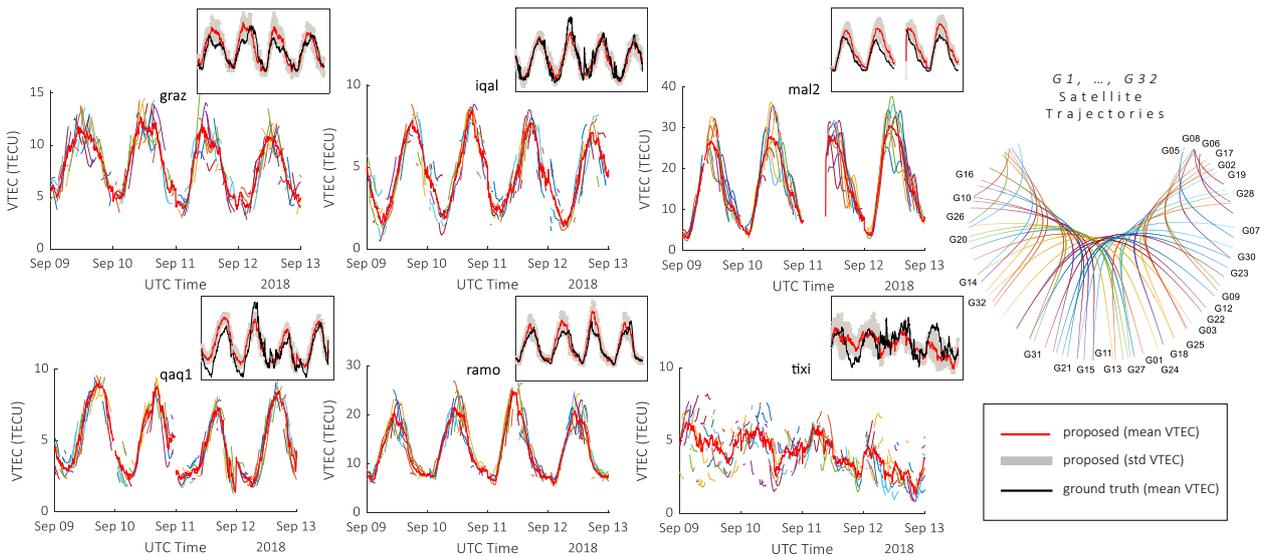

Fig. 5.7 Vertical TEC sequences per satellite (colorful lines) of each ground station for GRU method for four days on September of 2018, and the mean VTEC timeseries per station (red line). On top of each diagram mean red line per station is compared to mean ground truth data per station, as illustrated with black line.

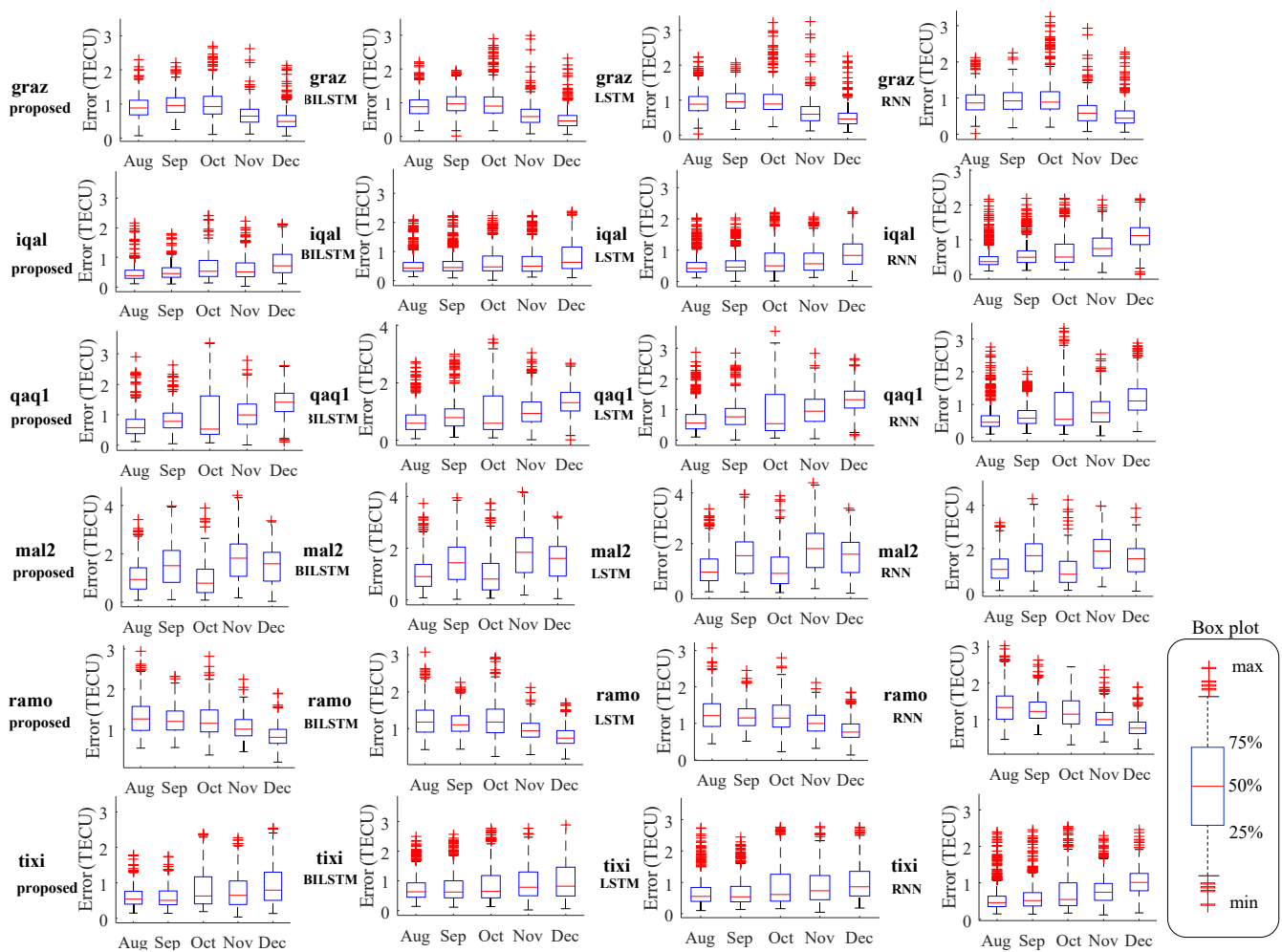

Fig. 5.8 Monthly box plots for the testing period during the second half of 2018, per station.

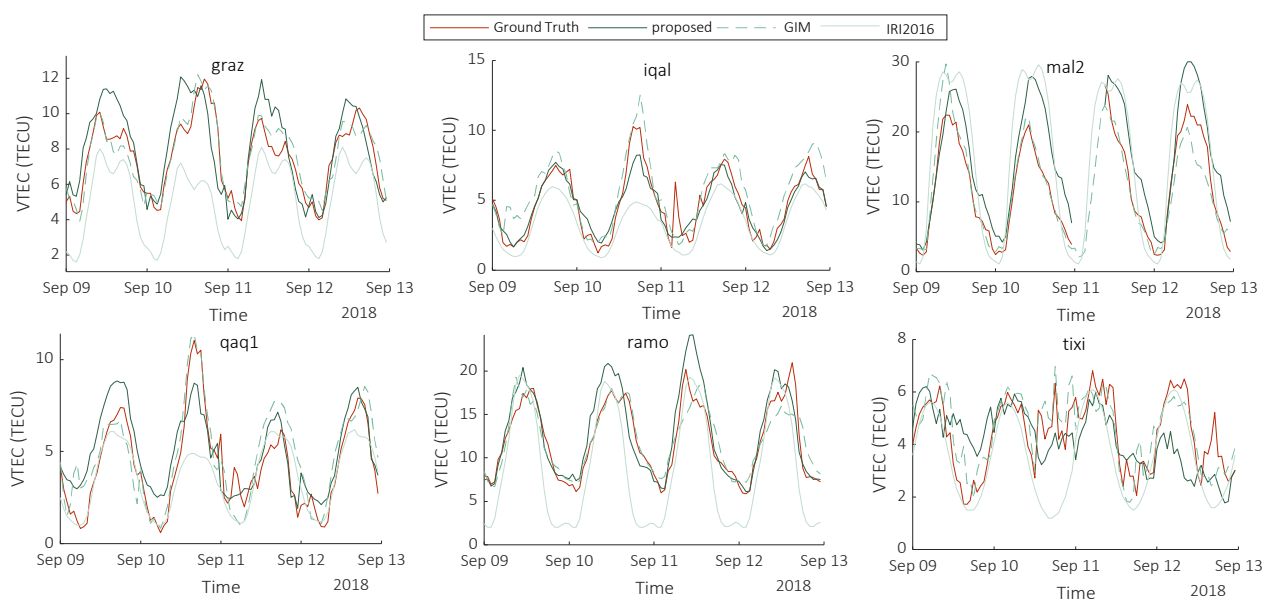

Fig. 5.9 Comparison between VTEC (TECU) RNN-based methods and the estimated GPS iono-spheric VTEC values (ground truth) as well as the VTEC models (GIM, NeQuick, IRI2001 and IRI01-cor), for days between 9 and 12 of September.

Fig. 5.6 shows the predicted VTEC values per station for a time period of a month (September 2018). The predicted values of the proposed CNN-GRU model are illustrated with green color whereas, the red line is the ground truth data. The up-left figure shows the F10.7 and DST values for the days of September. As observed, CNN-GRU predicted values achieve good performance. Also, the proposed CNN-GRU model catches the local maximum VTEC values during the days of increased ionospheric activity as mentioned above (i.e., the 11th of September).

Fig. 5.7 shows the estimated mean VTEC values, as illustrated with red line, for each station, for four different days between 9 and 13 September of 2018, compared to ground truth values with black line. Here, we can observe the diurnal changes, as the VTEC values are varying during each day, depending on the electron density in the ionosphere. As observed, the ionospheric delay changes slowly through a daily cycle. It has its minimum values ($2-4\,TECU$) between midnight and early morning, and reaches its peak during the daylight hours. The 'mal2' stations reaches the bigger mean VTEC values ($\sim 30\,TECU$), whereas the 'tixi' station (located in the aurora region) has the minimum observed values. However the 'tixi' station appears various fluctuations. The mean standard deviation values are 1.20 for 'graz', 0.80 for 'iqal', 2.81 for 'mal2', 0.70 for 'qaq1', 1.78 for 'ramo' and 1.04 for 'tixi' ground station. This is illustrated in Fig. 5.7 where we have depicted with gray the standard deviation values.

### 5.3.4   Performance Evaluation for Different Months

At every time instance (point), the difference between the ground truth VTEC values and the estimated ones from the various non-linear learning models is computed. Thus, for the test time period of a half a year (second half of 2018), we have an error timeseries for every station, deriving from the time-wise differences of VTEC values (see Section 5.2.2). Fig. 5.8 depicts the distribution of these errors per month for the testing period (second half of the year 2018). Each box plot graphically depicts groups of *vtec* error values through their quartiles. Outliers are plotted as individual points, with the red cross symbol. The horizontal red line is the median value of the *vtec* error (50%), whereas the first and third quartile, that is the 25*th* and the 75th percentiles, respectively, are illustrated in black dashed line. As observed, the performance evaluation per month is slightly different, however, these difference are not significant. In most of the cases, the proposed CNN-GRU method has better performance as concerns the outlier values, that are slightly fewer compared to other methods outliers(difference errors > 75%). In August, there are 18% fewer outliers in the proposed CNN-GRU compared to the RNN method. In the other months this percentage is: 5% for September, 13% for October, 17% for November and 9% for December.

### 5.3.5   Performance Evaluation between Different Ionosphere Models

Fig. 5.9 shows the mean VTEC values at every station, for IRI2016 and GIM TEC estimates compared to the ground truth values and CNN-GRU TEC estimations, during four days between 9 and 12 September. The VTEC maxima is shown for all stations between 8:00 and 12:00 AM. In most cases, the CNN-GRU predicted values are similar to ground truth. GIM-aided TEC values are also close to CNN-GRU TEC values. IRI2016 values seems to underestimate the ionospheric variability compared to the ground truth TEC and and TEC values from the GIM model.

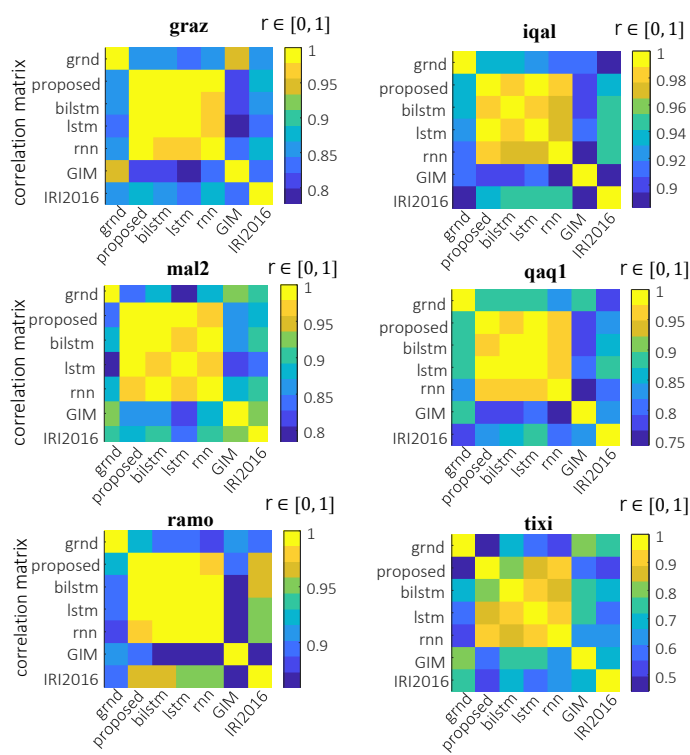

Fig. 5.10 Correlation matrices per station between VTEC (TECU) RNN-based predictions (GRU, BILSTM, LSTM and RNN) and the estimated GPS ionospheric VTEC values (Ground Truth) as well as the VTEC models (GIM and IRI2016), for the time interval between 9 and 12 September. In most of the stations, except for 'tixi' station in aurora region, the correlation coefficient r ranges between the interval [0.75, 1).

Fig. 5.10 shows the correlation coefficient between the proposed CNN-GRU method, the compared non-linear learning methods, the ground truth GPS TEC data and the values derived from IRI2016 and GIM models. Yellow blocks indicate high correlation, whereas, blue blocks show smaller values for the correlation index $r$. Overall, the stations 'iqal', 'mal2' and 'ramo', attain higher correlation values ($r > 0.75$), whereas 'tixi' (located near the equator) performs the worst values of correlation.

## 5.4    Conclusion

In this chapter, regarding the choice a spatio-temporal hybrid deep learning model for TEC modelling, we have proposed a combined CNN-GRU deep learning architecture for TEC modeling. The proposed model has been compared with other linear (e.g., AR and ARMA) and non-linear (RNN, LSTM, BILSTM) regression methods. Our main conclusions are the following:

- The proposed CNN-GRU model achieves an improvement of 16.77% with respect to the RNN architecture and of 12.71% and 9.49% with respect to the LSTM and BILSTM model respectively in case of 98% percentile. This implies the capability of the proposed CNN-GRU network to capture TEC values both in normal and high solar activity periods.

- The average mae improvement of the proposed model compared to the traditional RNN architecture is 23.9% with respect to the validation set, a set that it has not be used during training. In case of the BILSTM model the mae improvement is of 12.65%.

- The computational complexity of the proposed model is retained slow although its higher performance accuracy. In particular, as far as model testing is concerned, our CNN-GRU architecture requires less that 1s (0.96s) to model the TEC values over a period of a half year. Instead, regarding training, our model requires about 67 min to be trained for 200 epochs, in contrast to the BILSTM architecture where training is executed within 280 min for the same number of epochs. It should be mentioned that the training of the model is carried out once.

- The mean standard deviation value of mae of our model is 1.38 over all the 6 examined ground stations. The worst performance is for the 'mal2' station which is located near equator while the best for 'graz' station (mid-latitude).

As future work, we intend to further investigate model reliability, given as inputs GNSS observational data from a regional network of stations close to each other, to examine the extent to which such models capture the regional anomalies. In addition, more complicated deep learning architectures such as semi-supervised learning [94], [95] and/or Generative Adversarial Networks (GAN) [90] can be examined to see if they can improve the results while retaining small computational cost.

# Chapter 6

# The NILM Problem

## 6.1 Introduction

Raising awareness of individuals on environmental protection and sustainability, is prerequisite to set climate policies, responses or solutions to climate change at global scale [96]. There are various ways that householders could contribute to sustainable living. One of them is by reducing their energy consumption. To this end, a change of energy related behavior in the household is required. First, consumers need to become aware of their energy consumption. However, end-consumers often lack knowledge about potential energy savings, existing policy measures and relevant technologies. Most household consumers are usually aware of general information related to their consumption through monthly electricity bills. Nonetheless, the information about their consumption is not translated into good practices and tailored advice for energy saving. Non-Intrusive Load Monitoring (NILM) uses the aggregate power signal of a household as input to estimate the extent to which each appliance contributes to the aggregate energy consumption signal [97]. Thus, NILM is an efficient and cost effective framework for energy consumption awareness, providing itemized energy bills and personalized energy savings recommendations. Power disaggregation is applied to enhance awareness on the energy consumption behavior of consumers in the household and therefore guide them towards a prudent and rational utilization of energy resources [98].

Another usage of the disaggregated electrical consumption is to identify malfunctioning appliances. As an instance, NILM approach is applied to detect the frosting cycle of a fridge with a damaged seal, which is more frequent than the normal one [98]. Applying automated analysis is essential to detect appliance's performance problem; this can effectively be accomplished through online energy disaggregation [99].

Demand side response provides the possibility of shifting demand away from the peak and thus decreasing the corresponding cost of energy, although the average daily consumption is not altered [99]. Furthermore, it helps modifying the demand based on the intermittent power generation of renewable energy sources; hence decreases the need for expensive energy storage systems. Energy disaggregation permits the utility companies to identify a device with a high consumption rate at a peak hour in a household and sends a message to the corresponding users asking them to postpone their usage to smooth out the current peak in the demand.

Disaggregation of households' power consumption allows grid operators to improve their predictions in energy demand and is an important part of providing a stable supply of power to all customers on a power grid [100]. In [101] and [100] the consumption profile of appliances is identified through disaggregation and then, the obtained appliance-level load profiles along with meteorological information are employed to predict the future usage.

For the above mentioned reasons energy disaggregation is of great importance for energy conservation and planning. Given the power consumption per appliance the forward problem is to predict the total power consumption in a household. We assume the aggregate signal $p(t)$ at a discrete time index $t$, to be equal to the summary of the individual appliances' power consumption $p_m(t)$ plus an additional noise $\epsilon(t)$. Thus, the total power consumption $p(t)$ is:

$$\tilde{p}(t) = \sum_{m=1}^{M} p_m(t) + \epsilon(t) \tag{6.1}$$

In Eq. (9.1) variable $m$ refers to the $m$-th out of $M$ available appliances. Under a NILM framework, the individual appliance power consumption $p_m(t)$ is not a priori available, assuming the absence of installed smart plugs. Instead, only $\tilde{p}(t)$ is given. The inverse ill-posed problem, called NILM, is to calculate the best estimates $\hat{p}_m(t)$ of the appliance power consumption, given the aggregate power value $p(t)$.

Various approaches have been proposed to solve the NILM problem, as presented in Section 6.2. Some of the most successful ones exploit deep learning neural network structures for modelling an energy disaggregation problem (e.g., [56]). Nevertheless, non-intrusive load monitoring is a difficult task, especially when considering nonlinear and continuous appliances, and until recently, there are barriers and limitations that have not been properly addressed. In particular, the proposed techniques have not been applied successfully across different households and datasets [57]. Thus, it is difficult to create algorithms with a good *generalization ability*. In addition, noisy aggregate energy consumption measurements significantly deteriorate the performance of NILM methods. Usually, the detected appliances have unsteady signatures or present abnormal behavior and additionally, the existence of noisy as well as inadequate datasets deteriorates overall models' performance [58]. The latter issue is more evident in case that deep learning structures are used for energy disaggregation, since a large number of labeled samples is required during training.

Here, we collect the good practices for solving the NILM problem, providing suggestions and recommendations in three main pillars: (i) data pre-processing, (ii) NILLM algorithms and (iii) outcomes validation. A short literature review, divided into three different periods follows.

## 6.2 A Brief NILM Literature Review

### 6.2.1 The early NILM era [1995–2014]

Hart was the first to propose a method for disaggregating electrical loads based on Combinatorial Optimization (CO) through clustering of similar events based on appliances' characteristics [97]. The first approach to NILM employed Combinatorial Optimization, which at the time was the standard technique for disaggregation problems. For historical review of the evolution of NILM techniques, see for instance [102, 1, 55]. This first approach had a major shortcoming: Combinatorial Optimization performed the power disaggregation on each instant independently of the others, without considering the load evolution through time. Most common approaches to solve the NILM problem are based on unsupervised event detection in the aggregate signal, whereas supervised classifiers are used to assign known appliances to detected events in order to estimate the power trace of individual appliances [103]. Different classification tools have been widely used, including Sup-port Vector Machines (SVM) [104], neural networks, Decision Trees (DT) [105], and hybrid classification methods [106], [107]. Contrary to the aforementioned classic methods, other methods such as Dynamic Time Warping (DTW) are used for comparing and grouping windows from daily profiles and identifying unique load signatures [108]. The main object of controversy in these approaches refers to the difficulty of classifying multi-state appliances [106]. Multi-state appliances require a long range pattern to be trained for their detection [109]. Graph Signal Processing (GSP) [107] is a concept that effectively captures spatio-temporal correlation among data samples by embedding the structure of signals onto a graph. Zhao et al. [110] propose a low-resolution, event-based, unsupervised GSP approach. Recently, a Modified Cross-Entropy method for event classification has been suggested [111], which is based on CO and formulates NILM as a Cross-Entropy problem.

Hidden Markov Models (HMM) and various extensions of this model are advocated in order to explore the possible combinations among the different appliances' state sequences [112, 113, 25, 114]. In this light, HMMs are state-based, so the studied appliances should have discrete states in their signatures [103]. As the number of appliances increases, the number of combinations of states sequences is increased exponentially, increasing respectively problem's complexity [103]. In addition to this, time complexity is also increased, leading to the reduction of model's classification performance [115]. Makonin et al. [116] have proposed a super-state HMM and a sparse Viterbi algorithm in order to reduce the complexity. Another limitation of HMM-based approaches is that they tend to fail in the presence of unknown appliances [103]. Rahimpour et al. [117] proposed a matrix factorization technique for linear decomposition of the aggregated signal using as bases of this learned model the appliances' signatures resulting in an efficient estimation of the energy consumption per appliance.

## 6.2.2 Deep learning based NILM [2015–2019]

Soon after various open datasets became available ([118], [119]), and thanks to the increased number of datasets coming from smart electric meters installed in domestic residences, the proposed solutions to NILM shifted to a supervised learning process. With the rise of deep learning, a new family of methods have been introduced that exploit deep neural network structures to solve the ill-posed NILM problem. Deep learning techniques have been applied mostly to low frequency NILM approaches since 2015 [99].

It's a common way to treat aggregated signal as a corrupted by noise signal of an appliance. Under this view, denoising autoencoders (DAE) are excellent techniques used to to reconstruct a signal from its noisy version. This architecture has been initially proposed by Kelly and Knottenbelt [99], while others expand the idea proposing alternative DAE architectures, such as [113].

Exploiting the temporal character and dependencies of the power signal, another deep learning scheme called Recurrent Neural Networks (RNN), have proved efficient under the NILM framework. Here, NILM is treated as a supervised learning problem with times series. RNNs and their variants, such as Long Short-Term Memory Networks (LSTM) and Gated Recurrent Units (GRU) have been primarily used, as they are very popular and effective with 1D time series data. Relevant studies have been carried out in the past ([57], [99], [109]). In a previous work of ours, we have also proposed a Bayesian optimized bidirectional LSTM model for NILM [56], whereas in [54] a context-aware LSTM model adaptable to external environmental conditions is presented.

Although Convolutional Neural Networks (CNN) are traditionally developed for two-dimensional imagery data ([36]), one-dimensional CNN can be used to model the temporal character of sequential timeseries data. Few researches [1] have tried to enrich CNN structures providing a recurrent character, such as CNN-LSTM and Recurrent Convolutional Networks. In [3] a causal 1-D convolutional neural network for NILM is proposed. Others introduce the concept of data sequences [99] to feed the classic structure with historical past values of power load. Others [120] propose a sequence to point CNN architecture, underscoring the importance of sliding windows to handle long-term timeseries. Alternatively, sequence to sequence architectures have also been proposed [121].

### 6.2.3 Advancements in NILM [2020– ]

Recently, there are various advanced machine learning methods applied for NILM. These methods provide competitive accuracy against the traditional NILM methods. Some of the worth mentioned works are presented here.

Generative adversarial networks recently have been applied for NILM. [122] adopt a GAN-based framework for solving NILM in an early attempt. Then, Kaselimi et al. [79] propose a generative adversarial network for sequence to sequence learning, whereas Pan et al. [2] achieve sequence to sub-sequence learning with conditional GANs. Chen et al. [123] propose a context aware convolutional network for NILM that has been trained adversarially. Most of these studies exploit the robustness to noise that the adversarial training process achieves.

Transformer models were explored as alternative architecture for neural machine translation tasks within the past two years [124]. Recently, a transformer based architecture that utilizes self-attention for energy disaggregation has been adopted by [125], to handle power signal sequential data.

Given that, most of the existing DNNs models for NILM use a single-task learning approach in which a neural network is trained exclusively for each appliance, propose UNet-NILM for multi-task appliances' state detection and power estimation, applying a multi-label learning strategy and multi-target quantile regression. The UNet-NILM is a one-dimensional CNN based on the U-Net architecture initially proposed for image segmentation.

Explainable AI (XAI) attempts to promote a more transparent AI through the creation of methods that make the function and predictions of machine learning systems comprehensible to humans, without sacrificing performance levels. Explainable NILM networks these methods provide competitive accuracy, the inner workings of these models is less clear.

## 6.3   Conclusion

Usually we treat NILM as a regression or classification problem, but actually NILM is an inverse problem and that explains its complexity. There are various open access datasets to help the research community, however the real life problem is too complex, and usually algorithms with good performance in a specific dataset, do not share the same performance for other noes. The variety of different appliances and types of appliances with various characteristics and

# Chapter 7

# The proposed CoBiLSTM model for NILM

## 7.1   Our contribution

Although there is a significant interest from the industry, the research community has not yet explored NILM techniques at a larger scale and across different households. This is mainly because: (i) techniques developed for one household cannot be applied to other households (applicability), (ii) the developed models do not have the ability to perform satisfactorily across different households (scalability), (iii) and these models cannot be transferred across different datasets (transferability) [126]. The main cause of the aforementioned difficulties lies in the fact that energy consumption patterns of the appliances are highly correlated with the contextual conditions on which these appliances operate. Hereby, the term context refers to: (i) the external weather and environmental conditions (seasonal variations), (ii) the geographic location of the houses, (iii) the technology and the configuration set-up of an appliance (eco versus normal or even turbo mode of an operation) and (iv) other appliances' operational anomalies (malfunction, etc.) Therefore, a dynamic context adaptive model is developed to capture the new functional modes of an appliance, as the contextual conditions are changing. The main innovative characteristics of the proposed CoBiLSTM method are the following:

**Sequence-to-sequence**:  Rather than estimating single time-points (Sequence-to-Point), our proposed method estimates a whole target sequence, given a sequence of aggregate signal (Sequence-to-Sequence). The final output is derived as the average of the different overlapping output values, thus increasing accuracy. Furthermore, we introduce a bidirectional LSTM network as in one of our earliest work in [103], which means that our approach remembers and draws inferences from future data points as well as from the past ones for the target appliance. However, our previous work [103] does not take into consideration contextual conditions and model adaptation techniques due to the evolution of the context.

**Adaptability**: We propose a modular and a self-trained energy disaggregation model with the capability of updating its behavior whenever a significant change of contextual conditions is detected. Since it is difficult to define the general polyparametric notion of context and consequently, decide when a model update is required, an alternative approach is adopted. We implicitly define the context through the distribution of the performance errors of the energy disaggregation model during its operation. For this, the pursued solution to the problem of detecting changes in appliance operation will be sought in the space of model performance deterioration observations, instead of the space of contextual factors' changes. In particular, a context change is considered to occur when the distance between the probability distribution of the errors during model operation and the probability distribution during the training and validation phase, is statistically significant (Section 7.3.2). Then, model retraining is activated, through appropriate modification of the network weights, to implement the context aware adaptive energy disaggregation strategy. As we have stated previously, we call this model CoBiLSTM, indicating its adaptive capability to fit to current context.

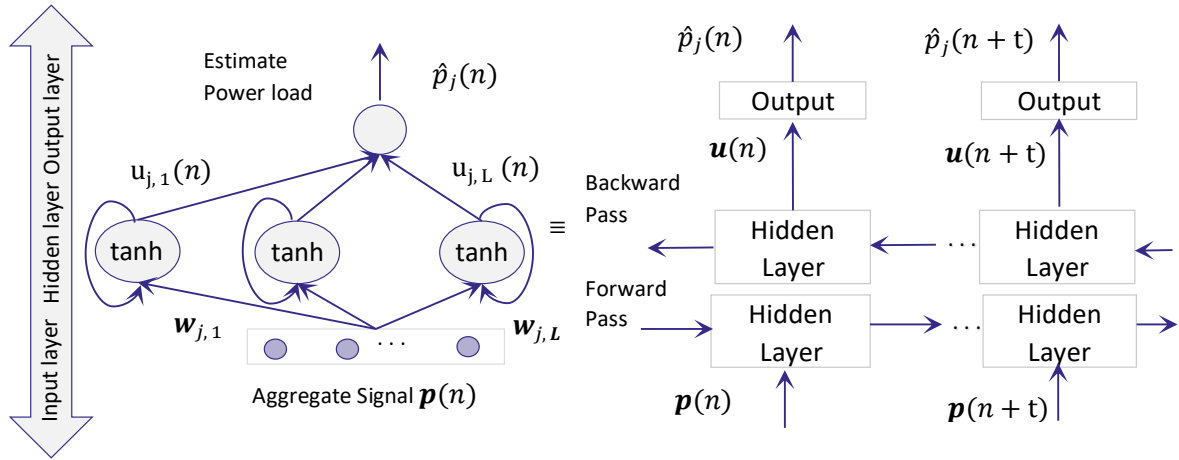

Fig. 7.1 The short-range recurrent regression model. (a) The bidirectional short-range recurrent regression model. (b) Unfolded representation in time where the forward and backward passes are depicted.

**Scalability**: The proposed method is capable of handling situations where statistically significant differences exist compared with the model training conditions. This means that the proposed model is capable of generalizing its behavior to data of quite different statistics than those they have been trained on. Instead, traditional methods [103, 127] fail to provide accurate regression results in datasets of different statistical behavior than the trained ones.

**Regression**: The proposed approach handles energy disaggregation as a regression problem instead of a classification one- as many traditional approaches do. Regression is beneficial as the model can correctly detect not only the switch-on/off events but also reproduce the shape of the target appliance load.

**Modularity**: Usually, the traditional energy disaggregation approaches (such as [113]) assume a single model for all appliances. However, such an approach increases model complexity especially in cases where the number of appliances increases. The proposed model adopts a modular design by assigning a different regressor for every type of appliance. Similar approaches are also adopted in other Recurrent Neural Network (RNN) methodologies such as [57].

Fig. 7.2 The memory cell of an LSTM network. It contains three different components; (i) the forget gate $F(n)$, (ii) the input gate $I(n)$ and input node $H(n)$ and (iii) the output node $O(n)$.

The remainder of this chapter is structured as follows: In Section 9.2, NILM is formulated as a non-causal problem for which a bidirectional LSTM regression model is proposed. In Section 7.3, a method for dynamic adaptation and optimal hyperparameter selection under a Bayesian framework is described. In Section 7.4, the proposed method is experimentally evaluated against state of the art energy disaggregation methods in publicly available datasets, while Section 7.5 concludes the final results.

## 7.2 Bidirectional LSTM Regression for Non-causal Energy Disaggregation

### 7.2.1 Bidirectional Short-Term Recurrent Regression

One way to model the unknown function $g(\cdot)$ of Eq. (2) is by means of a Feedforward Neural Network [32]. Assuming L hidden neurons and one linear output layer, the estimate $p_j(n)$ is given by

$$\hat{p}_j(n) = \mathbf{u}_j(\mathbf{n})^T \cdot \mathbf{v}_j \qquad (7.1)$$

$$\mathbf{u}_j(n) = \begin{bmatrix} u_{j,1}(n) \\ \vdots \\ u_{j \cdot L}(n) \end{bmatrix} = \begin{bmatrix} tanh(\mathbf{w}_{j,1}^T \cdot \mathbf{p}(n)) \\ \vdots \\ tanh(\mathbf{w}_{j,L}^T \cdot \mathbf{p}(n)) \end{bmatrix} \qquad (7.2)$$

$tanh(\cdot)$ refers to the hyperbolic tangent, so each element of the vector $u_j(n)$ ranges between -1 and 1. The weights $w_{j,i}, i = 1, \ldots, L$, connect the input p(n) with the i-th hidden neuron. Similarly, $v_j$ are the weights that connect the hidden neurons with the output neuron. Henceforth, subscript j is omitted for simplicity. Vector u(n) gathers the outputs of all L hidden neurons $u_{j,i}$ and is a state vector expressing latent operational modes of the j-th appliance. Their values range within the $[-1,1]$ interval. Each appliance has its own hidden states and these latent variables are estimated through a learning process. Since the power load of an appliance follows a non-causal relationship, the value of a state depends not only on its previous values and but also on its future ones (Fig. 7.1). Therefore,

$$u_i(n) = tanh(\mathbf{w}_i^T \cdot \mathbf{p}(n) + \vec{\mathbf{r}}_i^T \cdot \mathbf{u}(n-1) + \bar{\vec{r}}_i^T \cdot u(n+1)) \qquad (7.3)$$

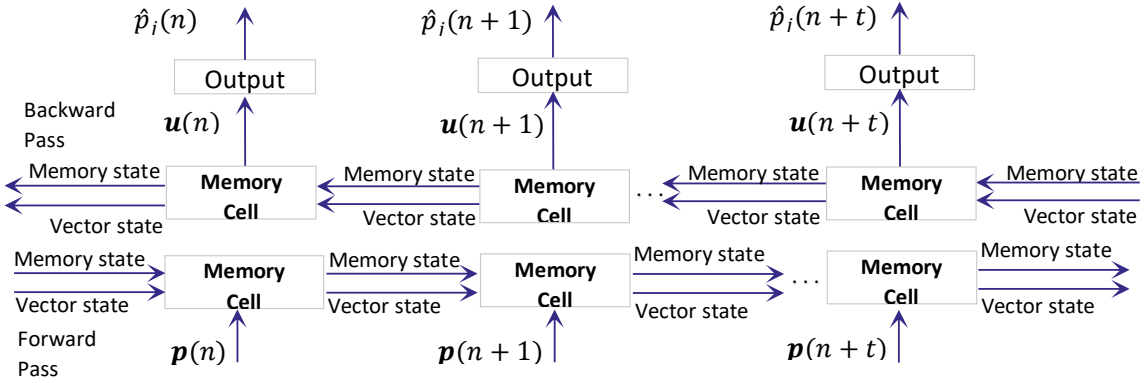

Fig. 7.3 Bidirectional LSTM regression model used for NILM estimation.

In Eq. 7.3, $\vec{r}_i$ is a weight vector expressing the dependencies of previous states, while $\bar{r}_i$ the future state dependencies [33].

### 7.2.2   Bidirectional Long-Term Recurrent Regression

Short-range dependencies are not adequate for power load estimation, as appliances often follow repeated patterns spanning long time periods. For this reason, bidirectional LSTM networks are adopted as an alternative regression model for power load estimation. LSTMs are of similar structure with the bidirectional recurrent regression model, but each node in the hidden layer is replaced by a memory cell, instead of a single neuron [128]. The structure of a single memory cell is depicted in Fig. 7.2, while Fig. 7.3 indicates an unfolded LSTM network over time. The memory cell contains three different components (see Fig. 7.2): (i) the forget gate, (ii) the input node and the input gate, and (iii) the output gate. Each component applies a non-linear relation to the inner product between the input vectors and respective weights (estimated through a training process). Some of the components have the sigmoid function, expressed as $\sigma(\cdot)$ in Fig. 7.3, while others the hyperbolic tangent function, $tanh(\cdot)$. Forget gate $F(n)$ separates the worth-remembering information from the unnecessary information, by keeping the latter out of the memory cell. Input node $H(n)$ activates appropriately the respective state (true or false output from the "tanh" activation). Input gate $I(n)$ regulates whether the respective hidden state is "significant enough" for the accurate estimation of the power loads $p_j(n)$. Output gate O(n) regulates whether the response of the current memory cell is "significant enough" to contribute to the next cell. In the context of washing machine operation, for example, the output gate indicates that if a pre-washing cycle is currently detected, a washing cycle operation is expected in the next time instances. One shortcoming of the memory cell of Fig. 7.2, is that it processes only previous state information. Bidirectional LSTM, on the other hand, processes data in both directions, including a forward and, additionally, a backward operation. The structure of a bidirectional LSTM, unfolded in time, is presented in Fig. 7.3. The operation of all the aforementioned modules is mathematically formulated by:

$$
\begin{aligned}
\{F(n),\ H(n), I(n), O(n)\} =\ \{\sigma, \tanh\}(\\
\mathbf{w}^{T,F,H,I,O} \cdot \mathbf{p}(n) + \vec{\mathbf{r}}^{T,\{F,H,I,O\}} \cdot \mathbf{u}(n-1) + \overleftarrow{r}_{j,i}^{T} F, H, I, O \cdot u_j(n+1))
\end{aligned}
\tag{7.4}
$$

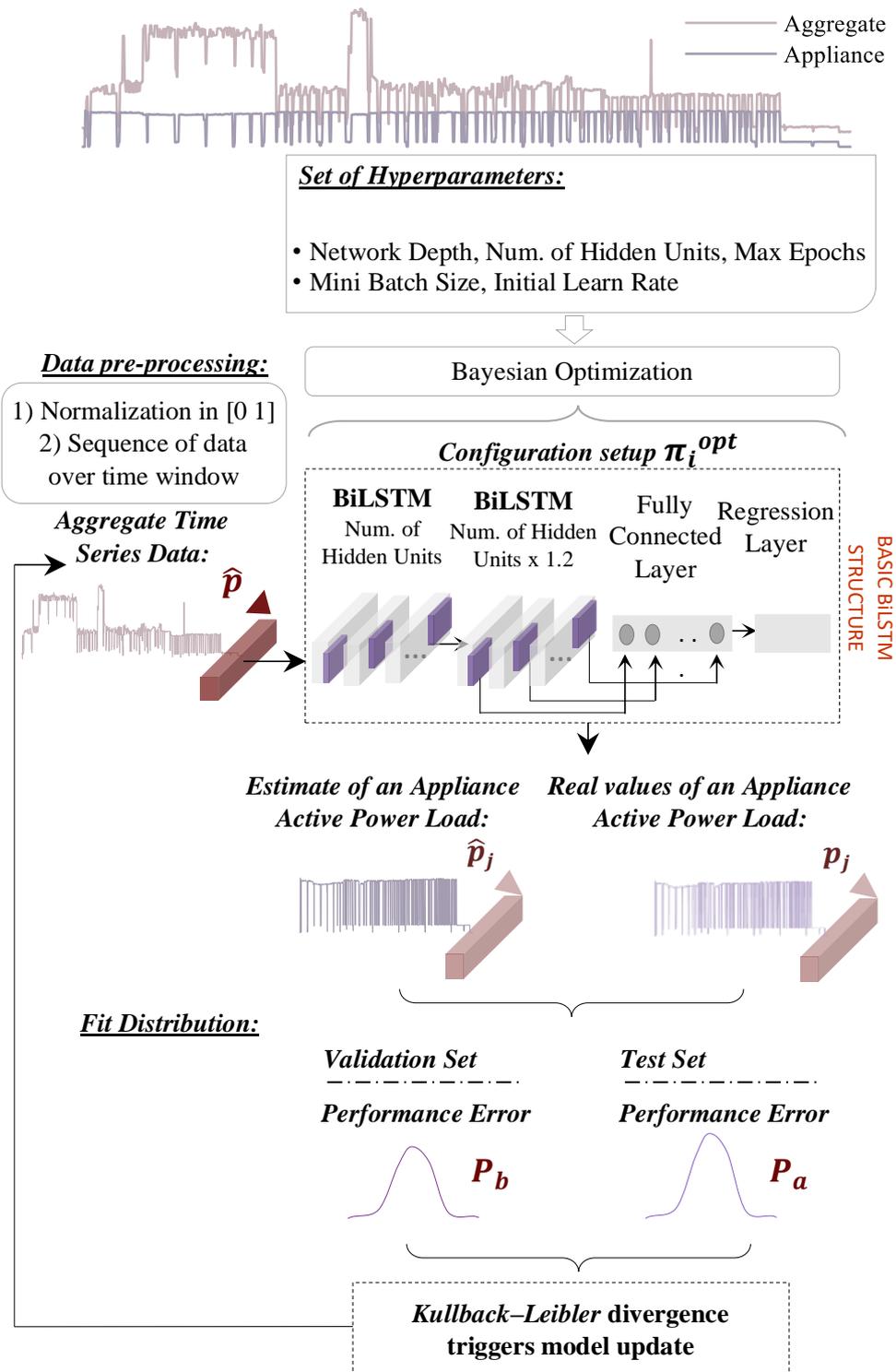

Fig. 7.4 The proposed CoBiLSTM deep learning structure.

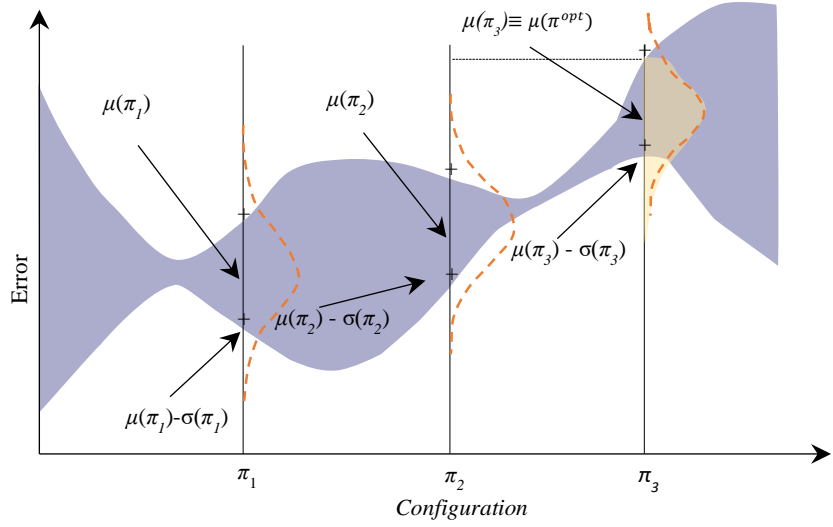

Fig. 7.5 A graphical illustration of Bayesian optimization.

In Eq. (5), variables $w^{(F,H,I,O)}$, $\vec{r}^{F,H,I,O}$ and $\overleftarrow{r}_{j,i}^{(F,H,I,O)}$ have similar meaning as the ones involved in Eq. (4), with the difference that each node or gate has its own unique parameters.

### 7.2.3 A Detailed Description of the Proposed LSTM Network Configuration

Aggregate energy consumption signals (timeseries data) were used as inputs for the described bidirectional LSTM networks. Initially, the input data are down-sampled in 1 min intervals (in datasets with denser time intervals) and additionally, are normalized so that their values range in [0, 1]. The input data represent a time window of an appliance operation. For appliances that do not belong to the Type-IV (permanently active) category, the time window is set to be the duration of the active operation of the appliance. For Type-IV devices, a time window of two hours is considered.

The parameters of the bidirectional LSTM network are selected through the Bayesian optimization approach described in Section 7.3.3. In particular, the following network parameters are optimized: network depth, number of hidden units, maximum number of epochs, mini batch size and initial learning rate. The optimization strategy starts from an initial configuration network set up and then, estimates the next network parameters in a way that model performance is improved. Therefore, the parameters of the bidirectional LSTM network are dynamically adjusted to fit appliance-specific characteristics as well as current contextual attributes. Fig. 7.4 shows the adopted strategy and the basic structure of the proposed CoBiLSTM model. A network configuration setup is provided for the clothes dryer: (i) the input layer, (ii) the bidirectional LSTM layers and (iii) the regression layer. The input layer receives the current data in a sequence-like form, of a time window with 60 min duration. The network includes two bidirectional LSTM layers: the first layer consists of 60 filters with a kernel size 1x5, while the second layer consists of 72 filters of the same kernel size. The regression layer consists of one fully connected hidden layer of 60 output neurons, and the final output regression layer. The output is a sequence of active power load with size 60 and not a single point. Further details regarding model's update constraints based on Kullback–Leibler (K-L) divergence are presented in Section 7.3.2.

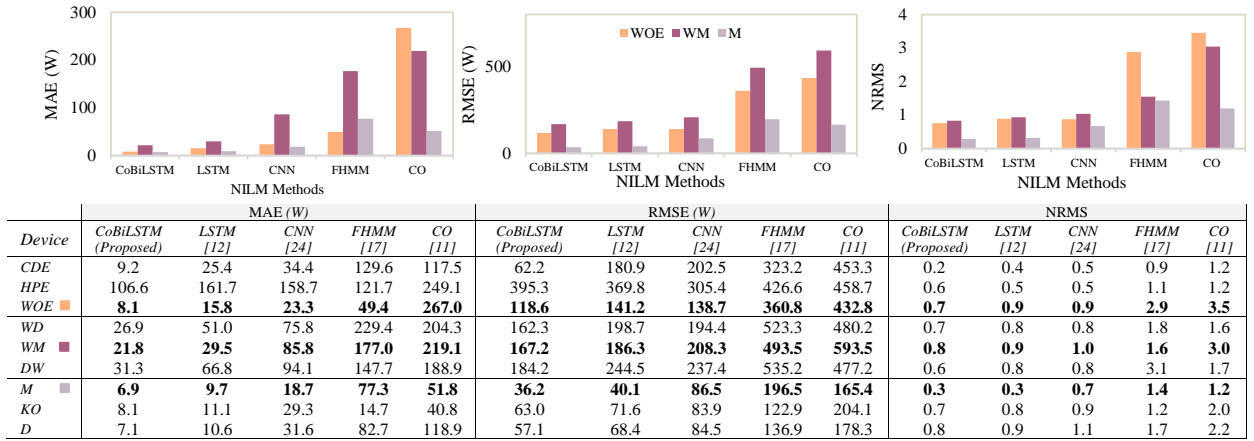

| Device | MAE (W) | | | | | RMSE (W) | | | | | NRMS | | | | |
|---|---|---|---|---|---|---|---|---|---|---|---|---|---|---|---|
| | CoBiLSTM (Proposed) | LSTM [12] | CNN [24] | FHMM [17] | CO [11] | CoBiLSTM (Proposed) | LSTM [12] | CNN [24] | FHMM [17] | CO [11] | CoBiLSTM (Proposed) | LSTM [12] | CNN [24] | FHMM [17] | CO [11] |
| CDE | 9.2 | 25.4 | 34.4 | 129.6 | 117.5 | 62.2 | 180.9 | 202.5 | 323.2 | 453.3 | 0.2 | 0.4 | 0.5 | 0.9 | 1.2 |
| HPE | 106.6 | 161.7 | 158.7 | 121.7 | 249.1 | 395.3 | 369.8 | 305.4 | 426.6 | 458.7 | 0.6 | 0.5 | 0.5 | 1.1 | 1.2 |
| WOE ■ | **8.1** | **15.8** | **23.3** | **49.4** | **267.0** | **118.6** | **141.2** | **138.7** | **360.8** | **432.8** | **0.7** | **0.9** | **0.9** | **2.9** | **3.5** |
| WD | 26.9 | 51.0 | 75.8 | 229.4 | 204.3 | 162.3 | 198.7 | 194.4 | 523.3 | 480.2 | 0.7 | 0.8 | 0.8 | 1.8 | 1.6 |
| WM ■ | **21.8** | **29.5** | **85.8** | **177.0** | **219.1** | **167.2** | **186.3** | **208.3** | **493.5** | **593.5** | **0.8** | **0.9** | **1.0** | **1.6** | **3.0** |
| DW | 31.3 | 66.8 | 94.1 | 147.7 | 188.9 | 184.2 | 244.5 | 237.4 | 535.2 | 477.2 | 0.6 | 0.8 | 0.8 | 3.1 | 1.7 |
| M ■ | **6.9** | **9.7** | **18.7** | **77.3** | **51.8** | **36.2** | **40.1** | **86.5** | **196.5** | **165.4** | **0.3** | **0.3** | **0.7** | **1.4** | **1.2** |
| KO | 8.1 | 11.1 | 29.3 | 14.7 | 40.8 | 63.0 | 71.6 | 83.9 | 122.9 | 204.1 | 0.7 | 0.8 | 0.9 | 1.2 | 2.0 |
| D | 7.1 | 10.6 | 31.6 | 82.7 | 118.9 | 57.1 | 68.4 | 84.5 | 136.9 | 178.3 | 0.8 | 0.9 | 1.1 | 1.7 | 2.2 |

Fig. 7.6 Performance metrics (MAE, RMSE, NRMS) per dataset (AMPds, REFIT, REDD). The vertical axis in the table shows the three selected appliances per dataset. The horizontal axis indicates MAE, RMSE and NRMS metrics per method. The three diagrams illustrate the performance (MAE, RMSE, NRMS) per method for three selected appliances: WOE/AMPds (orange), WM/REFIT (purple) and M/REDD (light purple), that proved to have the best performance.

## 7.3 Context-Aware Adaptive Energy Disaggregation

The way that some appliances operate depend on specific contextual attributes of the environment. For example, in case of an air-conditioning device, seasonality, external weather conditions or even technology adopted (e.g., inverter or not), affect the energy consumption patterns. Another characteristic example lies in energy signatures generated by a washing machine which highly depend on laundry loads. Such parameters are in effect contextual conditions affecting NILM performance. The purpose of the context-aware adaptation mechanism is to dynamically adapt the deep learning model so that it captures the new contextual operational conditions of the appliances, thereby improving the performance of NILM modeling.

| | Minimum Power Difference (W) | | | | | Maximum Power Difference (W) | | | | | Average Power Difference (Standard Deviation) (W) | | | | |
|---|---|---|---|---|---|---|---|---|---|---|---|---|---|---|---|
| *Device* | *CoBiLSTM (Proposed)* | *LSTM [12]* | *CNN [24]* | *FIMM [17]* | *CO [11]* | *CoBiLSTM (Proposed)* | *LSTM [12]* | *CNN [24]* | *FIMM [17]* | *CO [11]* | *CoBiLSTM (Proposed)* | *LSTM [12]* | *CNN [24]* | *FIMM [17]* | *CO [11]* |
| *CDE* | 0.1 | 2.2 | 0.1 | 0.2 | 1.0 | 594.8 | 635.5 | 793.4 | 1518.0 | 1627.1 | 19.4 (82.7) | 31.9 (94.8) | 29.3 (113.0) | 66.1 (89.9) | 58.3 (141.4) |
| *HPE* | 2.2 | 1.5 | 21.6 | 1.9 | 0.4 | 1105.9 | 834.1 | 741.3 | 389.7 | 882.0 | 106.2 (215.3) | 161.5 (182.1) | 136.1 (132.4) | 156.8 (110.4) | 78.07 (172.4) |
| *WOE* | 0.0 | 0.0 | 3.1 | 1.6 | 5.3 | 672.2 | 692.3 | 725.4 | 1143.9 | 1102.4 | 8.0 (51.9) | 15.8 (66.8) | 17.5 (64.6) | 60.7 (134.0) | 122.2 (141.7) |
| *WD* | 0.1 | 0.1 | 0.1 | 0.1 | 0.1 | 1090.9 | 1280.8 | 1051.9 | 2034.8 | 1770.6 | 26.9 (99.3) | 50.9 (143.6) | 45.4 (135.7) | 176.9 (260.2) | 219.0 (193.3) |
| *WM* | 0.1 | 0.1 | 0.6 | 0.1 | 0.1 | 518.9 | 519.8 | 568.2 | 788.5 | 1462.8 | 21.8 (71.2) | 29.4 (79.9) | 65.2 (87.9) | 139.1 (154.2) | 188.7 (278.1) |
| *DW* | 0.1 | 0.1 | 0.7 | 0.1 | 0.1 | 552.6 | 769.7 | 699.8 | 1624.3 | 1787.7 | 31.3 (69.2) | 66.8 (114.4) | 67.8 (99.7) | 229.4 (350.3) | 204.2 (245.8) |
| *M* | 0.9 | 3.4 | 4.4 | 19.6 | 5.5 | 103.9 | 112.1 | 151.1 | 742.9 | 226.5 | 6.9 (13.5) | 9.6 (14.2) | 15.8 (23.0) | 77.3 (109.0) | 51.7 (32.1) |
| *KO* | 0.7 | 0.4 | 0.4 | 1.5 | 67.0 | 100.6 | 113.2 | 160.8 | 369.5 | 361.9 | 8.1 (16.3) | 11.1 (20.6) | 18.5 (26.2) | 82.7 (43.7) | 118.8 (41.4) |
| *D* | 0.6 | 0.7 | 0.9 | 0.1 | 1.0 | 231.8 | 269.3 | 196.9 | 233.3 | 524.1 | 7.1 (27.1) | 10.6 (31.8) | 18.9 (31.8) | 14.6 (34.9) | 40.7 (83.4) |

Fig. 7.7 Metrics (min, max, average and standard deviation) of the difference between the real and the estimated values of appliance active power.

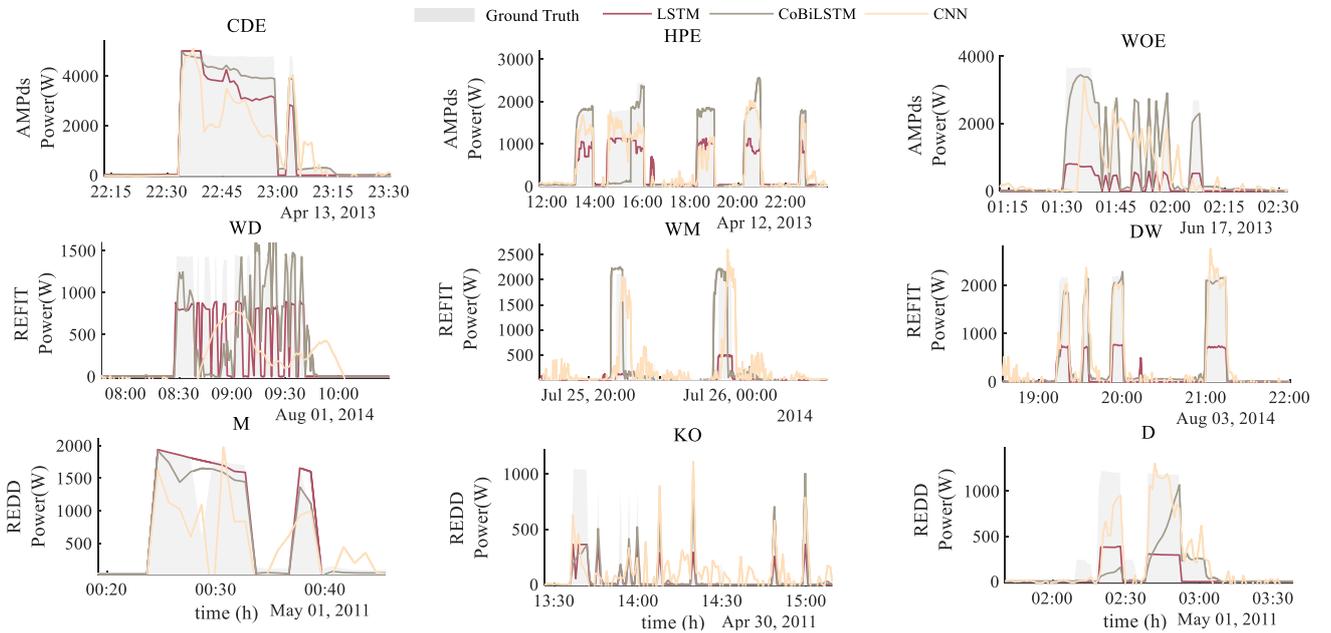

Fig. 7.8 Comparison between CoBiLSTM, CNN and LSTM methods.

### 7.3.1 Context-Aware Energy Disaggregation Framework

Let us denote by $g_j^b(\cdot)$ the non-linear model of the j-th appliance before model updating due to a change of the contextual conditions. Model $g_j^b(\cdot)$ has been trained and validated on data of a set $S^b$, which is divided into two mutually exclusive sets; the training $S_{tr}^b$ and the validation $S_v^b$ dataset. Set $S_{tr}^b$ trains the network, whereas set $S_v^b$ assesses the energy disaggregation model performance on data outside the training set. In order to build up an updated version of the model $g_j^b(\cdot)$, we need new additional samples capturing new contextual conditions of the j-th appliance. Let us denote by $S^a$ the set containing the samples, which is again divided into a training and a validation subset. Considering the new samples of $S^a$ that model the current context as well as samples of the previous knowledge in $S^b$, we deliver a new updated model for the j-th appliance, denoted as $g_j^a(\cdot)$.

The superscript variable a stand for the energy disaggregation model after updating. Data of the training set $S^a$ are obtained by using a set of houses equipped with smart plugs where the actual active power load of each appliance is available. In our application scenario, we assume that the smart-plug houses are very limited and they are only used to collect data for model updating. Then, the updated version of the model is delivered to non-smart-plug houses. Therefore, smart plugs are only used for network updating, that is, for collecting the new training samples in order to retrain the network to capture the new contextual conditions. It should be mentioned that the proposed scheme for dynamic model updating does not negatively impact on the practical applicability of the algorithm. Instead, it provides a framework that enables the model to be dynamically modified to include new operational ways of an appliance. In other words, the proposed context-aware adaptive energy disaggregation framework is a methodology to dynamically assess the performance of the model in a pool of representative houses. In case that model performance is not satisfactory, we appropriately retrain the network (i.e., modify its weight parameters) so that they can capture new contextual operational conditions of the appliances (i.e., different consumption signatures).

### 7.3.2 The Algorithm for Model Updating

Using Eq. (2), we can compute the performance error of the model $g_j^b(\cdot)$ over data of the validation set $S_v^b$. Let us denote by $e^b(i_n)$, $i_n \in S_v^b$ the performance error over all data of the set $S_v^b$. These errors form a probability distribution function (pdf), denoted by $P_b$, expressing the error distribution of the NILM model at those contextual conditions on which the model has been trained to over the $S_v^b$ data. Let us now assume that the trained energy disaggregation model is applied to another contextual condition than the one on which the model has been trained. Then, we evaluate the performance of the model over a time window of duration T. This evaluation is carried out only on the representative houses equipped with smart plugs. In our experiments, T is assumed as equal to 30 days. Therefore, assuming a sampling rate of 1 min, a collection of $30 \times 1{,}440 = 43{,}200$ errors is gathered. Let us denote by $Q(n)$ the probability distribution of all errors over the time window T. $Q(n)$ models the current contextual conditions at a time instance n where the energy disaggregation model is applied to. Therefore, $Q(n)$ can be considered as a metric of the current context $c(n)$ which an appliance operates on.

$$c(n) \sim Q(n) \tag{7.5}$$

Depending on the values of distribution $Q(n)$, two cases are considered. In the first case, $Q(n)$ close to the probability $P_b$, i.e., the probability expressing the initially trained contextual conditions. This means that no model updating is required. The second case refers to the condition where the probability $Q(n)$ is far away from $P_b$. Then, the model is updated. For comparing the two distributions $Q(n)$ and $P_b$, the Kullback-Leibler divergence entropy [43] is adopted here.

Fig. 7.5 A graphical illustration of Bayesian optimization.

In statistics, the K-L divergence entropy is a measure indicating of how close a probability distribution is against a reference probability. A K-L divergence of zero indicates that the two distributions are identical. Instead, larger values of K-L entropy mean that the two distributions are quite different. The K-L entropy is defined as

$$D_{K-L}(P_b||Q) = \sum P_b log \left( P_b / Q(n) \right) \tag{7.6}$$

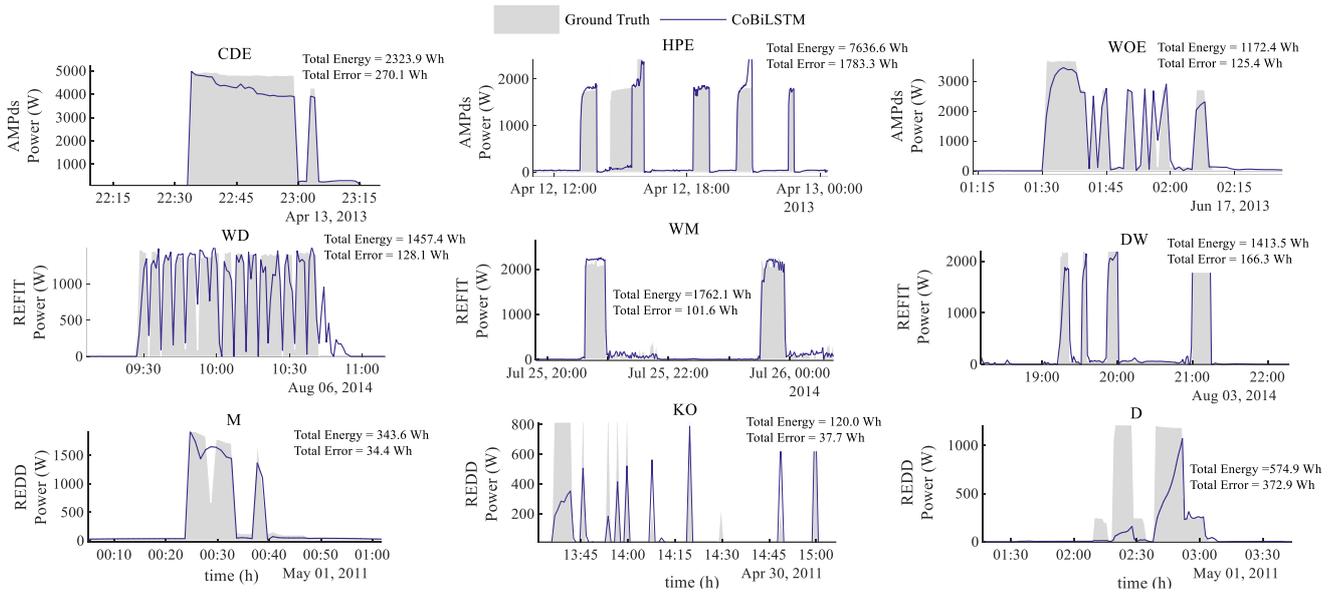

Fig. 7.9 Comparison of the proposed method (blue solid line) with ground truth (in gray) for selected appliances in AMPds, REFIT and REDD dataset. The actual total energy consumption and total error (between estimation and ground truth) for given time periods in every appliance are also depicted.

Eq. 7.6 implies that the distributions $Q(n)$ and $P_b$ are discrete, namely frequency distributions.

### 7.3.3 Context-Aware Adaptive Bayesian Optimization

In this Section, a context-aware adaptive Bayesian optimization strategy is used for optimally configuring the parameters of the bidirectional LSTM network. Bayesian optimization is a design strategy for optimizing an unknown, expensive and noisy objective function [129]. In particular, the Bayesian framework estimates a set of configuration parameters of the LSTM network, such as network depth and number of hidden units. More details about the configuration parameters can be found in Section 7.4. Let us denote by $\pi_j$ a vector that contains the configuration parameters as far as the model $g_j(\cdot)$ is concerned. Then, the goal of the Bayesian optimization strategy is to estimate an optimal configuration network $\pi_l^{(opt)}$ in a way that the performance error of the network is minimized.

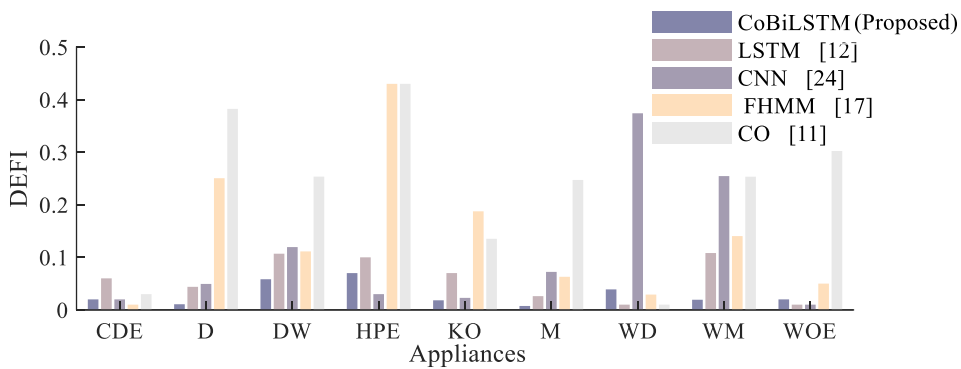

Fig. 7.10 Performance comparison of CoBiLSTM method with benchmarks LSTM, CNN, CO, FHMM for AMPds and REFIT (House 9) dataset.

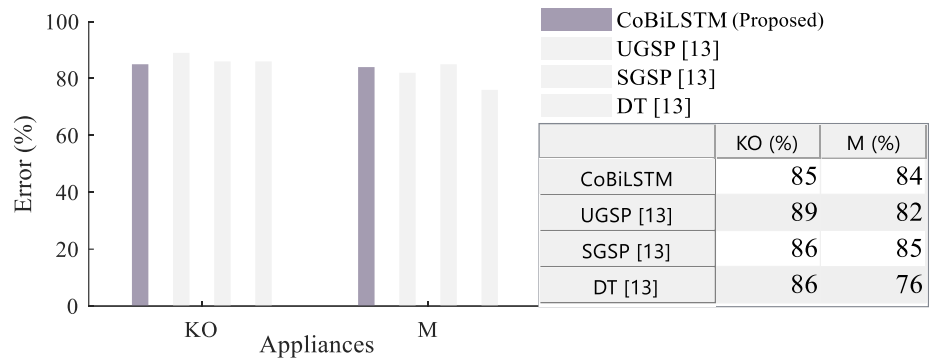

Fig. 7.11 Performance comparison of CoBiLSTM method with UGSP, SGSP and DT for REDD House 2.

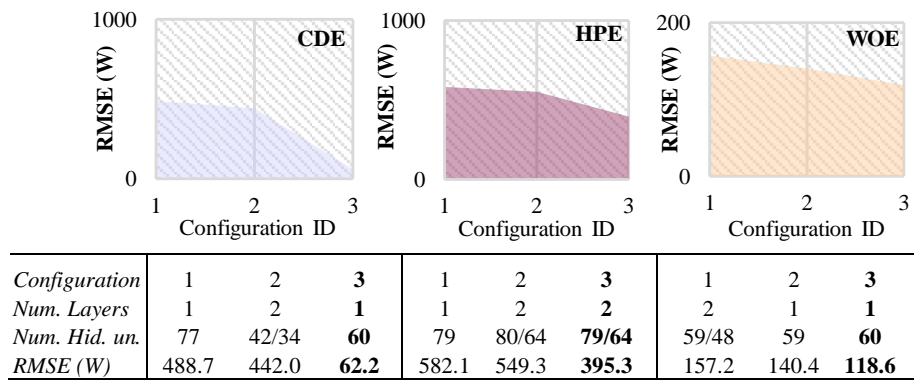

Fig. 7.12 Bayesian optimization results for CDE, HPE and WOE appliances of AMPds dataset. The y-axis shows the RMSE performance error for every different configuration set ID (1, 2, 3), as indicated in x-axis.

$$\pi_l^{(opt)} = arg \min_{\pi} E(\pi) \qquad (7.7)$$

where $E(\cdot)$ denotes the error of the estimated values as provided by the model $g_j(\cdot)$ and the ground truth data of the training and validation set. The main difficulty in solving Eq. 7.7 is that the error function $E(\cdot)$ is unknown. Therefore, a Bayesian optimization approach is considered, modeling the error $E(\cdot)$ as a Gaussian Process (GP). A GP models a distribution over function in a similar way that a Gaussian distribution models a distribution of a random variable and it is fully specified by the respective mean value and covariance [3].

Fig. 7.5 clarifies the way that the Bayesian optimization works, while more details can be found in [37]. In Fig. 7.5, we assume 1D configuration vector for visualization purposes. We have depicted 3 configuration values, namely $\pi_1, \pi_2, \pi_3$. Each configuration value corresponds to $E(\pi_i), i = 1, \ldots, 3$. However, the exact form of the function $E(\pi_i)$ is actually unknown. For this reason, we model the errors $E(\pi_i)$ as Gaussian distribution functions. In this particular example, we have three Gaussian distributions, each corresponds to one of the three configuration values $\pi_1, \pi_2, \pi_3$. These functions are depicted in Fig. 7.5 as dashed lines. Assembling, all these Gaussian distribution functions together, a GP is formed, as indicated with purple, Fig. 7.5. The scope of Bayesian optimization is to estimate a better configuration parameter from the current one. This means that it estimates the next configuration value, improving the error $E(\cdot)$. For example, let us assume that configuration value $\pi_2$ has been selected as the current optimal one. Then, a better choice for the Bayesian optimization strategy is the configuration $\pi_3$, since it provides a better mean value than the current one even when standard deviation divergence is considered. In general, there is a trade-off between the mean value and standard deviation, called Exploration-Exploitation trade-off in a Bayesian context. More details can be found in [130]. Following these steps, the optimal configuration parameters of the LSTM network are derived.

## 7.4 Experimental Evaluation

### 7.4.1 Dataset description and experimental setup

The evaluation of our context-aware bidirectional LSTM (CoBiLSTM) model for energy disaggregation has been conducted on three publicly available datasets; AMPds (a single Canadian house) [119], REFIT (21 UK houses) [131] and REDD (6 US houses) [112]. These open-access energy consumption datasets provide the aggregate power measurements of the whole house and sub-metered readings (smart plugs) from individual appliances at different time resolutions; 60s for AMPds, 6-8s for REFIT and 1s for REDD. The time duration of these datasets varies, covering two years for AMPds and REFIT and only a few weeks for REDD dataset. AMPds is selected, as it covers measurements from two years, so contextual variations (such as seasonal settings) are easily observed. On the other hand, REFIT is challenging, as its measurements contain noise due to numerous unknown appliances and measurements errors. Finally, REDD is selected as it is the first public available dataset released by MIT in 2011 and is widely used for NILM research. The proposed algorithm assumes low-sampling rate measurements so the measurements from REFIT and REDD are down sampled to 60s resolution. CoBiLSTM is implemented using MATLAB software. We trained our models using the adaptive moment estimation optimization algorithm (adam) with a learning rate of 10-4. Model weights and coefficients are updated using a mini-batch size of 50 samples at each training iteration. The maximum number of epochs for training is selected to be 400. Deep learning performance is improved through data balance and normalization of the data used as input to the CoBiLSTM network to the range [0, 1]. We have individual networks, one per appliance of the aforementioned datasets. The network configurations are optimally selected using the Bayesian optimization approach. When a statistical difference is detected, the energy disaggregation model is updated to fit the current contextual conditions, while retaining a satisfactory performance on the already gained knowledge.

## 7.4.2  Case Study 1: The effect of bidirectionality

In this study, we evaluate the effect of bidirectionality, that is the effect of future samples, on energy disaggregation. In particular, we compare the proposed method against traditional deep learning models such as CNNs [107] and unidirectional LSTMs [1], which receive as inputs only past time instances. Additionally, we compare the proposed method against additional approaches such as FHMM [132] and CO [97].

Fig. 7.6 presents the comparative results based on objective metrics of (i) Mean Absolute Error (MAE), (ii) Root Mean Square Error (RMSE) and iii) Normalized RMSE (NRMS), which are commonly used metrics for the evaluation of energy disaggregation [102]. The performance metrics are defined as follows:

$$MAE = \frac{\sum_n |\hat{p}_j(n) - p_j(n)|}{n} \tag{7.8}$$

$$RMSE = \sqrt{\frac{\sum_n (\hat{p}_j(n) - p_j(n))^2}{n}} \tag{7.9}$$

$$NRMS = \sqrt{\frac{\sum_n (\hat{p}_j(n) - p_j(n))^2}{\sum_n (p_j(n))^2}} \tag{7.10}$$

In this experimental setup, we have selected three appliances per dataset (AMPds, REFIT and REDD) for simplicity. Particularly, we have used (i) clothes' dryer (CDE), heat pump (HPE) and wall oven appliance (WOE) of AMPds, (ii) washer dryer (WD), washing machine (WM) and dishwasher (DW) of House 9 in REFIT and (iii) microwave (M), kitchen outlet (KO) and dishwasher for REDD (House 2) dataset. We select these appliances, as the most common in relevant studies focusing on deep learning low frequency NILM [99], [57]. As observed, the proposed method outperforms the compared ones. Our proposed CoBiLSTM as well as unidirectional LSTM generally perform best mainly due to their capability to effectively model long-range dependencies. Comparing these two, the proposed method attains the minimum error, since it is able to model non-causal behavior and has been optimized using the Bayesian framework. Selected appliances for REDD and AMPds dataset seem to have better performance than the case of the noisy REFIT dataset. It should be mentioned that the model for detecting the HPE appliance (AMPds) is not so accurate mainly due to recurring signal changes caused by external (seasonal) contextual conditions. It is important to note that in Section 7.4.4, we consider these external factors that cause signal appliances' changes, further improving the performance of our proposed model. It should be mentioned that for all scenarios the metrics have been calculated over the entire examined time period, in which the appliances can be either in operation or not. Treating the outcome of the model's network as a timeseries of the estimated active power, we extract some summary descriptive statistics to get an insight in models' performance, through the comparison of the appliance's estimated timeseries with the ground truth. Thus, Table I provides minimum, maximum and average difference between the estimated power and the real one, as well as the standard deviation.

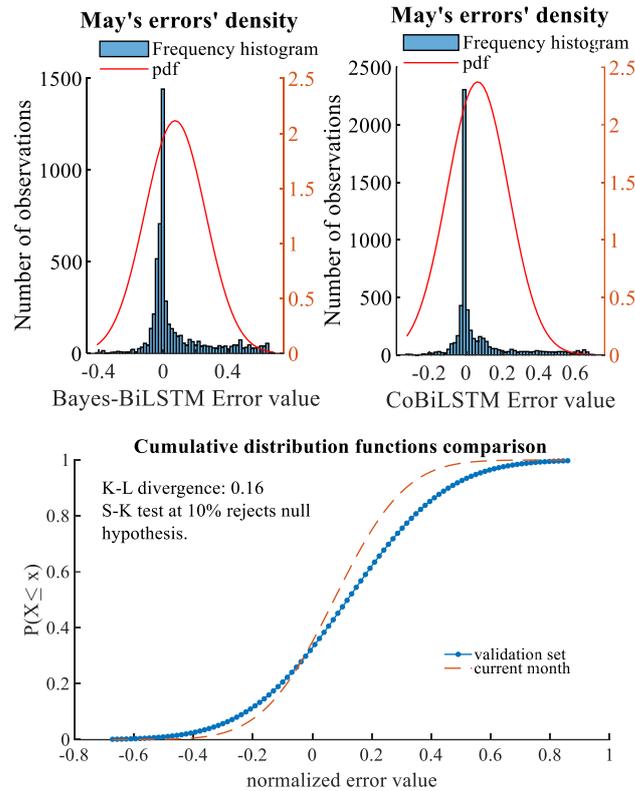

Fig. 7.13 Top: Performance error distribution for May with Bayes-BiLSTM [3] model (left) and CoBiLSTM model (right) for washer dryer. Bottom: the divergence K-L between distributions of May and the given distribution for the validation set.

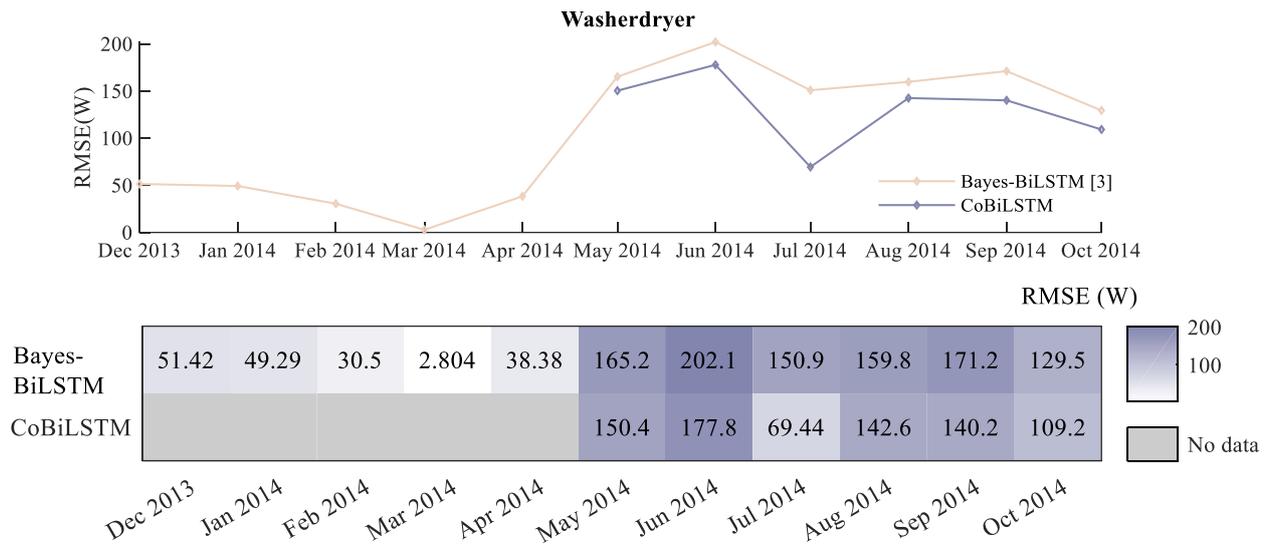

| | Dec 2013 | Jan 2014 | Feb 2014 | Mar 2014 | Apr 2014 | May 2014 | Jun 2014 | Jul 2014 | Aug 2014 | Sep 2014 | Oct 2014 |
|---|---|---|---|---|---|---|---|---|---|---|---|
| Bayes-BiLSTM | 51.42 | 49.29 | 30.5 | 2.804 | 38.38 | 165.2 | 202.1 | 150.9 | 159.8 | 171.2 | 129.5 |
| CoBiLSTM | | | | | | 150.4 | 177.8 | 69.44 | 142.6 | 140.2 | 109.2 |

Fig. 7.14 Comparison between the Bayes-BiLSTM approach [1] and CoBiLSTM models for washer dryer appliance.

| | | RMSE (W) | | |
|---|---|---|---|---|
| | Bayes-BiLSTM [3] | CoBiLSTM (Proposed) | CNN [24] | CoCNN |
| **Dehumidifier** | 66.7 | 61.8 | 211.5 | 123.0 |
| **Heat Pump** | 135.7 | 125.6 | 291.6 | 314.6 |
| **Tumble Dryer** | 94.0 | 80.2 | 142.6 | 146.6 |
| **Washer Dryer** | 112.0 | 95.0 | 162.9 | 158.4 |

Fig. 7.15 Performance error per appliance and per method (Bayes-BiLSTM, Co-BiLSTM, CNN, CoCNN).

The majority of the minimum values are close to zero. On the contrary, maximum values in some appliances are quite high. This is mainly because in some cases there is a delay between model estimates and ground truth label data, leading to high maximum deviations. However, the mean and standard deviation indicate that the proposed CoBiLSTM model is an adequate energy disaggregation estimator. It should be mentioned that the proposed method has one further advantage compared to FHMM and CO approaches. In particular, the FHMM and CO methods fail to provide adequate estimates as the number of the appliances under consideration increases, mainly due to modularity related issues. Instead, the performance of the proposed CoBiLSTM model is independent of the number of appliances considered. Fig. 7.7 shows the temporal behavior performance of the proposed model compared with unidirectional LSTM and CNN. As is observed, all models can correctly detect the time intervals when the target appliance consumes energy. However, LSTM and CNN methods fail to follow the exact consumption pattern of the appliance. For example, WOE appliance energy consumption pattern is detected with high accuracy using the CoBiLSTM approach. The LSTM model tries to follow appliance's pattern but underestimates the amount of power that the appliance reaches. The CNN model detects the existence of WOE appliance, but the method can't follow the energy consumption profile of WOE appliance. Additionally, Fig. 7.8 shows the real amount of energy that a specific appliance consumes at a given time period and the error (losses) of our CoBiLSTM deep learning model. In the following, we perform comparisons using the Estimated Energy Fraction Index (EEFI) and Actual Energy Fraction Index (AEFI) indicators which is used in the literature [102]:

$$EEFI(j) = \sqrt{\frac{\sum_n \hat{p}_j(n)}{\sum_n \sum_j \hat{p}_j(n)}} \tag{7.11}$$

$$AEFI(j) = \sqrt{\frac{\sum_n p_j(n)}{\sum_n \sum_j \hat{p}_j(n)}} \tag{7.12}$$

where we recall that $\hat{p}_j(n)$ and $p_j(n)$ is the estimated and ground truth power load for the j-th appliance respectively. Then, the absolute difference is exploited:

$$DEFI(j) = |EEFI(j) - AEFI(j)| \tag{7.13}$$

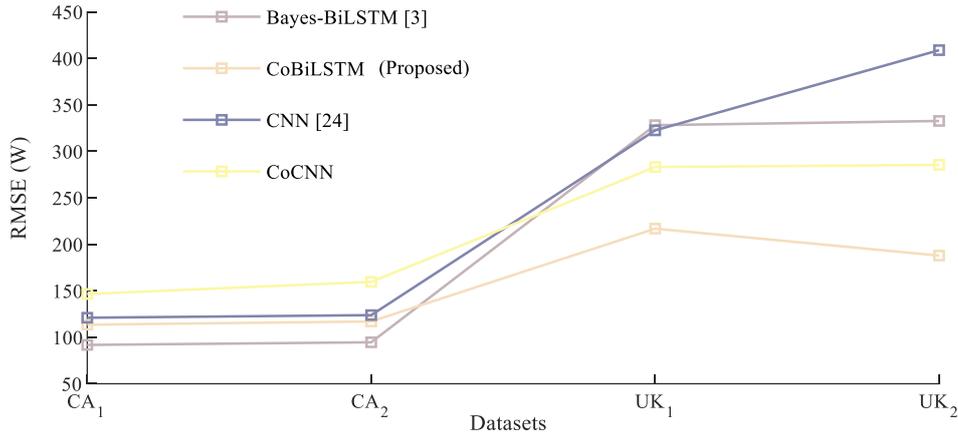

Fig. 7.16 Performance error for dishwasher appliance per method (Bayes-BiLSTM, CoBiLSTM, CNN, CoCNN).

This metric is depicted in Fig. 7.9 for the three selected appliances of the AMPds, REFIT and REDD dataset (i.e., total 9 appliances). In this figure, we also perform comparisons using other NILM methods (unidirectional LSTM, CNN, FHMM, CO), verifying that CoBiLSTM yields the minimum DEFI value (e.g., values close to zero). The performance of CoBiLSTM method is also compared with the state of the art GSP-based method of [7], [13]. Particularly, Fig. 7.10 shows the comparison of our method with algorithms of [107], [133], i.e. unsupervised GSP (UGSP), supervised GSP (SGSP) and decision trees (DT) as far as the Kitchen Outlet (KO) and Microwave (M) of REDD House 2. As is observed, our method presents almost similar behavior. It should be mentioned that the operation of these devices is not highly affected by seasonal (contextual) settings.

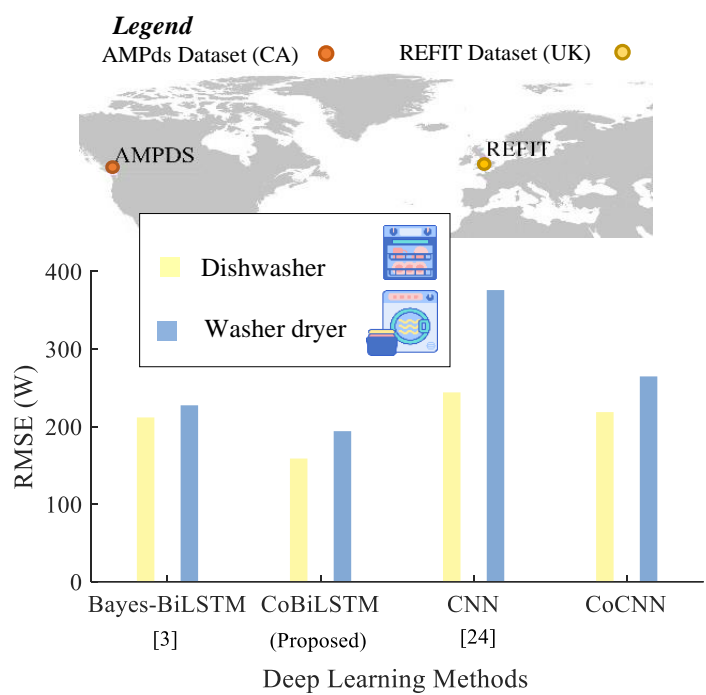

Fig. 7.17 Performance error for dishwasher and washer dryer appliances per method (Bayes-BiLSTM, CoBiLSTM, CNN, CoCNN).

### 7.4.3 Case Study 2: The effect of Bayesian optimization

Hyper-parameter tuning in a deep network is a major issue. In general, hyper-parameters are not optimized and are assumed to be fixed through time. However, seasonal attributes affect appliance electricity loads, influencing energy disaggregation performance. Bayesian optimization strengthens model performance through an optimal hyper-parameter selection. Fig. 7.11 shows the performance error (RMSE) of the three selected devices (CDE, HPE, WOE) of the AMPds dataset. In Fig. 7.11, we plot the RMSE performance error versus the number of iterations of the Bayesian optimization. As the number of iterations increases, RMSE is reduced. Similar results are derived for other devices of other datasets.

### 7.4.4 Case Study 3: The effect of seasonality

Consumers' energy consumption habits are characterized by heterogeneity across houses in different regions with varying weather conditions. The proposed CoBiLSTM model handles these heterogeneities through a dynamic learning process in which we update the basic parameters of the bidirectional LSTM model in a way to fit context changes (either internal or external). In this particular example, the effect of seasonality is validated. In our approach, the core idea lies in efficient triggering the update process, for the employed disaggregation networks. Trigger points are strongly connected to contextual factors (i.e., seasonal variations). The quantification of such factors is achieved by modeling the changes in model's performance (i.e. disaggregation error values). If the updated process is triggered, a new disaggregation network is created, using the same optimization approaches, over a renewed dataset. The process operates on a monthly base and evaluates the differences in error distribution functions. Every month, we fit a probability distribution function on the observed performance errors. Typically, we have a normal distribution, from which we can calculate the corresponding cumulative distribution function (cdf). Then, a comparison between the distributions, i.e. monthly errors and validation set (reference distribution) indicates whether these distributions are close enough (no need to update the model) or if those distributions are far away (a model updating is required). Two types of statistical tests were employed to estimate the proximity between the two compared distributions: Kullback-Leibler (K-L) divergence and Kolmogorov-Smirnov (K-S). K-S test, at a significance level of 10% (since variation in error is expected between months), evaluates the difference between the reference cdf distribution and the cdf of the current month, as illustrated in the bottom diagram in Fig. 7.12. If K-S test accepts the null hypothesis (the error distribution remains the same), there is no need for the model to be updated. On the other hand, that is i.e. K-S rejects the null hypothesis, we should probably update the model. However, K-S rejects too frequently the null hypothesis. Thus, the necessity for model's update is further investigated. In this case, K-L divergence is employed.

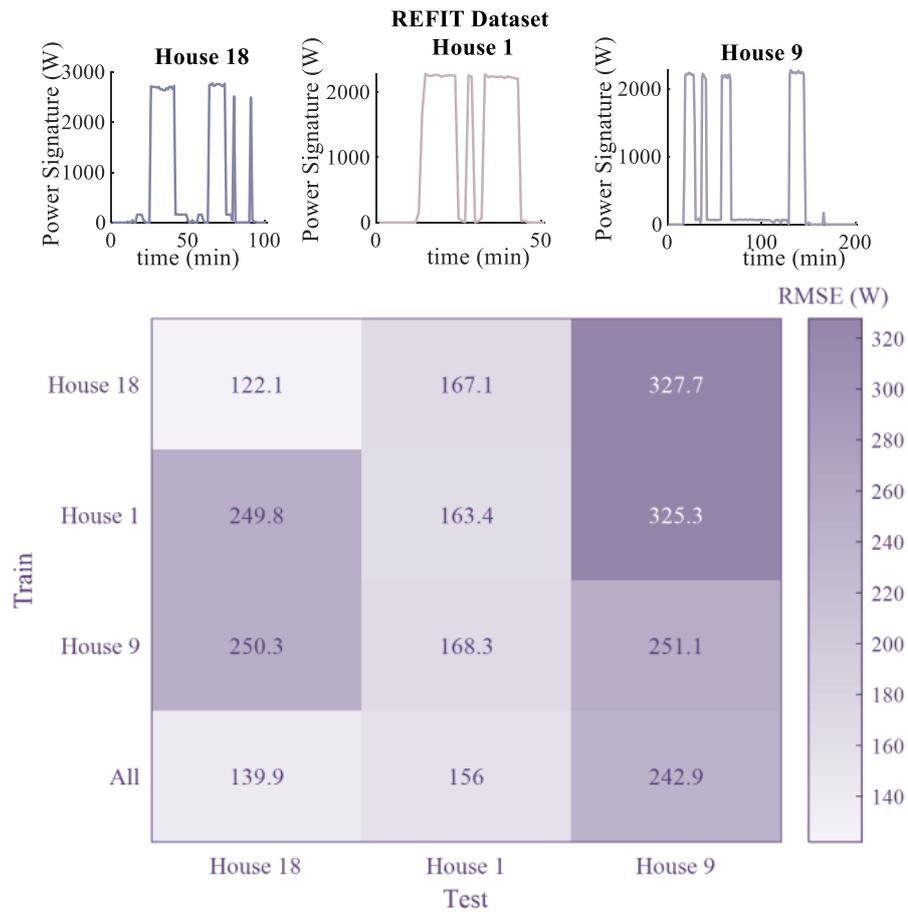

Fig. 7.18 Top: Differences in signature patterns for three different models of the dishwasher appliance, in three different houses. Bottom: Heat map showing the performance error for three different Bayes-BiLSTM models trained separately for each house and the integrated proposed CoBiLSTM model (last line of the colored table).

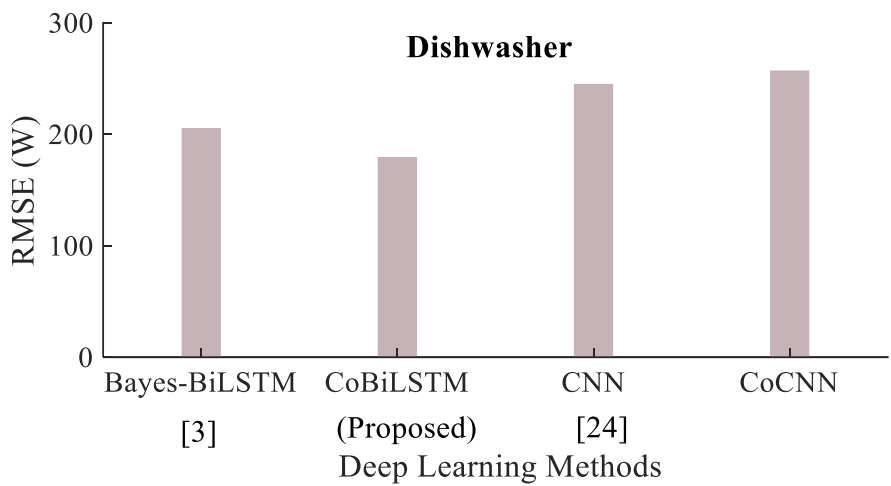

Fig. 7.19 Performance error for dishwasher appliance for the four deep learning methods (Bayes-BiLSTM, CoBiLSTM, CNN and CoCNN).

For example, in May, washer dryer's model performance function has changed, as Fig. 7.12 illustrates. K-S test rejects the null hypothesis and the K-L divergence score is 0.16. Thus, according to both tests, the applied disaggregation model should be updated. Consequently, we update the model, expanding the training dataset to include measurements of the month, in which difference in distribution of the performance error is observed. Fig. 7.13 shows the progress for the washer dryer RMSE performance error per month for a year, assuming a static model [1] and the improved performance achieved using the proposed Co-BiLSTM model. Furthermore, Fig. 7.14 shows the exact RMSE values as presented through a heat map to highlight the difference in performance between the approach of [1] and the CoBiLSTM model. The heat map is used here as a graphical representation tool of models' performance per month where the individual performance values are co-tained in the matrix and presented as colors of a gradient color palette. In case that the static model of [1] is considered, the performance for the next months will be worse in comparison to the proposed approach when dynamic model updating is carried out (Fig. 7.14).

Washer dryer, tumble dryer, dehumidifier and heat pump are typical examples of appliances that their usage patterns are highly dependent on external weather conditions. For those appliances, we proceed to a similar procedure as in the case of washer dryer. In Fig. 7.15, we compare our CoBiLSTM model performance with other methods such as CNN, as well as an improved version of it, called Context-aware CNN (CoCNN), where a similar approach for model updated is considered as in CoBiLSTM model. CoBiLSTM model has the minimum mean RMSE error over a period of about a year.

### 7.4.5 Case Study 4: The effect of geographical location

Here, we consider the transferability of our CoBiLSTM model against different regions. Thus, we evaluate our approach on two different datasets in different regions, namely the AMPds in Canada and the REFIT in UK. The experiment was conducted for dishwasher and washer dryer appliances. Firstly, we have our model trained in AMPds dataset. We check our model's performance for two different periods in time (CA1 and CA2) with mean RMSE error about 93.18 W for the dishwasher (Fig. 7.16). Having our model trained in AMPds dataset, we test its performance in the REFIT dataset, for the same appliance. The results are considerably worse (328.11W and 332.66 W for UK1 and UK2 different periods of time respectively). Then, we create an updated CoBiLSTM model that includes additional information regarding the houses in UK. We test our CoBiLSTM model performance separately in the two aforementioned datasets (AMPds and REFIT). The attained accuracy for AMPds dataset is now a little lower for the Canadian house (113.51W and 117.00W for CA1 and CA2) but is significantly higher in the case of the UK dataset (216.82W and 187.84W). We also compare our method with a CNN network, and its respective adaptive context-aware implementation (CoCNN). The CoCNN version has a context-aware adaptive character similar to our approach. The performance of the CoCNN model is improved compared to the CNN model, but it is still far from the performance achieved with the proposed CoBiLSTM model. It has been proved that CNNs work excellent with image data, but LSTMs can deal effectively with 1D signal data [134]. Fig. 7.17 shows the performance error regarding the effect of regionality. In particular, the work of [1] and the CNN methods have been statically trained using data from one geographical region, Canada, in our specific example. Instead, the CoBiLSTM and CoCNN models have dynamically changed their behavior to include new training samples for the new region of UK. As is observed, the proposed method outperforms the compared ones.

### 7.4.6 Case Study 5: The effect of appliance's operation mode

As observed (Fig. 7.18), different types of the dishwasher appliance, have different signature patterns. This is mainly due to the fact that an appliance operates at different modes (e.g. eco versus turbo mode). In this case, a model update is required to capture the new internal contextual operational attributes of the appliance. Our experiment uses three different types of dishwasher from three different houses (House 1, House 9 and House 18 of REFIT dataset). In the first row of heat map (Fig. 7.18), we have the model of [56] trained with dishwasher appliance of house 18 and test its performance in House 18 as well as Houses 1 and 9. The second row of the heat map tests model performance, using as training set the signature of dishwasher (House 9) and test in these three Houses. The third row also presents model performance in the three houses. The model is trained using dishwasher measurements of House 1. The last row presents CoBiLSTM model's performance on the three examined houses. The CoBiLSTM model in the last line uses as training set the total measurements of the three houses. Fig. 7.19 shows the RMSE error between the proposed CoBiLSTM model and the three other compared ones (the work of [1], a conventional CNN model and a context-aware CNN). As is observed, our proposed method outperforms the other compared ones, indicating that our method is capable of modelling the different internal contextual attributes of an appliance.

## 7.5 Conclusions

In this chapter, we propose a contextually adaptive and Bayesian optimized bidirectional LSTM model for modeling appliances' consumption patterns in a NILM operational framework. The proposed CoBiLSTM model is a generalized bidirectional LSTM network, adaptable to different contextual factors with an internal process of adaptation based on model performance error. The proposed approach based on progressive model updates can effectively adapt to a wider and more diverse range of instances and conditions, thus making a successful step towards scaled up NILM applicability. A bespoke, appliance-specific and specific to particular contextual configuration model would have better performance than our scalable context aware model, but it would only be applicable in a very constrained set of settings. The proposed model provides an updating solution that is sufficiently general, while at the same time managing to capture individual particularities in appliance operation. In our future work, we will further investigate the model's scalability and applicability to different contextual factors that constitute an obstacle to the solution of NILM problem, by leveraging a new NILM dataset across 3 countries (Greece, Austria and France) being currently created in the context of BENEFFICE EU project.

# Chapter 8

# The Proposed MR-TDLCNN model for NILM

## 8.1   The Proposed MR-TDLCNN model for NILM

The proposed methodology deals with the disaggregation problem by utilizing consecutive deep learning multi-input/multi-output regression models.

Let $M$ be the number of household appliances and $p(t_n)$ be the measured aggregate active power over all appliances at a time instance $t_n$. Considering a discrete time sampling, we can express $p(t_n)$ as $p(t_n) = p(nT) = p(n)$, where $T = t_n - t_{(n-1)}$ is the sampling interval. Similarly, we denote $p_j(n)$ the active power load of the $j$-th appliance out of the $M$ available. Then, the aggregate signal $p(n)$ can be given as [105]

$$p(n) = \sum_{j=1}^{M} p_j(n) + e(n) \tag{8.1}$$

where $e(n)$ denotes the additive noise of the measurements. In a NILM modeling framework, the measurements $p_j(n)$ are not available, since there are no smart plugs installed. Instead, only $p(n)$ is given. Therefore, the problem is to estimate $p_j(n)$ from $p(n)$. Let us note hereby that Table 8.1 includes the notations used in this chapter.

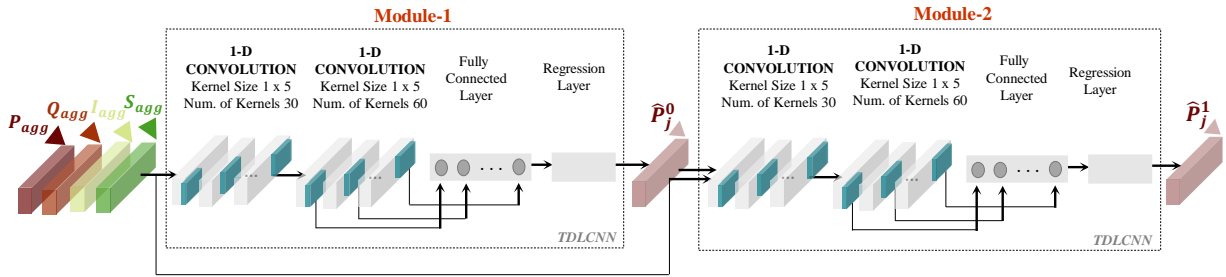

Fig. 8.1 The proposed non-linear pipelined recurrent CNN structure.

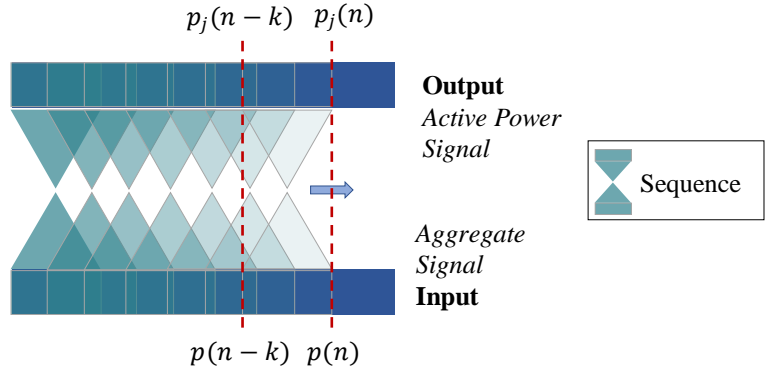

Fig. 8.2 The convolutional sequence to sequence NILM model. Time windows progress incrementally in time. The output is derived as the average of the different overlapping output values.

Table 8.1 List of the notations.

| Notation | |
|---|---|
| $j = 1, ..., M$ | number of appliances |
| $T$ | sampling interval |
| $t_n = nT = n$ | time instance |
| $p$ | active power |
| $r$ | reactive power |
| $q$ | apparent power |
| $I$ | current |
| $p(n)$ | aggregate signal |
| $p_j(n)$ | $j$-th appliance signal |
| $\hat{p}_j(n)$ | $j$-th appliance estimated signal |
| $e(n)$ | noise |
| $k$ | window of samples |
| $\hat{p}_j^0(n)$ | output of Module-1 TDLCNN |
| $\hat{p}_j^1(n)$ | output of Module-2 TDLCNN |
| $f(\cdot), g(\cdot)$ | non-linear function |

### 8.1.1 TDLCNN Model: Designing a baseline CNN for NILM

Each appliance has a unique spectral signature. This is the main principle we exploit to decompose the aggregate signal $p(n)$ into its components $p_j(n)$. Aggregate signal is actually derived as an integration of individual appliances' power consumption values over time. Thus, in order to get the estimates $\hat{p}_j(n)$ of $p_j(n)$, we need to assemble measurements of the aggregate signal $p(n)$ over a time window $[p(n)\, p(n-1) \ldots p(n-k)]^T$. Variable $k$ expresses the number of previous samples that should be considered for estimating the $p_j(n)$. The time window of the aggregate measurements (sequence) covers the mean appliance's operational duration in order to provide the full information regarding appliance's operational states for optimal feature maps selection later. Output sequences are created considering the same length T as in the input sequences, respectively (Fig. 8.2).

According to the already implemented approaches [1], sequence to sequence learning for NILM maps the input sequence of the aggregate signal to a same length output sequence of appliance's active power via LSTM networks. Here, we introduce an architecture based entirely on 1D convolutional neural networks that incorporates time series data. At instant $n$, the set of input to the module is a sequence of $k$ most recent measurements of aggregate power values which can be represented as vector $\mathbf{p}(n)$ given by:

$$\mathbf{p}(n) = [p(n)\, p(n-1) \dots p(n-k)]^T \tag{8.2}$$

proposing a Tapped Delay Line CNN model (TDLCNN). The purpose of the convolutional layer is to apply non-linear transformations on the input data to maximize regression performance. A set of parameterizable filters (e.g., learnable kernels) is convolved with the input data selecting appropriate feature modalities and estimating kernel parameters, so that performance error on a labeled training set is minimized. The L feature maps, say $f_1, f_2, \dots, f_L$, optimally selected by the convolutional layer, will be used as input to the final regression layer. The output is a sequence of the $j$-th appliance active power data, formed as:

$$\hat{\mathbf{p}}_j^0(n) = [\hat{p}_j^0(n)\, \hat{p}_j^0(n-1) \dots \hat{p}_j^0(n-k)]^T \tag{8.3}$$

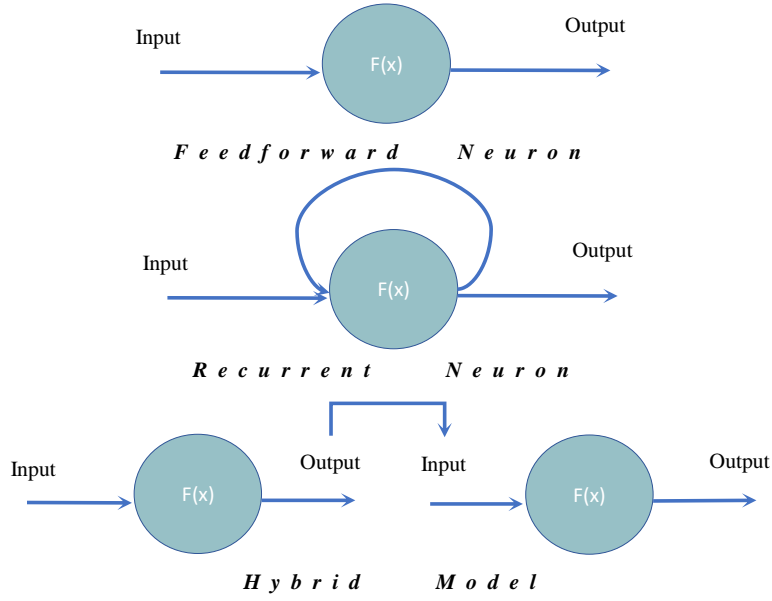

Fig. 8.3 Schematic illustration of the proposed hybrid model in conjunction to a simple feedforward and a recurrent neuron.

Therefore, we have that:

$$\hat{\mathbf{p}}_j^0(n) = g(\mathbf{p}(n)) + e(n) \tag{8.4}$$

where $g(\cdot)$ is a nonlinear relationship modeled by the learning process. As derived in this section, the first step to suitably decompose the aggregate signal to its components $p_j(n)$, is to consider several previous observations of the aggregate signal over a time window, in a way to maximize model's performance.

Table 8.2 Performance metrics (MAE, RMSE, NRMS) for AMPds dataset.

| | **CDE** | | | **DWE** | | | **HPE** | | | **WOE** | | |
|---|---|---|---|---|---|---|---|---|---|---|---|---|
| | MAE(W) | RMSE(W) | NRMS | MAE(W) | RMSE(W) | NRMS | MAE(W) | RMSE(W) | NRMS | MAE(W) | RMSE(W) | NRMS |
| TDLCNN | 19.2 | 103.7 | 0.2 | 32.0 | 84.6 | 0.8 | 102.2 | 102.2 | 1.1 | 42.2 | 201.3 | 1.3 |
| R-TDLCNN | 11.9 | 95.8 | 0.2 | 16.0 | 74.8 | 0.7 | 49.3 | 181.0 | 0.9 | 20.4 | 139.2 | 0.9 |
| M-TDLCNN | 16.3 | 65.3 | 0.1 | 23.9 | 71.7 | 0.6 | 24.9 | 65.9 | 0.3 | 27.0 | 99.1 | 0.6 |
| MR-TDLCNN | 4.7 | 41.6 | 0.1 | 14.3 | 64.5 | 0.6 | 18.4 | 58.3 | 0.3 | 15.2 | 76.68 | 0.5 |
| LSTM[103] | 25.3 | 180.9 | 0.4 | 24.0 | 72.1 | 1.0 | 161.7 | 369.8 | 0.5 | 15.8 | 141.2 | 0.8 |
| ConvLSTM[99] | 14.1 | 107.0 | 0.2 | 24.3 | 87.3 | 0.7 | 94.7 | 236.8 | 1.2 | 55.9 | 298.0 | 2.0 |
| CO[128] | 117.5 | 453.3 | 1.2 | 156.2 | 317.6 | 4.4 | 249.1 | 458.7 | 1.2 | 267.0 | 432.7 | 3.4 |
| FHMM[128] | 129.5 | 323.2 | 0.9 | 313.6 | 459.0 | 4.4 | 121.6 | 426.6 | 1.1 | 49.3 | 360.8 | 2.8 |

## 8.1.2 R-TDLCNN Model: Introducing a Recurrent character to CNN for NILM

It is intuitively clear that the active power signal observations per appliance are not independent over time. A widely accepted way to model this dependence and dealing with these inherently recursive data is through recurrent neural networks (RNN). RNNs can use the feedback connection to store information over time in form of activations, successfully handling sequential data and time series. On the other hand, CNNs capture patterns through non-linear relationships allowing weights to be dynamically updated in a complicated way. An approach based entirely on CNN is not adequate (see Section III.A), since significant variations in the disaggregated signal are observed that should be taken into consideration.

Thus, a hybrid CNN model is hereby introduced that incorporates RNN model's characteristics into the basic CNN structure, leading to a novel CNN model with a recurrent character, as illustrated in Fig. 8.3. The difference is that in the case of RNN model the update occurs in the hidden state, whereas in our approach the update is carried out directly to the output. Except for matching sequence inputs with sequence outputs, as introduced in the previous step (Section IV.A), the model incorporates an a posteriori state estimation per appliance, at time $n$, given observations up to and including time $n$.

Thus, based on (4) and given (2) and (3), we form an updated non-linear framework:

$$\hat{\mathbf{p}}_j^1(n) = f(\mathbf{p}(n), \hat{\mathbf{p}}_j^0(n)) + e(n) \tag{8.5}$$

where:

$$\hat{\mathbf{p}}_j^1(n) = [\hat{p}_j^1(n)\,\hat{p}_j^1(n-1)\ldots\hat{p}_j^1(n-k)]^T \tag{8.6}$$

The main difficulty in (5) is that the non-linear relationship $f(\cdot)$ is actually unknown. To address the fact, machine learning methods can be applied to approximate $f(\cdot)$ in a way that minimizes error $e(n)$.

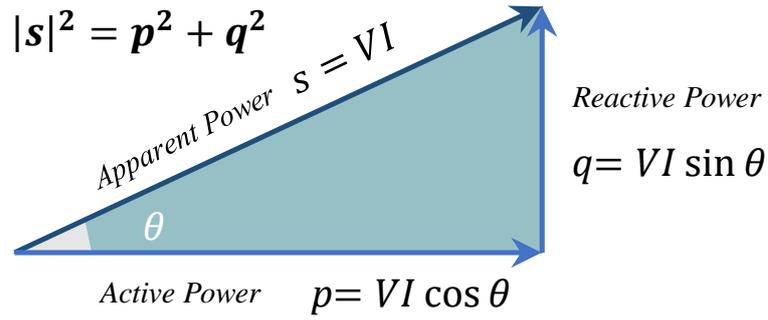

Fig. 8.4 The power triangle.

### 8.1.3 M-TDLCNN Model: Introducing a multi-channel CNN architecture for NILM

An important benefit of using CNNs is that they can support multiple inputs. In the literature, the majority of the proposed methods adopt a solution that uses only active power measurements. However, power utility companies are generally concerned about both active power ($P$) loads ($W$) and reactive power ($Q$) loads ($VAR$). In active power loads (e.g., an electric stove) dissipation of the performed worked takes place, whereas reactive power loads (e.g., a capacitor) store the power received from the grid, and release it back in the opposite direction later without dissipation. Loads can often be both active and reactive, as for example an air conditioning unit. Mathematically, active power results from in-phase voltage and current, whereas reactive power results from out of phase voltage and current [3]. Apparent power $S$, sometimes referred to as total momentary power, can also be a useful cue for disaggregation. Apparent power is conventionally expressed in volt-amperes (units in $VA$). These quantities are related as (Fig. 8.4):

$$s = I \cdot V \quad p = s \cdot \cos\theta \quad q = s \cdot \sin\theta \tag{8.7}$$

where $I$ is current, $V$ is voltage, and $\theta$ is the phase of voltage relative to current (i.e., the phase angle).

The $s, p, q$, and $I$ are inserted in the CNN model as multi-variable timeseries in order to strengthen the model's reliability, resulting in a non-linear regression problem. In correspondence to Section IV.A, each variable $(p, s, q, I)$ adopts the tapped delay line structure in order to be feed in the model. Each input sequence $(\{p\}, \{s\}, \{q\}, \{I\})$ is then, passed as a separate channel, in correspondence to different channels of an image (e.g. red, green and blue), forming a multi-channel CNN structure. As a result, a fused input that resembles a tensor is created. The tensorized input ensures that the model encapsulates all necessary information to produce the output, using a M-TDLCNN model.

### 8.1.4 MR-TDLCNN Model: Leveraging M-TDLCNN and R-TDLCNN models into a novel NILM model

The architecture, as shown in Fig. 8.1, consists of a pipelined recurrent structure. As shown in the figure, it is composed of two modules in parallel, TDLCNN Module-1 and TDLCNN Module-2, in such a manner that the output of Module-2 is used as an input to Module-1. MR-TDLCNN model combines simple recursive approach with CNN architecture in order to allow learning meaningful data-dependent weights. Furthermore, it exploits multiple input features succeeding high performance.

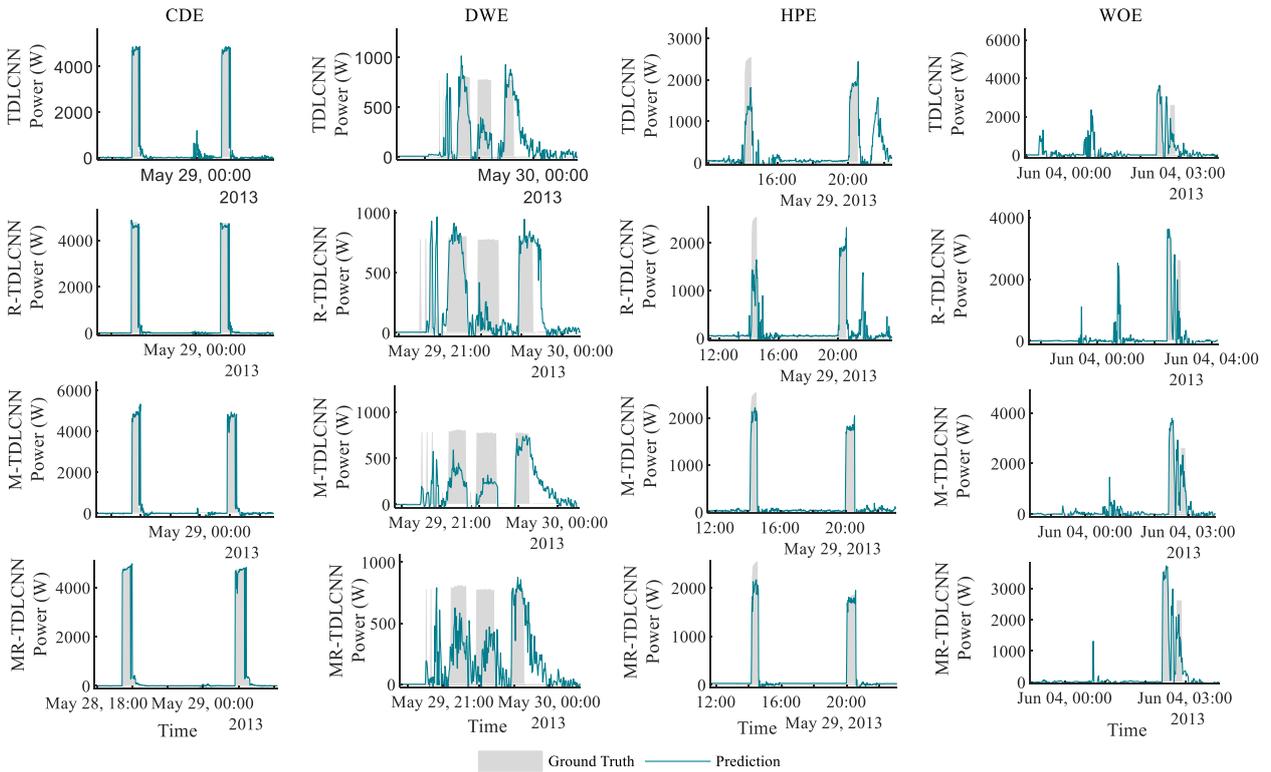

Fig. 8.5 Comparison of the proposed method (green solid line) with ground truth (in gray) for selected appliances in AMPds dataset.

## 8.2   EXPERIMENTAL EVALUATION

In this section, we will experimentally validate the superiority of the proposed MR-TDLCNN method in comparison to (i) the basic TDLCNN model, (ii) individual updates (R-TDLCNN and M-TDLCNN models) and, more importantly, (iii) other state of the art methods. Among them an LSTM network [103] and a hybrid CNN-LSTM method [99] are included. LSTM has been selected as a typical network for timeseries processing. Furthermore, an improved hybrid CNN-LSTM model [135] is used with CNN layers for feature extraction on input data combined with LSTMs to support sequence prediction. We also compare the aforementioned results with the state-of-the-art NILM algorithms i.e., FHMM-based and CO-based methods from NILMTK [128], a Python-based extension which is widely used in energy disaggregation research. We evaluate the model's accuracy and convergence speed across different models and different appliances. RMSE error of training was selected as a frequently used measure of accuracy in order to keep track of the performance measure of our model during training.

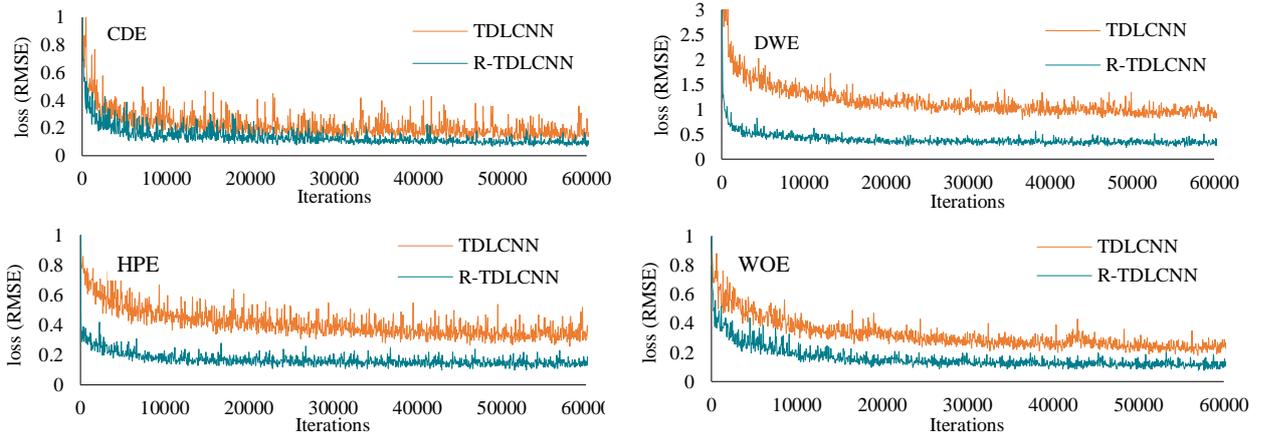

Fig. 8.6 Comparison of loss curves between the basic TDLCNN (orange line) and R-TDLCNN (green line) models.

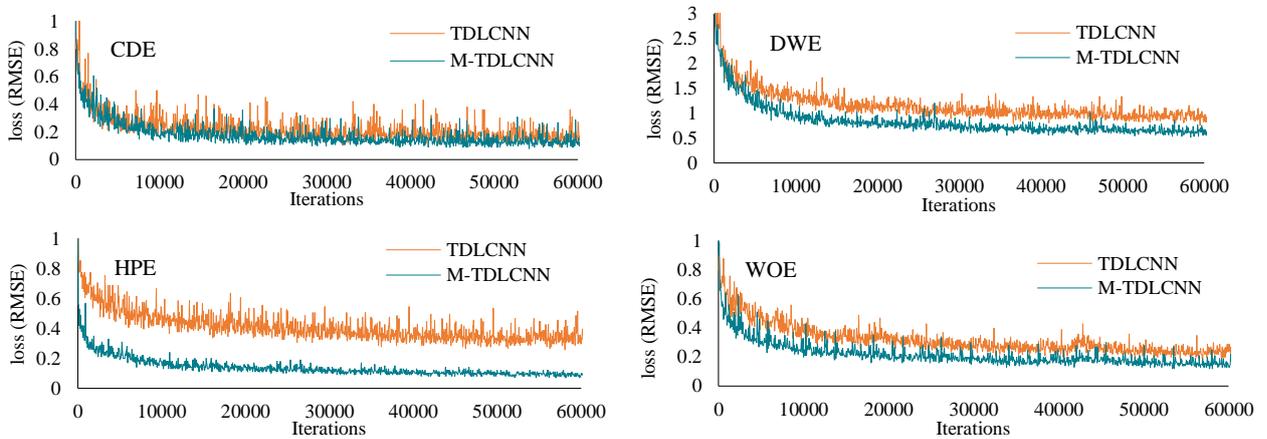

Fig. 8.7 Comparison of loss curves between the basic TDLCNN (orange line) and M-TDLCNN (green line) models.

### 8.2.1 Dataset description and experimental setup

The evaluation of the proposed method is conducted on the public AMPds dataset [116]. The AMPds contains active, reactive and apparent power values as well as current measurements from a Canadian house, at one-minute interval over a 2-year period. The selected appliances are of multi-state type, making difficult their respective power load estimation (see Fig. 8.5 with grey filled color).

### 8.2.2 Performance Evaluation and Comparisons

We trained our models using adam (adaptive moment estimation) optimization with a learning rate of 1e-4. Model weights and coefficients are updated using a mini-batch size of 50 at each training iteration. The maximum number of epochs for training is selected to be 400. Training period starts at 18 August 2012 and ends at 13 April 2013; 30 days were used as test sample (17 May 2013-17 June 2013). This split is representative of the problem and in addition, the testing period is a transitional period, so we can evaluate the ability of the model to adapt to seasonal variations.

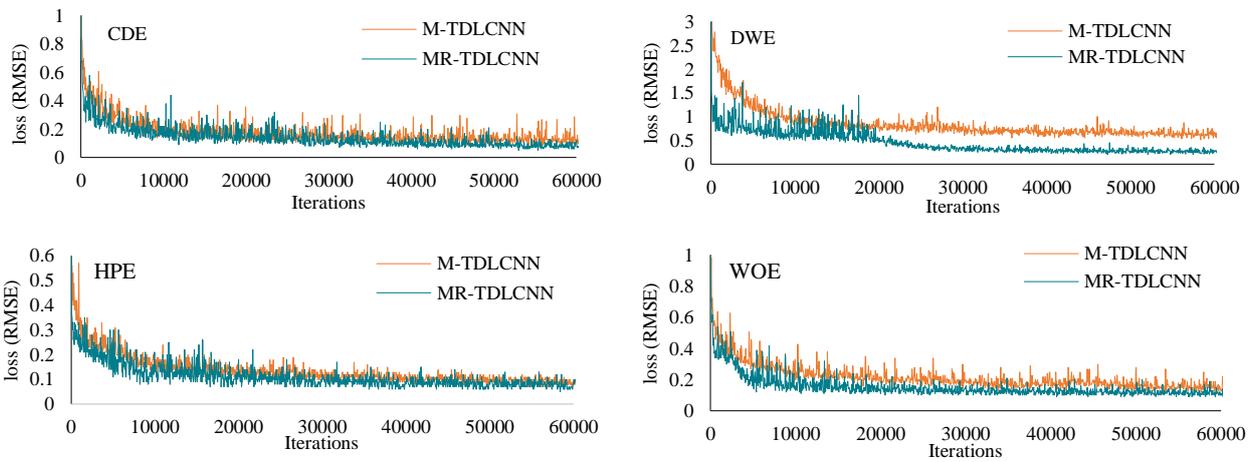

Fig. 8.8 Comparison of loss curves between the basic M-TDLCNN (orange line) and MR-TDLCNN (green line) models.

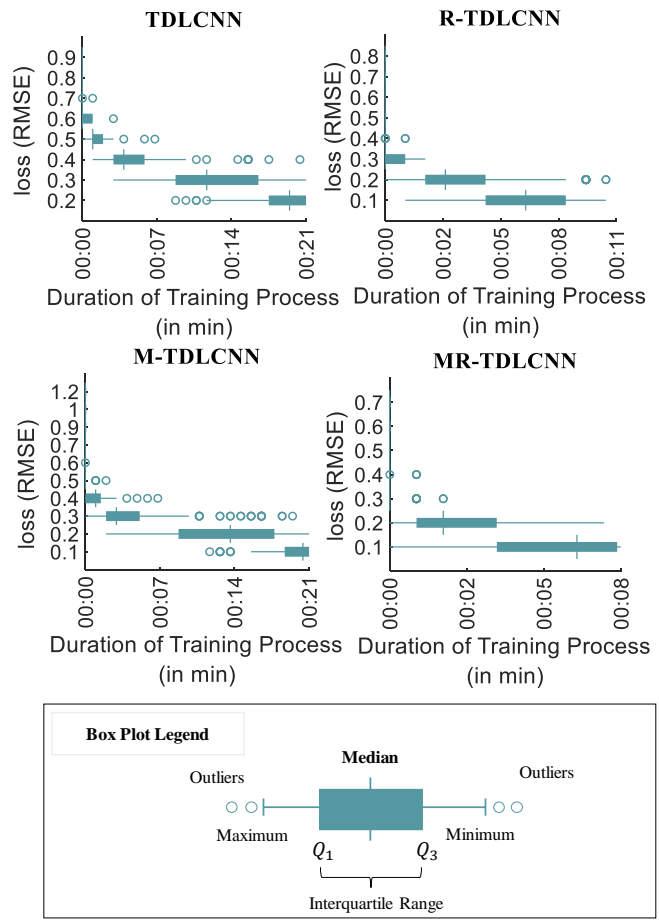

Fig. 8.9 The proposed model architecture. The model leverages the strength of rapid progress in CNNs, among with the need to apply the convolutional models for temporal sequences of data. Thus, a sequential input is entered into the model, a CNN layer is used for the optimal feature extraction and then, the stacked recurrent layers follow.

Deep learning performance is improved through data balance and normalization to 0-1 for each channel. The proposed MR-TDLCNN model, along with TDLCNN, M-TDLCNN, R-TDLCNN and LSTM are implemented using MATLAB software. CNN-LSTM algorithms have been trained and deployed using Python with Tensorflow and Keras libraries. CO and FHMM methods have implemented in NILMTK. Regarding the dataset, training and testing splits have already been pre-split and pre-normalized, to ensure that the conditions are the same and the results are comparable.

The proposed MR-TDLCNN regression model satisfies a set of crucial characteristics making it superior than the other existing methods in literature, for NILM. Its modularity is one of its main advantages in comparison to FHMM and CO approaches, in which dimensionality is a major issue. In addition, the introduction of deep learning as part of the solution of NILM problem is also a comparative advantage. Furthermore, model's performance strengthens with the use of all four components (current, active power, reactive power, and apparent power) available in AMPds dataset, achieving faster convergence and higher performance than state-of-the-art results for the same dataset.

Table 2 presents the comparative results based on objective metrics of (i) Mean Absolute Error (MAE), (ii) Root Mean Square Error (RMSE) and iii) Normalized RMSE (NRMS), which are commonly used metrics for the evaluation of energy disaggregation. In this experimental setup, four appliances for AMPds dataset are presented. Particularly, we have used clothes' dryer (CDE), dishwasher (DWE), heat pump (HPE) and wall oven appliance (WOE) of single AMPds house. Our proposed MR-TDLCNN method generally performs best mainly due to its capability to effectively model time dependencies and its ability to incorporate different data observations (p, s, q, I) strengthen model's performance. R-TDLCNN and M-TDLCNN models proved to have better performance compared to basic TDLCNN model. Here, it is worth mentioning that, CNN-LSTM model's performance is quite high and is a good alternative as a proposed solution to solve NILM problem. It should be mentioned that the model for detecting the HPE appliance (AMPds) is not so accurate mainly due to seasonal signal's changes caused by external contextual conditions / parameters. It should be mentioned that for all scenarios the metrics have been calculated over all the examined time period, in which the appliances can be either in operation or not.

Fig. 8.5 shows the comparison among the predicted signal and the ground truth for clothes dryer (CDE), heat pump (HPE) and wall oven appliance (WOE) of single AMPds House. Fig. 8.5 is representative of the MR-TDLCNN method's superiority against the remaining TDLCNN, M-TDLCNN and R-TDLCNN methods that have been presented before. Thus, the integrated MR-TDLCNN model succeeds better performance in comparison to the results that the basic model architecture succeeds. Particularly, the baseline TDLCNN model presents the worst results among others for all the presented appliances. Also, we can notice the existence of false detections, especially in HPE and WOE appliance. M-TDLCNN and R-TDLCNN models' have adequate performance, while MR-TDLCNN presents high-levels of performance and additionally, false detections have been eliminated.

A way to get insight into the model's learning behavior is through evaluation on the training dataset. Thus, a model's learning rate can be described using performance/epochs diagram. Fig.

8.6, 8.7, 8.8 illustrate loss curves for the four aforementioned models, namely are TDLCNN, R-TDLCNN, M-TDLCNN and MR-TDLCNN, in pairs.

Fig. 8.6 shows TDLCNN and R-TDLCNN models' loss progress during training using RMSE error and considering 60000 iterations per appliance for each of the presented CNN based models. In general, the loss function is being minimized during training. As observed, R-TDLCNN model's performance is slightly better than TDLCNN model's, as the former has a lower loss than the latter. It is worth mentioning that the training loss for HPE and DWE appliances present considerably lower values in R-TDLCNN than the values deriving from TDLCNN model. The loss curve, as illustrated with green for R-TDLCNN model, has lower starting point value, decreases with a smaller rate and is smoother than TDLCNN model's loss curve (with orange line).

Fig. 8.7 shows loss progress for basic TDLCNN and M-TDLCNN models during training. In general, the loss function is being minimized during training. In most cases, M-TDLCNN losses are lower compared to TDLCNNs ones. DWE appliance presents intense slope with high starting point values (3 for TDLCNN and 1.5 for M-TDLCNN) that converge at values <1. Finally, the loss curve for M-TDLCNN and MR-TDLCNN models is presented with orange and green color line, respectively, as Fig. 8.8 shows.

Taking as example WOE appliance, the box plot has been used to display the distribution of RMSE error reduction during training in conjunction to elapsed time, based on the five-number summary: minimum, first quartile, median, third quartile, and maximum. Furthermore, surprisingly high and low values called outliers, are illustrated by dots. The central rectangle spans the first quartile to the third quartile. A segment inside the rectangle shows the median and "whiskers" above and below the box show the locations of the minimum and maximum. Comparing TDLCNN and R-TDLCNN in Fig.9, we notice an increase in convergence speed leading to the reduction of needed time for training in comparison to the time needed for TDLCNN , even though the initial RMSE error (0.9) is greater than RMSE error of the simple CNN single channel model (0.8). Furthermore, TDLCNN succeeds RMSE error 0.2 while R-TDLCNN reaches the value of 0.1. Also, the presence of RMSE error value 0.1 starts early (the first minute of training time), even though as an "outlier". As regards the other two models, M-TDLCNN model reaches the value of 0.1 RMSE error, while MR-TDLCNN reaches the same value earlier and the training phase stops 10 minutes earlier.

# Chapter 9

# The proposed EnerGAN++ model for NILM

## 9.1 Motivation and Contribution

The main limitation of the aforementioned GAN-based NILM approaches is that the discriminator is represented by a generic classifier which has not been optimised for the specific particularities of energy disaggregation. More specifically, energy consumption follows long-range dependencies which cannot be approximated by the traditional shallow-based (short-term) discriminant components. Moreover, data consumption time-series of an appliance present non-linear auto-regressive behavior; the output values at a time instance is non-linearly related with the output values of previous time instances. Therefore, the discriminator classifier should simulate recurrent capabilities. Finally, a GAN model used for NILM modelling is implemented using the inverse GAN framework, since we need to approximate real energy consumption signals of an appliance instead of generating realistic outputs.

In this section, we extend the EnerGAN approach of [79] by including a deep learning classifier in the discriminator component of GAN [136], [137]. In particular, we enrich the concept of adversarial learning in NILM by introducing a more efficient discriminator in our proposed EnerGAN++ model which is now constitute of a combined convolutional layer with a recurrent GRU unit instead of a simple binary classifier. Advanced structures such as the recurrent GRU approximate long range recurrent dependencies in a better way, compared to traditional recurrent neural networks suffering from the vanishing gradient problem [138], [40]. In this work, we leverage the strengths of rapid progress in CNNs and the desire to apply these models to time-varying power consumption data sequences, under an adversarial training framework [139]. Here, we emphasize that alternative ways of applying CNNs for sequential data with temporal character for NILM have been proposed, such as [3]. However, none of the above mentioned methods that combine CNNs properties with temporal character have been used in adversarial learning for energy disaggregation. In particular, EnerGAN++ model has good performance, especially in case that (a) noisy aggregate signals are used as input triggers and (b) abrupt changes in the appliance energy signals are encountered.

We name this extended model, which is also based on an inverse GAN structure, EnerGAN++. Therefore, the main contributions can be summarized as follows:

- The use of a deep learning recurrent classifier to model the discriminator component of the EnerGAN++, with the capability to approximate long-term and regressive data signal behaviour, instead of the traditional GAN discriminators relied on shallow and generic classifiers.

- EnerGAN++ generates the entire signal waveform and it is capable of making more long-term estimations, rather than predicting solely the single current value of the individual power signal

- Our approach uses the aggregate energy measurements as the noise input vectors triggering the GAN model. In addition, since our target is to approximate real energy consumption of a specific appliance, EnerGAN++ uses labeled time-series of single appliance as input vectors of the generator during the training phase in contrast to conventional GAN modelling where only random noise signals are considered as input triggers. The non-linear deep discriminator unit competitively acts with the generator to reject generated sequences which are not well approximates of energy time-series of an appliance during training. Instead during testing, only the aggregate energy signal is used as input and the EnerGAN++ isolates the energy consumption sequence of a specific appliance from aggregate measurements.

- EnerGAN++ are inherently more robust to noisy input signals and less sensitive to inconsistencies or gaps in the input aggregate signal, since more sophisticated deep learning GAN components are considered.

The structure of the chapter is as follows: In Section 9.2 a detailed description of the NILM problem formulation is provided. Sections 9.3 and 7.4 describe the two EnerGAN++ network's components: the generator and the discriminator, explaining at the same time the challenges and barriers that we should overcome to adopt a GAN network for solving NILM problem as well as, the respective adaptations that EnerGAN++ introduces. Section 7.5 presents the proposed EnerGAN++ method configuration. In Section 9.6, the EnerGAN++ approach is experimentally evaluated against state of the art energy disaggregation methods in publicly available datasets, whereas Section 9.7 concludes the results.

## 9.2 The NILM Problem Formulation

The residential total power consumption is measured using smart meter devices, thus the consumers be aware of their total (aggregate) power consumption. However, as far as energy efficiency is concerned, energy consumption awareness at appliance level is essential. One way to measure consumption per appliance is through the usage of smart plugs, a solution which is economically unaffordable. For this reason, usually NILM or energy disaggregation methods are applied. NILM is the problem of decomposing the total power consumption of a household, into individual appliance

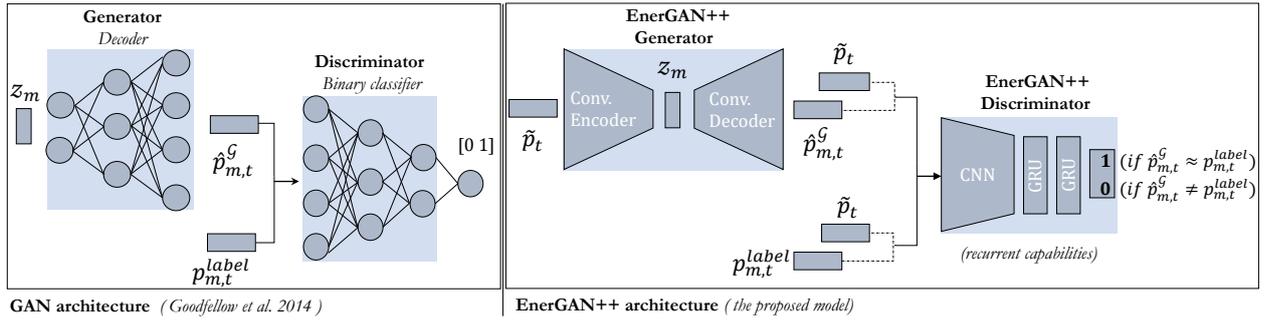

**GAN architecture** *( Goodfellow et al. 2014 )*     **EnerGAN++ architecture** *( the proposed model )*

Fig. 9.1 Comparison between a typical GAN approach and the EnerGAN+ structure approach. Three main specifications are introduced there: (i) the use of the aggregated signal as input noise, (ii) the encoding layer used to map the aggregated signal into a desired latent feature, and (iii) the advanced CNN-GRU discriminator dedicated to achieve the optimal sequence classification, that ensure the EnerGAN++ ability to achieve energy disaggregation, even if the data input samples are noisy.

power signal components, using signal processing and machine learning tools, without prior existence of smart-plug equipment.

At a discrete time index $t$, we assume $\tilde{p}(t)$ the noisy aggregate measured energy signal for the whole household under study. Signal $\tilde{p}(t)$ is the sum of the individual appliances' power consumption $p_j(t)$ plus an additional noise $\epsilon(t)$. Thus, in a NILM framework [97], we express the total power consumption $\tilde{p}(t)$ as:

$$\tilde{p}(t) = \sum_{m=1}^{M} p_m(t) + \epsilon(t) \tag{9.1}$$

In Eq. (9.1) variable $m$ refers to the $m$-th out of $M$ available appliances. Here, we need a robust to noise model able to separate the total noisy power measurements $\tilde{p}(t)$ into the individual -free of noise- appliance source signals $p_m(t)$. Under a NILM framework, the individual appliance power consumption $p_m(t)$ is not a priori available, assuming the absence of installed smart plugs. Instead, only $\tilde{p}(t)$ is given. Therefore, the problem is to calculate the best estimates $\hat{p}_m(t)$ of the appliance power consumption, given the noisy $\tilde{p}(t)$ values.

NILM methods often look at the problem as decomposing a mixture signal into individual appliances signals (based on single-channel source separation) and formulate the task as an optimization problem [see Eq. (9.1)]. Traditional generative models such as independent component analysis [140] and non-negative matrix factorization [117] have been proposed to solve the NILM problem. Linear independent component analysis (ICA) has been especially popular as a method for blind source separation (BSS), with applications in various domains including audio source separation and image processing. However, it would be interesting to replace the linear ICA model, with an alternative model to model the non-linearities existing in energy data consumption signals. In the light of the recent success of deep learning methods, auto-encoders (AEs) have been proposed as an approach to supervised non-linear source separation, by for example Pandey et al. [141] and Grais and Plumbley [142]. In the context of NILM, the work of Kelly [99] is one of the first works that have used denoising autoencoders.

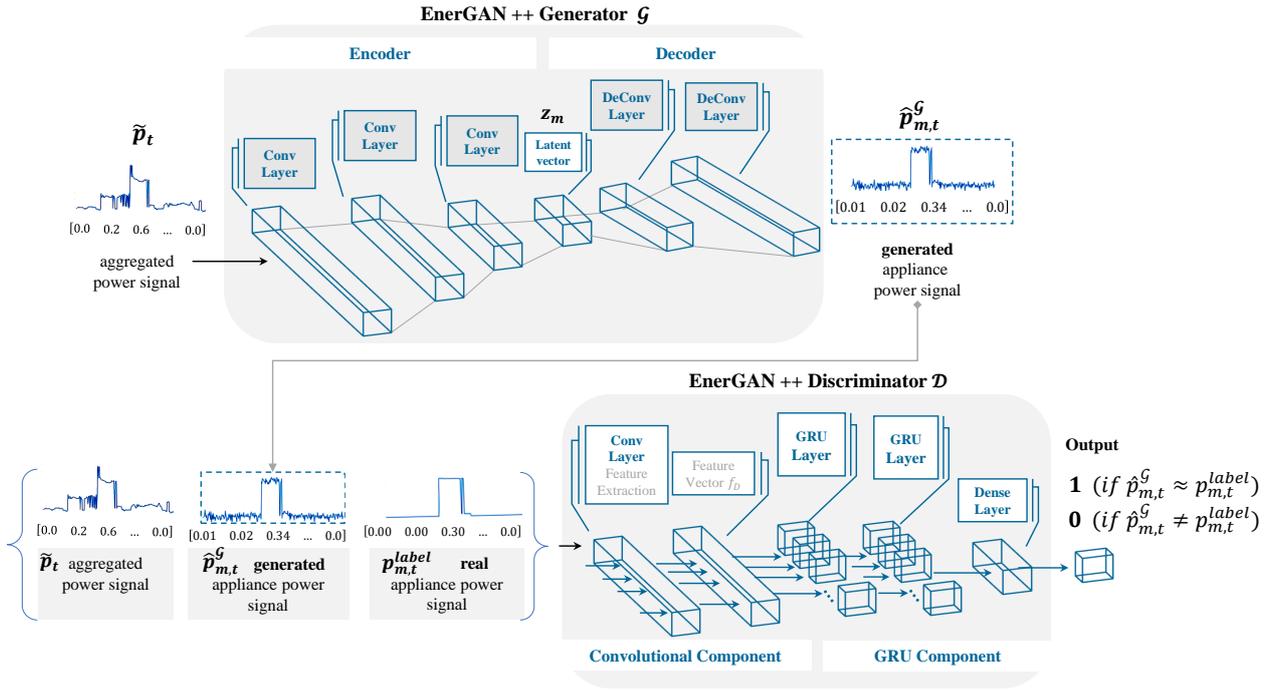

Fig. 9.2 The proposed EnerGAN++ architecture. The proposed method aims to unify existing AE and GAN architectures into a single framework, in which AE achieves a non-linear total power signal source separation, and adversarial training enhances model's robustness to noise. The generator is a convolutional autoencoder and as input has the noisy version of the aggregated power signal. The autoecoder is trained adversarially with the discriminator. The discriminator is a long-term convolutional recurrent network for sequence classification, and is conditioned with the aggregated power signal.

### 9.2.1 Adversarial learning in Energy Disaggregation - The EnerGAN++ approach

In general estimates of $\hat{p}_m(t)$ given as input the aggregate signal $\tilde{p}(t)$ is approximated through a non-linear relationship and therefore neural network structures are considered as universal approximators [143], [33]. However, unlike other deep learning neural networks that are trained with a loss function until convergence, a GAN model is trained in an adversarial manner between two main components: *the generator $\mathcal{G}$ and the discriminator $\mathcal{D}$*. In our NILM modelling, the *generator $\mathcal{G}$* is trained to produce time sequences resembling energy consumption of an appliance, while the *discriminator $\mathcal{D}$* is trained to check whether the produced time series by the *generator* coincide with a specific (real) appliance energy consumption sequence or not. Therefore $\mathcal{G}$ is trained in a way to cheat $\mathcal{D}$ in the sense that it produces data sequences that they are not be able to be distinguished by $\mathcal{D}$. In other words $\mathcal{G}$ and $\mathcal{D}$ play a two-play minmax game with value function of $V(\mathcal{D}, \mathcal{G})$

$$\min_{\mathcal{G}} \max_{\mathcal{D}} V(\mathcal{D}, \mathcal{G}) = E_{p_m \sim p_{data(p_m)}}[\log \mathcal{D}(p_m)] + E_{z \sim p_z(z)}[\log(1 - \mathcal{D}(\mathcal{G}(z))] \tag{9.2}$$

Eq. (9.2) means that the discriminator $\mathcal{D}$ is trained to maximize the probability of assigning the correct label between the ground truth samples and the "fake" samples produced by $\mathcal{G}$, while simultaneously $\mathcal{G}$ to minimize $log(1-\mathcal{D}(\mathcal{G}(z))$ that is the generator to produce samples of being indistinguishable by $\mathcal{D}$. The discriminator is trained to reject the artificially generated sequences by the $\mathcal{G}$ and the generator to produce sequences confusing $\mathcal{D}$.

A typical generative model takes a noise signal vector $z$ in space $R^k$ and generates another signal, $\mathcal{G}(z)$, of a higher dimensional space $R^n$ with $n > k$. The generative models produce data that resemble the real data [144]. Probabilistic generative models demonstrate excellent performance for various tasks such as denoising, inpainting, texture synthesis, video and natural language processing [145].

In our case, a typical generative model will try to successfully reproduce the power consumption signal of an appliance in households. In particular, in the following, we denote as $\{p^{label}_{m,t}\}^{t=t+T}_{t=t}$ the target signals over a time window $T$ and as $\{\hat{p}^{\mathcal{G}}_{m,t}\}^{t=t+T}_{t=t}$ the data generated by the generator $\mathcal{G}$ over the same time interval. A traditional GAN generator is optimized to create data sequence resembling the labeled data. This is achieved by propagating the noise signal into a higher dimension space through a neural network structure (i.e., a decoder) with the purpose of transforming the noise input $z$ into a data sequence $\{\hat{p}^{\mathcal{G}}_{m,t}\}^{t=t+T}_{t=t}$. This is illustrated in Fig. 9.1, where the noise input signal $z$ is forwarded into a convolutional decoder to generate a data sequence of $\hat{p}^{\mathcal{G}}_m$ of similar statistical properties with the labelled data $p^{label}_m$.

Even though the generated data $\{\hat{p}^{\mathcal{G}}_{m,t}\}^{t=t+T}_{t=t}$ are indistinguishable from the real data $\{p^{label}_{m,t}\}^{t=t+T}_{t=t}$, as they follow the same statistical properties, their resemblance to the actual power values is purely coincidental. Until now, we have underlined the GAN's ability to reproduce appliances power signals that resemble to the actual appliance operation power signal. However, solving NILM problem implies the need for a network trained not only to reproduce the power signal of an appliance but to know the exact operation and consumption of the appliance at a given time epoch. Thus, our method extends beyond the need for randomly generated appliance power consumption value having the same statistics. The trained network should be able to provide information regarding the estimated appliance power consumption value at a specific time, and this, in turn, means that $z$ values should not be purely randomly generated but to move in a space appropriate for extracting the desirable sequential data $\{p^{label}_{m,t}\}^{t=t+T}_{t=t}$ [146]. In case that the desirable output (i.e. the disaggregated appliance values) $\{p^{label}_{m,t}\}^{t=t+T}_{t=t}$ is a priori known, we need to solve the inverse problem, which corresponds to finding the latent vector $z$ that explains the output measurements as much as possible.

To address the aforementioned difficulties, we modify the traditional GAN configuration to fit the particularities of energy disaggregation problem. In energy disaggregation the purpose of a NILM-based GAN network is to generate real appliance power sequence which is almost identical with the labelled data over a time interval $T$, that is $\{\hat{p}^{\mathcal{G}}_{m,t}\}^{t=t+T}_{t=t} \approx \{p^{label}_{m,t}\}^{t=t+T}_{t=t}$ instead of producing realistic data sequences. Thus, it is important to find a way of inverting the "non-invertible" generator. This is achieved through the following modifications of the traditional GAN structure (see Fig. 9.1).

- First, instead of using a noise input signal $z$ the aggregate energy consumption signal $\tilde{p}(t)$ is considered. Aggregate signal $\tilde{p}(t)$, which is the sum of appliance's energy consumption signals, can be considered as an "noise" input trigger of the EnerGAN++ model. In this way, the generator $\mathcal{G}$ is capable of producing real and not realistic energy consumption sequences for the $m$-th appliance, confusing the discriminator $\mathcal{D}$.

- Second, in case that the aggregate signal $\tilde{p}(t)$ is used as input trigger of the proposed GAN-based model for NILM, an additional layer is required for the generator $\mathcal{G}$. In particular, we add an encoder prior to the decoder with the main purpose of compressing the input aggregate signal $\tilde{p}(t)$ in a way to produce a noise vector resembling $z$ of the traditional GAN structure.

- Third, the traditional GAN models considers simple classification structure for the discriminator $\mathcal{D}$. Energy consumption signal presents high temporal relationships and auto-regressive properties. Therefore, the traditional GAN discriminator can be easily cheated by the generator, reducing the overall NILM performance. For this reason, the proposed EnerGAN++ model modifies the conventional GAN discriminator by using GRU unit which has recurrent capabilities in order to address the temporal appliance's energy consumption properties. EnerGAN++ has a more clever discriminator, forcing the generator to produce almost identical appliance energy consumption to cheat $\mathcal{D}$.

## 9.3 The EnerGAN++ Generator

Fig. 9.2 presents the structure of the generator of the proposed EnerGAN++ model. Our method extends the traditional operation of a GAN beyond the need for randomly generated appliance's power data sequence of the same statistics with the labeled data. The NILM framework implies the need for a network trained to generate the exact operation and consumption of an appliance at a given time instance $t$.

This is achieved by using as input trigger of the EnerGAN++ the aggregate signal $\{\tilde{p}_t\}_{t=t}^{t=t+T}$ over a time window of $T$ duration. The signal $\{\tilde{p}_t\}_{t=t}^{t=t+T}$ is considered as a noise trigger since it is the summation of independent energy consumption signals of the appliances. In addition, during training, the ground truth data of the $m$-th appliance $\{p_{m,t}^{label}\}_{t=t}^{t=t+T}$ over a time window of $T$ duration is considered in order to initiate the EnerGAN++ generator to simulate the real energy consumption data of the $m$-th appliance. We denote as $\mathcal{I}_{train}$ the input trigger vector of the $\mathcal{G}_{EnerGAN++}(\cdot)$ generator during the training phase,

$$\mathcal{I}_{train} \equiv \left[ \{\tilde{p}_t\}_{t=t}^{t=t+T} \ \{p_{m,t}^{label}\}_{t=t}^{t=t+T} \right]^T \tag{9.3}$$

To handle the aggregate signal $\{\tilde{p}_t\}_{t=t}^{t=t+T}$ and the ground truth labels of $\{p_{m,t}^{label}\}_{t=t}^{t=t+T}$ as input trigger of the EnerGAN++ model, an encoder layer is added prior to the decoder. The encoder generates a compressed "noise" signal $z_m$ (by encoding the input vector signal of $\mathcal{I}_{train}$) [99] which is used as input trigger of the decoder of the EnerGAN++ generator to produce a real appliance energy consumption time series $\mathcal{G}(z_m)$ for the $m$-th appliance. Thus, the pipeline of the EnerGAN+ generator during the *training phase* is the following:

$$\textit{the pipeline of EnerGAN++ generator during training:}$$
$$\mathcal{I}_{train} \to Encoder\,(\mathcal{I}_{train}) \to z_m \to Decoder\,(z_m) \to \mathcal{G}_{EnerGAN++}(z_m) \tag{9.4}$$

Eq. (9.4) means that the input signal $\mathcal{I}_{train} \equiv \left[\{\tilde{p}_t\}_{t=t}^{t=t+T} \{p_{m,t}^{label}\}_{t=t}^{t=t+T}\right]^T$ is transformed (compressed) to a latent noise trigger $z_m$, through a convolutional encoder and then, the noise signal $z_m$ is decompressed to generate a signal $\mathcal{G}_{EnerGAN++}(z_m)$ that resembles the real energy consumption of the $m$-th appliance. The encoder with convolutional layers, is forced to be an inverted version of the decoder (with transposed convolutional layers), where corresponding layers perform opposite mappings and share parameters [147]. The model tries to minimize the difference between the $\{p_{m,t}^{\mathcal{G}}\}_{t=t}^{t=t+T}$ sequence values and the actual sequence values $\{p_{m,t}^{label}\}_{t=t}^{t=t+T}$, generating data sequences $\mathcal{G}_{EnerGAN++}(z_m)$ that confuse the discriminator $\mathcal{D}$.

During *the testing phase*, the generator of the EnerGAN++ model $\mathcal{G}_{EnerGAN++}(\cdot)$ receives as input only the aggregate signal $\{\tilde{p}_t\}_{t=t}^{t=t+T}$ and not the appliance ground truth data of $\{p_{m,t}^{label}\}_{t=t}^{t=t+T}$ since it has been learned during the training phase to produce identical energy consumption time series of the $m$-th appliance. Thus, the pipeline of the EnerGAN++ generator $\mathcal{G}_{EnerGAN++}(\cdot)$ during *the testing phase* is the following:

$$\textit{the pipeline of EnerGAN++ generator during testing:}$$
$$\{\tilde{p}_t\}_{t=t}^{t=t+T} \to Encoder\,\left(\,\{\tilde{p}_t\}_{t=t}^{t=t+T}\,\right) \to \hat{z}_m \to Decoder\,(\hat{z}_m) \to \mathcal{G}_{EnerGAN++}(\hat{z}_m) \tag{9.5}$$

It is clear that the output of the EnerGAN++ generator during the testing phase approximates its output during training since the ground truth data of the $m$-th appliance are available only during training. Therefore, we have that $\mathcal{G}_{EnerGAN++}(\hat{z}_m) \approx \mathcal{G}_{EnerGAN++}(z_m)$, meaning that the propoduce data time series by the generator $\{\hat{p}_{m,t}^{\mathcal{G}}\}_{t=t}^{t=t+T}$ is very close to the labeled data of $\{p_{m,t}^{label}\}_{t=t}^{t=t+T}$

## 9.4 The EnerGAN++ Discriminator

In the EnerGAN++ model there is not a loss function for training the network until convergence. Instead, adversarial learning is adopted based on a min-max two-player game between the generator $\mathcal{G}_{EnerGAN++}$ and the discriminator $\mathcal{D}_{EnerGAN++}$. The generator produces a sequence of data $\{p_{m,t}^{\mathcal{G}}\}_{t=t}^{t=t+T}$ to cheat the discriminator, while the discriminator seeks to distinguish whether the data sequence produced by $\mathcal{G}$ is real (assigned a value of 1) or fake (assigned a value of 0). The traditional discriminator of a GAN network does not take into account the regressive temporal characteristics of the energy consumption signals. Therefore, it can be easily cheated by the generator which has been appropriately modified to produce almost identical time series for the $m$-th appliance (see Section 9.3).

# GRU Cell

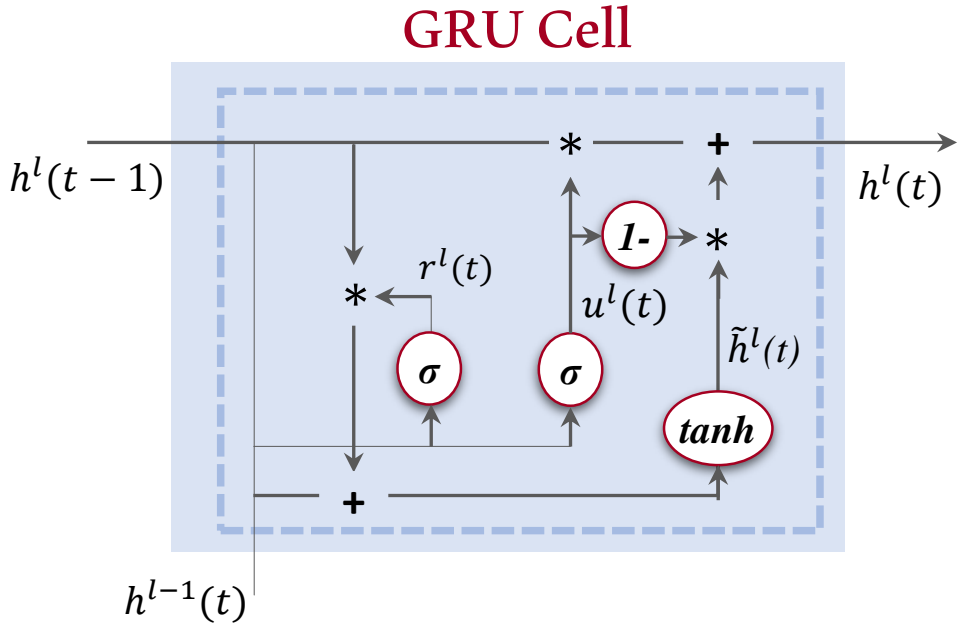

Fig. 9.3 The computation of a hidden state in a GRU cell.

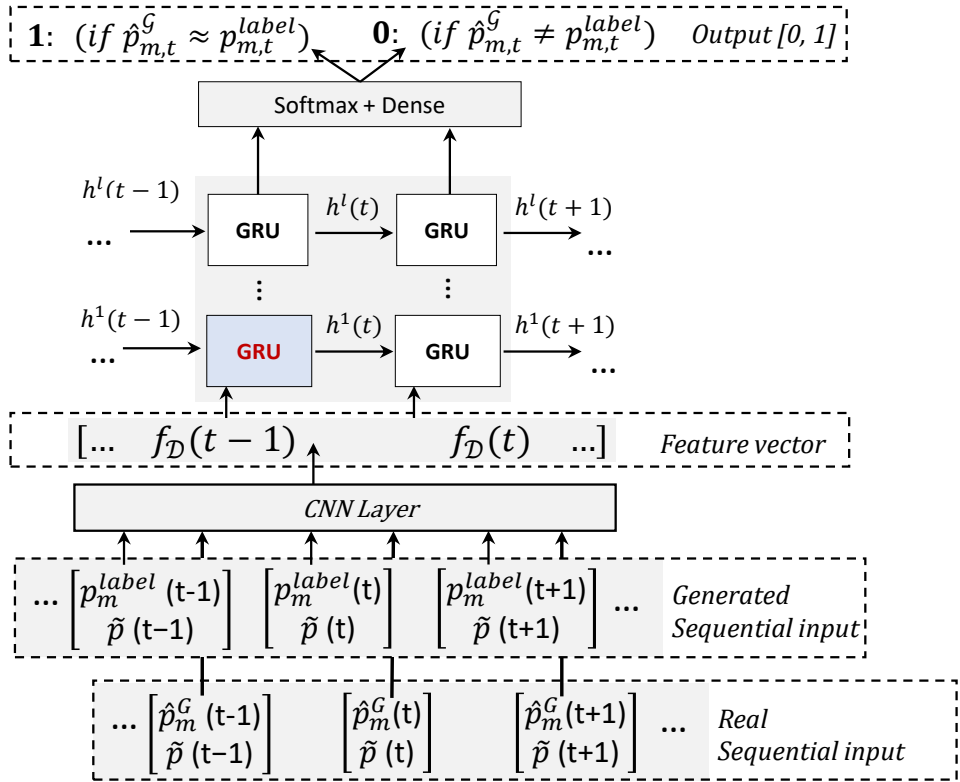

Fig. 9.4 The discriminator is a long-term convolutional recurrent network for sequence classification. The proposed architecture of the discriminator leverages the strength of rapid progress in CNNs, among with the need to apply the convolutional models for temporal sequences of data. Thus, a sequential input is entered into the discriminator, a CNN layer is used for the optimal feature extraction and then, the stacked recurrent layers follow. The scope of the discriminator is to decide whether the given sequence is the generated or the real one.

Here, the discriminator solves a binary sequence classification problem. Sequence classification is posed as a problem of assigning a label to a sequence of observations. Recently deep learning reccurent networks have been proposed in the literature for sequence classification. Examples are the LSTM [40] and the GRU [148] structures. These networks have an internal mechanism to balance between current and previous time steps, and thus, memorizing the temporal information flow.

### 9.4.1 CNN Enriched - Gated Recurrent Networks for EnerGAN++ Discriminator

In the EnerGAN++ model, a combined CNN enriched - GRU classifier is adopted as the discriminator unit. GRU networks are appropriate for modelling the temporal auto-regressive properties of a time series (Fig. 3). However, GRU structures are not able to extract features from the input data in a way to optimize the overall classification performance. For this reason, we adopt a combined approach by introducing a CNN model [36] prior to the GRU framework. Hereby, CNNs have been used as feature extractors for the GRUs. In other words, the combination of CNN as an efficient feature extractor with the GRU model, is capable of representing, synthesize and therefore distinguish the temporal dynamic nature of the power sequence signals.

GRU is a modern version and very similar to the traditional LSTM cell, albeit less complex. It merges the cell state and hidden state of the traditional LSTM cell, and in addition, it combines the forget and input gates into a single update gate. Thus, it is more computational efficient as it need less parameters for training and it has less complex structure. GRUs make use of a gated activation function and are designed so as to have more persistent memory thereby making it easier to capture long-term dependencies.

In the proposed EnerGAN++ diciminator, a stack of $l$ GRU layers is considered. In general, stacking improves discrimination performance [149]. Stacked GRU layers have two main operations of dependence; *the in-depth dependence* and *the temporal dependency*. The in-depth dependence implies that the output of the current GRU layer is related with the output of the previous layer, while temporal dependency assumes that the hidden GRU states in a layer are inter-related in the time domain. Each GRU unit has two additional control variables; the reset gate $r^l(t)$ and the update gate $u^l(t)$ (see Fig. 9.3). The reset gate $r^l(t)$ is responsible for determining how much of information to forget. The update gate $u^l(t)$ is responsible for determining the worth-remembering information of the previous states that should be forwarded to the next state. Therefore, the gates $r^l(t)$ and $u^l(t)$ are related with the hidden states $h^l(t)$ and $h^l(t-1)$ as follows

$$\begin{bmatrix} u^l(t) \\ r^l(t) \end{bmatrix} = \begin{bmatrix} \sigma \\ \sigma \end{bmatrix} \mathbf{W}_l \begin{bmatrix} h^l(t-1) \\ h^{l-1}(t) \end{bmatrix} + \begin{bmatrix} b^l_u \\ b^l_r \end{bmatrix} \tag{9.6}$$

In Eq. (9.6) $\sigma$ is the sigmoid function and $b^1_u$ and $b^1_r$ are the respective biases of each component for the GRU cell. Variables $\mathbf{W}$ and $\mathbf{U}$ are the transition matrices of the $l$-th GRU. In case of $l=1$, the convolutional layer is considered as previous layer and the input vector is the feature vector [see Eq. (9.9)]. This means that $h^{l=0}(t) \equiv f_{\mathcal{D}}(t)$.

In stacked GRUs configuration a recursive approach is considered as regards the operation of each GRU cell. In particular, a new memory state, denoted as $\tilde{h}^l(t)$, is considered, acting as the consolidation of the hidden state of the previous layer $h^{l-1}(t)$ and the previous hidden state $h^l(t-1)$ of the current layer. The consolidated hidden state $\tilde{h}^l(t)$ is given by:

$$\tilde{h}^l(t) = \tanh\left(r^l(t)\mathbf{U}h^l(t-1) + \mathbf{W}h^{l-1}(t)\right) \tag{9.7}$$

In Eq. (9.7) function $tanh(\cdot)$ refers to the hyperbolic tangent relationship. As we stated previous, for $l = 0$ the value of the hidden state equals $h^{l=0}(t) \equiv f_{\mathcal{D}}(t)$ (see Fig. 9.4). Eq. (9.7) means that the consolidated state is related with the output of the hidden state $h^l(t-1)$ at the time instance $t-1$ and the output of the previous hidden layer $h^{l-1}(t)$ at the time instance $t$. Using the values of the consolidated state $\tilde{h}^l(t)$ [see Eq. (9.7)] and the values of the update gate $u^l(t)$ [see Eq. (9.6)] the value of the hidden state of the $l$-th GRU element is estimated

$$h^l(t) = \left(1 - u^l(t)\right)\tilde{h}^l(t) + u^l(t)h^l(t-1) \tag{9.8}$$

In more detail, the GRU related above mentioned operations are illustrated in Fig. 9.4.

### 9.4.2 Operation of the EnerGAN+ discriminator

The proposed discriminator has two main components; *the convolutional* layer and the *GRU* unit. The convolutional layer transform the input signal to a reliable feature vector $f_{\mathcal{D}}(t)$, while the GRU unit performs the discrimination. As input signal the generated power signal $\mathcal{G}_{EnerGAN++}(\hat{z}_m)$ of the $m$-th appliance is used [see Eq.(9.5)]. In addition, the labeled training samples of the respective appliance $\{p_{m,t}^{label}\}_{t=t}^{t=t+T}$ and the aggregate measurements $\{\tilde{p}_t\}_{t=t}^{t=t+T}$ is used as input triggers for classification comparisons. The discriminator has been optimized through training to distinguish the "fake" data sequences produced by the generator $\{\hat{p}_{m,t}^{\mathcal{G}}\}_{t=t}^{t=t+T}$ from the real ones.

Initially, the input vector of the discriminator, that is the data produced by the generator $\{\hat{p}_{m,t}^{\mathcal{G}}\}_{t=t}^{t=t+T}$, the real labelled data $\{p_{m,t}^{label}\}_{t=t}^{t=t+T}$ and the aggregate measurements $\{\tilde{p}_t\}_{t=t}^{t=t+T}$ are fed as inputs to a CNN structure with the main purpose of transforming them into optimized feature maps of $f_{\mathcal{D}}(t)$.

$$f_{\mathcal{D}}(t) \sim Conv_{\mathcal{D}_{EnerGAN++}}\left(\mathcal{I}_{input}\right)$$
$$\text{with}$$
$$\mathcal{I}_{input} = \left[\{p_{m,t}^{label}\}_{t=t}^{t=t+T}, \{\hat{p}_{m,t}^{\mathcal{G}}\}_{t=t}^{t=t+T}, \{\tilde{p}_t\}_{t=t}^{t=t+T}\right]^T \tag{9.9}$$

The features $f_{\mathcal{D}}(t)$ at the time instance $t$ are fed to the GRU structure trained to distriguish the "fake" sequence produced by $\mathcal{G}_{EnerGAN++}$ from the real one (available in the training set). Therefore, we have that

$$\mathcal{D}_{EnerGAN++} \equiv GRU\left(f_{\mathcal{D}}(t)\right) = \begin{cases} 1 & \text{if } \{\hat{p}_{m,t}^{\mathcal{G}}\}_{t=t}^{t=t+T} \approx \{p_{m,t}^{label}\}_{t=t}^{t=t+T} \\ \\ 0 & \text{if } \{\hat{p}_{m,t}^{\mathcal{G}}\}_{t=t}^{t=t+T}\text{not} \approx \{p_{m,t}^{label}\}_{t=t}^{t=t+T} \end{cases} \tag{9.10}$$

Table 9.1 Performance metrics(MAE, SAE, RMSE) for nine appliance of the AMPds and REFIT datasets. In bolds, we have highlighted the method that succeeds the best performance per appliance.

| | Clothes Dryer | | | Heat Pump | | | Oven | | |
|---|---|---|---|---|---|---|---|---|---|
| | MAE | SAE | RMSE | MAE | SAE | RMSE | MAE | SAE | RMSE |
| EnerGAN++ | 17.700 | **0.018** | 192.523 | **80.032** | **0.065** | 227.704 | 8.051 | **0.011** | **108.909** |
| ENERGAN [79] | 25.030 | 0.034 | 260.748 | 80.125 | 0.255 | 256.610 | 10.258 | 0.619 | 145.82 |
| BabiLSTM [56] | **10.009** | 0.062 | **99.211** | 88.215 | 0.354 | 230.155 | 17.645 | 1.420 | 146.459 |
| DAE [99] | 37.355 | 0.170 | 310.081 | 55.608 | 0.261 | 228.690 | 19.255 | 1.306 | 137.930 |
| seq2seq CNN [121] | 15.473 | 0.116 | 149.819 | 107.119 | 1.501 | 260.468 | 67.578 | 7.536 | 322.287 |
| unidirectional LSTM [109] | 90.200 | 1.104 | 219.296 | 154.947 | 2.181 | 285.355 | 57.685 | 6.094 | 242.572 |
| FHMM [128] | 129.568 | 3.422 | 323.225 | 121.689 | 2.641 | 426.662 | 49.383 | 9.047 | 360.852 |
| CO [128] | 120.191 | 2.712 | 468.625 | 249.389 | 2.731 | 459.612 | 267.089 | 50.242 | 433.586 |
| | **Dishwasher** | | | **Kettle** | | | **Microwave** | | |
| | MAE | SAE | RMSE | MAE | SAE | RMSE | MAE | SAE | RMSE |
| EnerGAN++ | **20.320** | **0.079** | 156.633 | **7.811** | **0.232** | 119.102 | **8.302** | 1.417 | **75.840** |
| ENERGAN [79] | 22.166 | 0.154 | 197.087 | 31.304 | 0.788 | 122.357 | 12.046 | 1.444 | 94.008 |
| BabiLSTM [56] | 29.200 | 0.716 | **144.776** | 41.286 | 4.835 | 137.877 | 15.263 | **0.210** | 89.678 |
| DAE [99] | 25.410 | 0.235 | 205.516 | 9.063 | 0.303 | **114.404** | 12.252 | 0.391 | 89.587 |
| seq2seq CNN [121] | 34.987 | 0.886 | 188.099 | 19.896 | 1.968 | 115.482 | 14.830 | 0.408 | 79.568 |
| unidirectional LSTM [109] | 102.171 | 3.432 | 241.053 | 41.120 | 4.332 | 149.885 | 15.922 | 0.168 | 92.983 |
| FHMM [128] | 147.701 | 3.902 | 493.499 | 40.856 | 2.433 | 204.101 | 77.342 | 3.599 | 196.579 |
| CO [128] | 138.805 | 3.629 | 492.172 | 40.659 | 2.567 | 203.586 | 51.826 | 2.042 | 165.419 |
| | **Toaster** | | | **Tumble Dryer** | | | **Washing Machine** | | |
| | MAE | SAE | RMSE | MAE | SAE | RMSE | MAE | SAE | RMSE |
| EnerGAN++ | **2.255** | **0.249** | **32.979** | **16.992** | **0.060** | **108.256** | 7.362 | 1.519 | 105.555 |
| ENERGAN [79] | 2.645 | 0.632 | 38.743 | 19.207 | 0.069 | 133.004 | 8.171 | **0.265** | 101.452 |
| BabiLSTM [56] | 12.810 | 11.897 | 45.886 | 48.749 | 1.097 | 121.309 | 17.696 | 0.811 | **88.099** |
| DAE [99] | 8.294 | 7.737 | 57.960 | 32.864 | 0.149 | 142.340 | 13.400 | 0.275 | 109.782 |
| seq2seq CNN [121] | 15.044 | 13.790 | 50.837 | 42.513 | 0.648 | 158.970 | 27.090 | 2.196 | 133.218 |
| unidirectional LSTM [109] | 26.708 | 14.471 | 50.413 | 87.832 | 1.966 | 183.102 | 31.859 | 2.617 | 121.371 |
| FHMM [128] | 32.457 | 15.443 | 57.786 | 91.554 | 2.456 | 193.771 | 177.015 | 2.747 | 535.198 |
| CO [128] | 35.664 | 12.544 | 67.729 | 91.988 | 3.211 | 201.533 | 210.956 | 2.936 | 458.503 |

## 9.5 The Proposed EnerGAN++ Model Configuration

The min-max game refers to the minimization of generator and the maximization of the discriminator.

### 9.5.1 The adversarial learning between the generator and the discriminator

Eq. (9.10) means that the disctiminator is trained not to be cheated by the generator $\mathcal{G}_{EnerGAN++}$. At the same time the generator is optimized to produce a sequence of the $m$-th appliance in a way to cheat the discriminator, that is $\mathcal{G}_{EnerGAN++}(\hat{z}_m) \approx \{p_{m,t}^{label}\}_{t=t}^{t=t+T}$. This results into a 2-player adversarial learning between the discriminator and the generator

$$\text{EnerGAN++: } \min_{\mathcal{G}} \max_{\mathcal{D}} V(\mathcal{D}_{EnerGAN++}, \mathcal{G}_{EnerGAN++}) =$$

$$E\left[\log \mathcal{D}_{EnerGAN++}(p_m^{label}, \mathcal{G}_{EnerGAN++}(\hat{z}_m))\right] + E\left[\log(1 - \mathcal{D}_{EnerGAN++}(\mathcal{G}_{EnerGAN++}(\hat{z}_m), \bar{p}))\right]$$

$$(9.11)$$

In Eq. (9.11) it is held that $\mathcal{G}_{EnerGAN++}(\hat{z}_m) \equiv \{\hat{p}_{m,t}^{\mathcal{G}}\}_{t=t}^{t=t+T}$.

### 9.5.2 The proposed EnerGAN++ network configuration

Fig. 2 shows the proposed methodology and the basic structure of the proposed model. As we have stated previously two main components are included in the EnerGAN++ model; *the generator* and *the discriminator*.

**EnerGAN++ generator configuration:** The EnerGAN++ generator consists of an autoencoder having (i) *a convolutional encoder* and (ii) *a decoder part*, which is an inverted convolutional version of the encoder. During the training phase the encoder part takes the aggregate signal measurements $\tilde{p}(t)$ and the labelled energy consumption time series of the $m$-th appliance $p_m^{label}(t)$ and maps it to a latent compressed vector of $z_m(t)$. The compressed vector $z_m(t)$ feeds the decoder part of the generator with the purpose of reconstructing an approximate version of $p_m^{label}(t)$ denoted as $\hat{p}_m(t)$. During the testing phase only the aggregate measurements are used as inputs to the encoder part of the generator and therefore, it produces an approximate latent compressed vector of $\hat{z}_m(t)$. The decoder part has been learned, during the training phase, to produce approximate energy consumption time series of the $m$-th appliance. Therefore, during the test phase, the decoder part of the generator produces a data sequence of $\hat{p}_m^{\mathcal{G}}(t)$ close to $p_m^{label}(t)$ from the input latent signal $\hat{z}_m(t)$. Therefore, the loss function of the generator autoencoder will be:

$$L_{\mathcal{G}_{EnerGAN++}}: \ min \left\| p_m^{label}(t) - Decoder(\hat{z}_m(t)) \right\|_2$$

with

$$\hat{z}_m(t) = Encoder(\tilde{p}(t))$$

(9.12)

Eq. (9.12) means that the encoder part of the generator has been trained to produce a latent compressed signal $\hat{z}_m$ with the capability of generating the energy time series $p_m^{label}$ for the $m$-th appliance.

The encoder configuration structure has three convolutional layers: the first layer consists of 256 filters with a kernel size 1×8, the second layer consists of 128 filters of 1×16 kernel size, whereas the third layer consists of 64 filters with size 1×32. The decoder maps the latent value $z_m$ to a higher feature space, which describes accurately the individual appliance waveform. The proposed topology layout has two deconvolution layers: the first layer consists of 64 filters with a kernel size 1×8 and the second layer consists of 128 filters of 1×16 kernel size. The latent vector is not randomly generated, on the contrary it comes from an inverted generator (the encoder) and encloses information from the aggregate signal.

**EnerGAN++ discriminator configuration:** The discriminator $\mathcal{D}$ comprises of a convolutional layer with 60 different kernels consisting of trainable parameters which can convolve the given input and extract the appropriate features. After the automated feature selection, a sequence classification with two stacked recurrent GRU layer follows. The first GRU layer comprises of 40 filters and the second with 30 filters, whereas the network architecture ends up with two dense layers. The first one consists of 30 neurons, whereas the second one is fully connected to the output neuron to predict whether the input pair is the actual data or the generated one. The neurons in output layer use as the activation function the sigmoid function, whereas all the remaining layers use ReLU as the activation function.

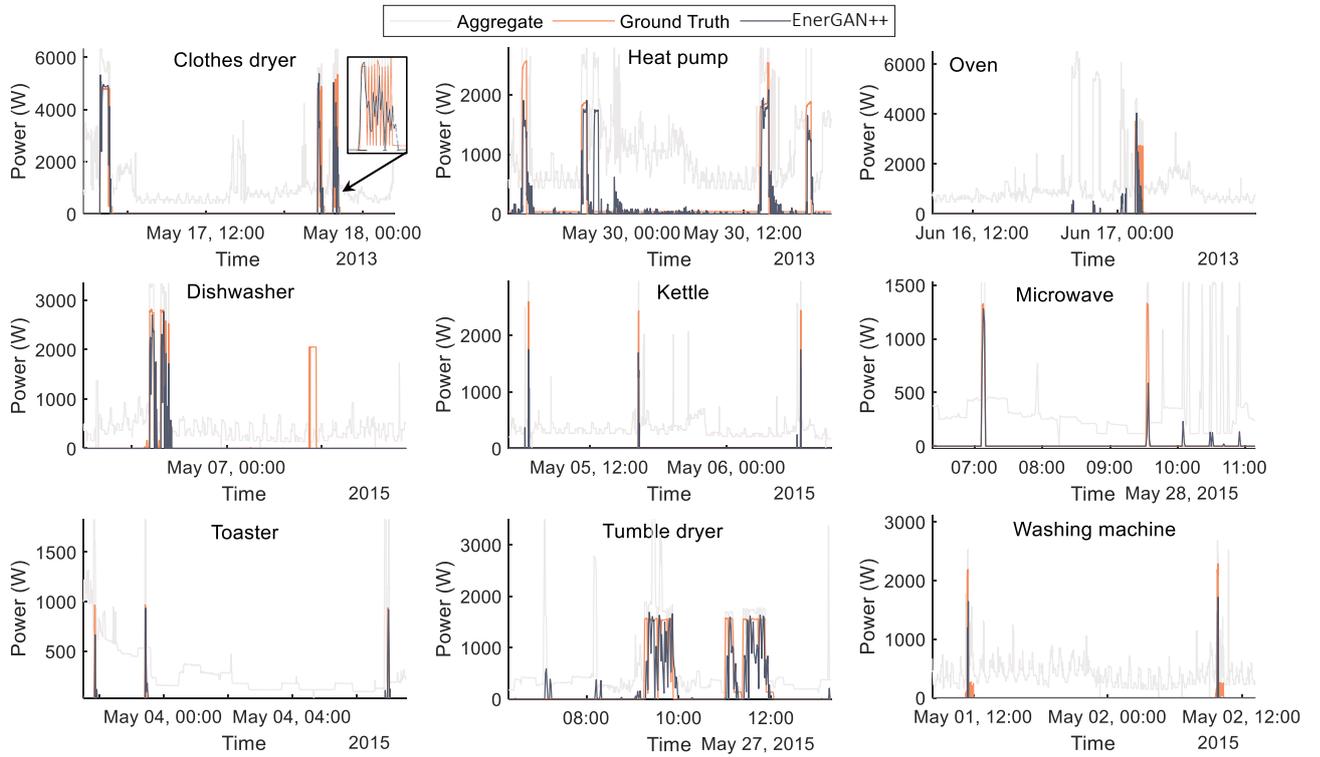

Fig. 9.5 Comparison of the proposed method (purple line) with ground truth (in orange) for selected appliances in AMPds and REFIT dataset. Furthermore, the aggregated data are also illustrated.

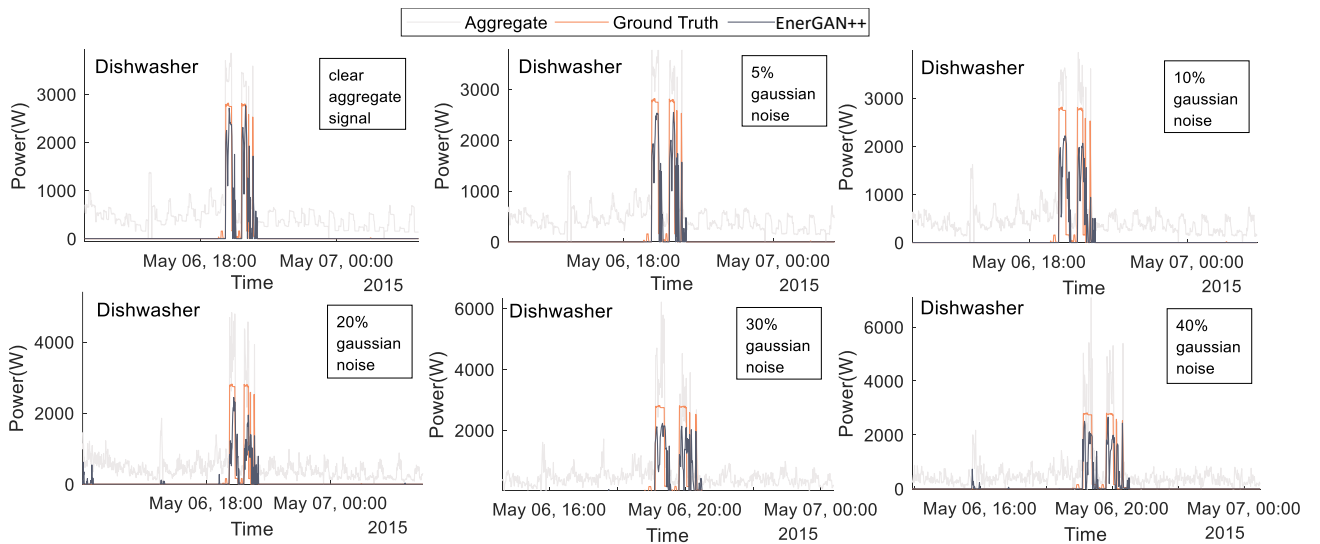

Fig. 9.6 Generated power appliance samples for dishwasher appliance (REFIT). The diagrams show the GAN's model robustness to noise, for all the cases (adding Gaussian noise in 5%, 10%, 20%, 30% and 40%).

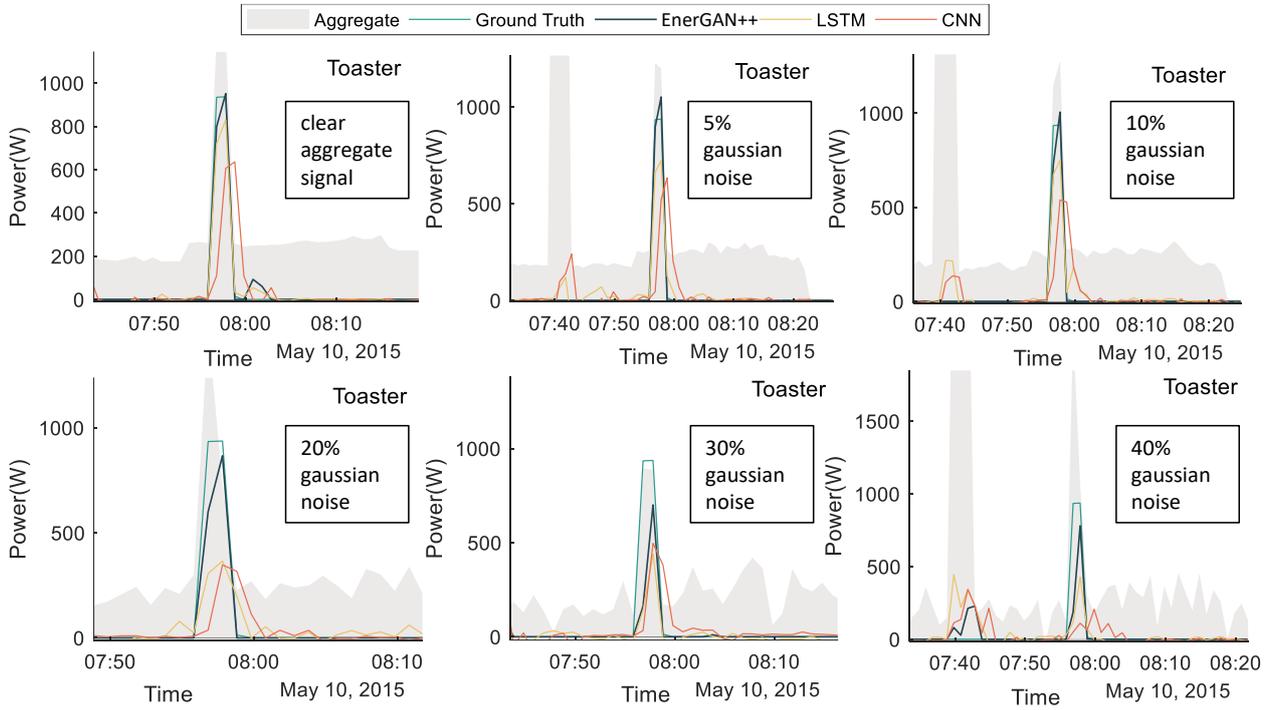

Fig. 9.7 Generated power values for toaster appliance, The diagrams indicate that the proposed model is robust to noise, whereas the baselines are weak for the cases with noise.

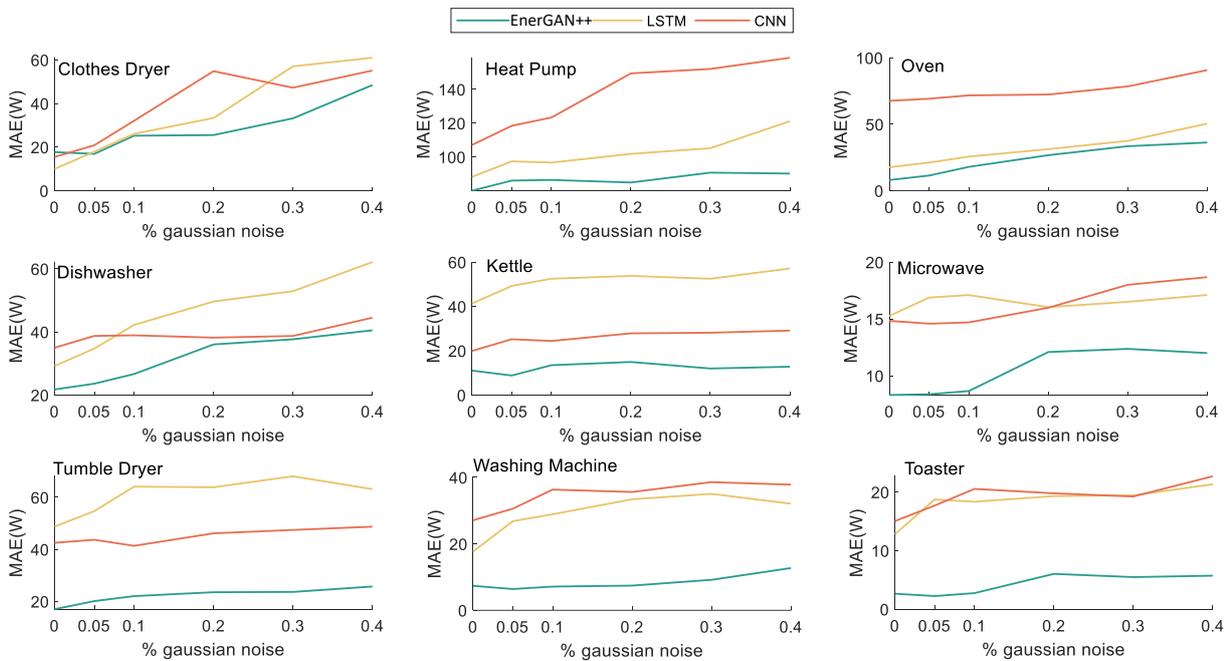

Fig. 9.8 Quantitative results on AMPds and REFIT datasets. The small MAE error values provide better performance.

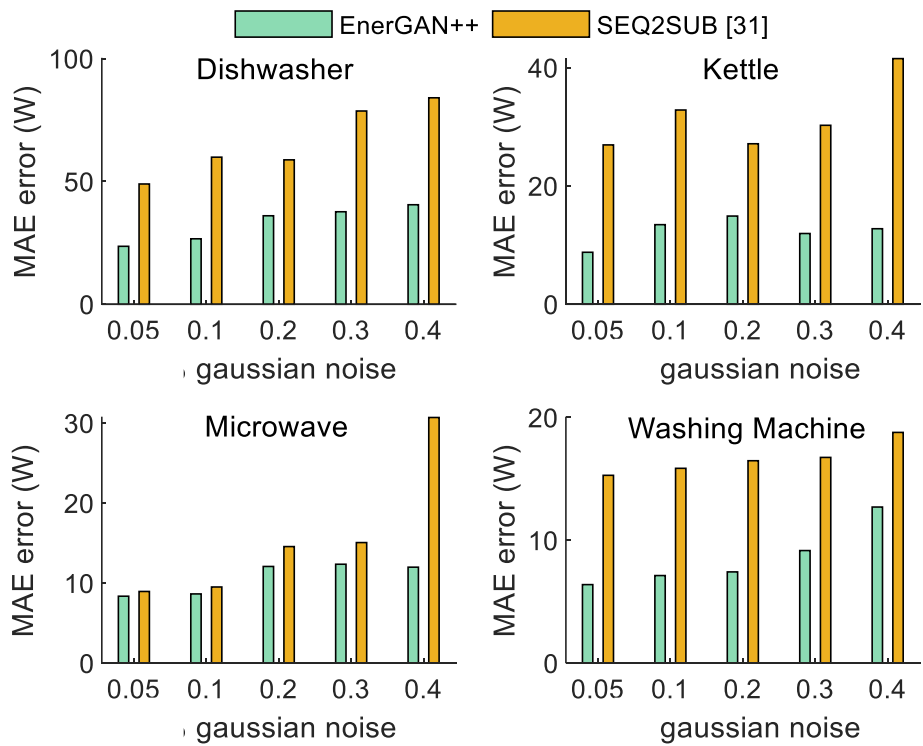

Fig. 9.9 Comparison between EnerGAN++ model results provided after applying Gaussian noise to the aggregate signal and the respective results of the disaggregation using the sequence to subsequence conditional GAN model of [2]. The small MAE error values provide better performance.

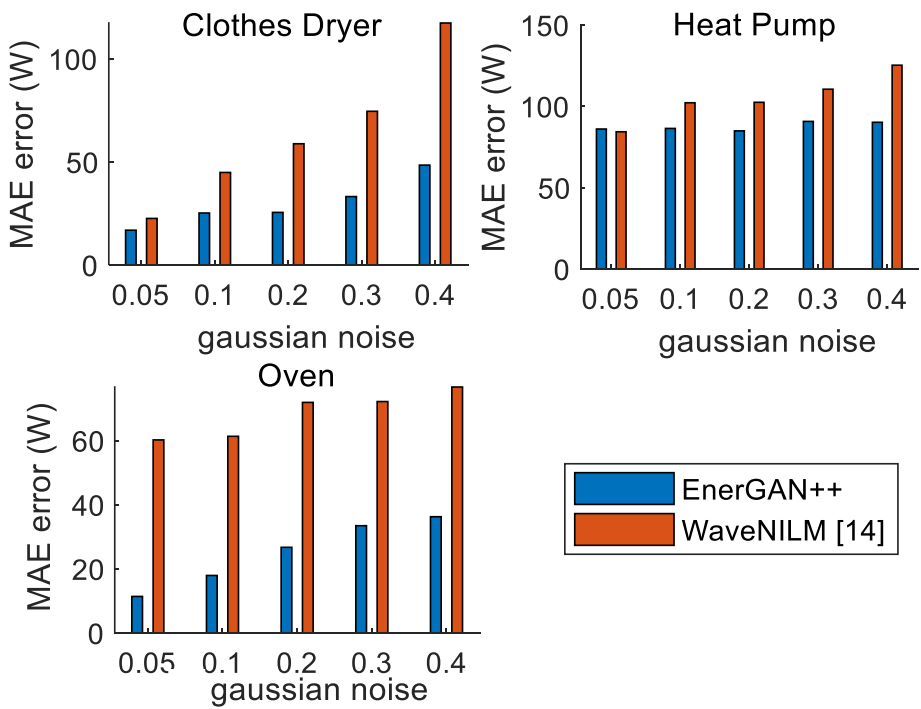

Fig. 9.10 Comparison between EnerGAN++ model results provided after applying Gaussian noise to the aggregate signal and the respective results of the disaggregation using the WaveNILM model of [3]. The small MAE error values provide better performance.

## 9.6   Experiments

### 9.6.1   Description of the Dataset

The evaluation of our generative adversarial model for energy disaggregation has been conducted on nine appliances derived from AMPds [119] and REFIT [131] datasets. These open-access energy consumption datasets provide the aggregate power measurements of the whole house and sub-metered readings (smart plugs) from individual appliances at different time resolutions; 60s for AMPds, 8s for REFIT. In our study, REFIT data are down sampled to 60s resolution. The AMPds consists of a single house in Canada, whereas the REFIT consists of 20 houses located in UK. Both the AMPds and REFIT datasets collected over a period of two years. The appliances are: clothes dryer, heat pump and oven appliances from AMPds and dishwasher, kettle, microwave, toaster, tumble dryer and washing machine appliances from REFIT dataset.

The section 9.6 presents: (i) the evaluation performance metrics for our proposed methods in contrast to the other state of the art methods (see section 9.6.3), (ii) the robustness of our method to noise, which is our comparative advantage in contrast to other conventional NILM methods (see section 9.6.4).

The chapter points out the EnerGAN++ method's comparative advantage for noisy aggregate data. Furthermore, REFIT measurements contain noise in the label ground truth data due to numerous unknown appliances and measurement errors. This introduces an additional challenges for our proposed method.

Our EnerGAN++ algorithm is implemented in Python 3.6 using the Keras API integrated into TensorFlow 2. The computer used for all of the training and testing was an Intel Core i7-8750H CPU at 2,20 GHz with 8 GB of random-access memory and an NVIDIA GeForce GTX 1050 with 4096 MB of DDR5 memory.

### 9.6.2   Quantitative Evaluation Metrics

In this study, we compare the proposed EnerGAN++ method for NILM, against other state of the art methods. In particular, our method is evaluated against the previous version of our proposed GAN (EnerGAN) [79], as well as the BabiLSTM network, as presented in [56]. Furthermore, the method is compared against the traditional deep learning models, applied for NILM, such as sequence to sequence CNNs (seq2seq CNN), [121], [56], unidirectional LSTMs [109], [57] and denoising autoencoders [99]. These models have also been implemented in Python 3.6 using Keras. Finally, we compare the proposed method against additional benchmark approaches such as FHMM [128] and CO [128]. These approaches are implemented in NILMTK library [128].

Table I is a summary that compares the aforementioned techniques with the proposed EnerGAN++ model. The metrics used are:

(i) the Mean Absolute Error (MAE) that measures the average magnitude of the errors in the set of predictions and is a commonly used metric for NILM [58]. MAE metric is used when we are interested in the error in power at every time point, and is less affected by outliers, i.e. isolated predictions that are particularly inaccurate:

$$MAE = \frac{\Sigma_{t=1}^{N} |\hat{\mathbf{p}}_{\mathbf{j}}(t) - \mathbf{p}_{\mathbf{j}}(t)|}{N} \quad (9.13)$$

(ii) the Signal Aggregate Error (SAE) is a common metric used to estimate the total energy consumed by each appliance over a period of time [58]. This measure is useful because a method could be accurate enough for reports of daily power usage even if its per-timestep prediction is less accurate,

$$SAE = \frac{\left| \Sigma_{t=1}^{N} \hat{\mathbf{p}}_{\mathbf{j}}(t) - \Sigma_{t=1}^{N} \mathbf{p}_{\mathbf{j}}(t) \right|}{\Sigma_{t=1}^{N} \mathbf{p}_{\mathbf{j}}(t)} \quad (9.14)$$

(iii) the Root Mean Squared Error (RMSE), that is more sensitive to large errors, occurring either due to time delays or even fault detection (outliers),

$$RMSE = \sqrt{\frac{\Sigma_{t=1}^{N} (\hat{\mathbf{p}}_{\mathbf{j}}(t) - \mathbf{p}_{\mathbf{j}}(t))^2}{N}} \quad (9.15)$$

From the above, we consider that the smaller the MAE, SAE and RMSE errors are, the better the performance of the examined method is.

Table 9.2 Rate of change in MAE error with an increase on Gaussian noise. The comparison is between the proposed EnerGAN++ method and the BabiLSTM as well as sequence to sequence CNN methods.

| Appliance | Clothes dryer | | | Heat pump | | | Oven | | |
|---|---|---|---|---|---|---|---|---|---|
| Method | EnerGAN++ | BabiLSTM | seq2seqCNN | EnerGAN++ | BabiLSTM | seq2seqCNN | Proposed | BabiLSTM | seq2seqCNN |
| $(0-5)\%$ | -0.16 | 1.58 | 1.07 | 1.20 | 1.84 | 2.26 | 0.67 | 0.70 | 0.32 |
| $(5-10)\%$ | 1.67 | 1.63 | 2.25 | 0.07 | -0.15 | 0.99 | 1.31 | 0.90 | 0.51 |
| $(10-20)\%$ | 0.03 | 0.73 | 2.27 | -0.15 | 0.52 | 2.61 | 0.88 | 0.56 | 0.06 |
| $(20-30)\%$ | 0.77 | 2.37 | -0.76 | 0.58 | 0.33 | 0.26 | 0.67 | 0.62 | 0.61 |
| $(30-40)\%$ | 1.53 | 0.39 | 0.79 | -0.05 | 1.60 | 0.66 | 0.28 | 1.30 | 1.22 |
| $\overline{\Delta r}$ | **0.83** | 1.34 | 1.43 | **0.41** | 0.89 | 1.36 | 0.76 | 0.82 | **0.55** |

| Appliance | Dishwasher | | | Kettle | | | Microwave | | |
|---|---|---|---|---|---|---|---|---|---|
| Method | EnerGAN++ | BabiLSTM | seq2seqCNN | EnerGAN++ | BabiLSTM | seq2seqCNN | Proposed | BabiLSTM | seq2seqCNN |
| $(0-5)\%$ | 0.37 | 1.10 | 0.75 | 0.19 | 1.59 | 1.05 | 0.01 | 0.32 | -0.05 |
| $(5-10)\%$ | 0.61 | 1.50 | 0.04 | 0.93 | 0.65 | -0.15 | 0.06 | 0.04 | 0.02 |
| $(10-20)\%$ | 0.94 | 0.74 | -0.07 | 0.15 | 0.13 | 0.33 | 0.34 | -0.11 | 0.13 |
| $(20-30)\%$ | 0.16 | 0.33 | 0.05 | -0.29 | -0.13 | 0.03 | 0.03 | 0.05 | 0.20 |
| $(30-40)\%$ | 0.29 | 0.92 | 0.58 | 0.08 | 0.46 | 0.10 | -0.04 | 0.06 | 0.07 |
| $\overline{\Delta r}$ | 0.47 | 0.92 | **0.30** | **0.33** | 0.59 | 0.34 | **0.09** | 0.12 | 0.09 |

| Appliance | Tumble dryer | | | Washing machine | | | Toaster | | |
|---|---|---|---|---|---|---|---|---|---|
| Method | EnerGAN++ | BabiLSTM | seq2seqCNN | EnerGAN++ | BabiLSTM | seq2seqCNN | Proposed | BabiLSTM | seq2seqCNN |
| $(0-5)\%$ | 0.62 | 1.19 | 0.23 | -0.19 | 1.82 | 0.70 | -0.08 | 1.18 | 0.53 |
| $(5-10)\%$ | 0.38 | 1.88 | -0.47 | 0.15 | 0.41 | 1.15 | 0.10 | -0.08 | 0.57 |
| $(10-20)\%$ | 0.15 | -0.02 | 0.48 | 0.03 | 0.45 | -0.07 | 0.33 | 0.10 | -0.08 |
| $(20-30)\%$ | 0.01 | 0.42 | 0.13 | 0.17 | 0.16 | 0.30 | -0.06 | 0.01 | -0.05 |
| $(30-40)\%$ | 0.21 | -0.49 | 0.13 | 0.35 | -0.30 | -0.08 | 0.02 | 0.19 | 0.35 |
| $\overline{\Delta r}$ | **0.27** | 0.80 | 0.29 | **0.18** | 0.63 | 0.46 | **0.12** | 0.31 | 0.31 |

### 9.6.3 Experimental Results

In Table 1, our proposed method has the lower values for MAE metric. This indicates that the proposed EnerGAN++ has the best performance compared to the other methods, in terms of MAE.

The exception is the clothes dryer appliance, where MAE as well as RMSE values seem to be higher, while SAE error succeeds the minimum value. This is probably occurs due to the "jagged edges" appeared in clothes dryer appliance pattern, that successfully cashed by the bidirectional-LSTM scheme, but only as outline in the proposed scheme (also see Fig. 9.5). SAE and RMSE metric values show also good performance for the majority of appliances. In terms of RMSE error, however, the "failures" observed refereed to time delays occurred between the generated and ground truth data. Furthermore, from previous studies in NILM, has been observed that architectures such as LSTM and CNN, underestimate power values during the training process. As a result, such schemes, produce patterns with lower power values, leading to small RMSE values, but, on the contrary, to increased MAE values. Furthermore, it should be pointed out that the proposed network has been significantly improved compared to our former EnerGAN network [79]. The improvement is due to the different discriminator configuration, i.e. the recurrent convolutional network for sequence classification that is trained to discriminate samples from data, as well as the different loss functions selected for our proposed GAN network.

Fig. 9.5 shows the aggregate signal (grey line) and the generated power timeseries from our GAN network (purple line) at a given time period. Also, with orange line the ground truth data are depicted. As shown, the operation of each appliance is detected at an adequate level. In Fig. 9.5, the generated timeseries of power data are identical with the actual operation (ground truth) of clothes dryer, oven, kettle, microwave, toaster, tumble dryer and washing machine appliances. However, during the snapshot in time as illustrated in Fig. 9.5, for the heat pump appliance, a false positive is appeared, since the appliance is detected in operation five times, whereas 4 times is actually ON. On the contrary, a false negative is detected, at first, for dishwasher appliance, but actually the orange undetected signal for this case indicates noise and not actual presence of the dishwasher appliance.

### 9.6.4 Robustness to noise

Hereby we compare the results of our proposed method after having applied additive white Gaussian noise in the aggregate signal in varying percentages (5%, 10%, 20%, 30% and 40%). The $n$% percentage values means that the power measurements of the aggregated signal (in W) have a deviation of $n$% with 68%. Fig. 9.6 shows the generated values and their respective ground truth, assuming clear aggregated signal (as illustrated in the up left diagram) and aggregated signal, corrupted with 5%, 10%, 20%, 30% and 40% Gaussian noise, respectively. As noticed, the aggregated power values are differentiated from the real ones, according to the percentage of the Gaussian noise, and gradually "new" peak values in the aggregate timeseries are appeared, that could lead to an erroneously detection of an appliance in operation or to degrade the quality of the generated appliance power signal. In Fig. 9.6, the aggregated signal distortion, is observed, for the different percentages of the Gaussian noise. For example, on the 6th of May, at 6:00 AM, the aggregated signal has its maximum, with peak values reaching the 4000 W, while, after adding the Gaussian noise (40%), the same maximum is now higher ( $6000W$). Despite of the additive Gaussian noise, it appears that the GAN's performance isn't affected.

Dishwasher appliance has a duration in operation of about a half hour or even 1 hour. The Gaussian noise, distorts the aggregated signal, however, this distortion is not enough to fool the EnerGAN++ and force it to fault detection. In the next step, the study examines the behaviour of the EnerGAN++ model in detecting appliances in operation for short time, since the peak values from the Gaussian noise act as a catalyst for techniques that detect the appliance pattern into the aggregated signal. Toaster, an appliance that remains ON for a short period of time (some minutes), has been selected. Fig. 9.7 compares the generated appliances power values (dark green line) using as input an aggregate signal corrupted with Gaussian noise (gray filled color), with the respective estimations after applying the traditional LSTM (yellow line) and CNN (orange line) methods. In contrast to EnerGAN++ method that performs learning in an adversarial framework (between generator and discriminator), CNN and LSTM methods try to catch the appliance signal from the aggregated signal using non-linear regression relationships. Thus, a possible aggregated signal distortion, leads to erroneous appliance signal detection and the opposite one. This is obvious, in Fig. 9.7, where the CNN and LSTM architectures underestimate the toaster appliance power values during operation at 8:00 AM, as the Gaussian noise increases from 5% to 40%. In addition, fault detection seems to be appeared for LSTM and CNN models. The right down diagram (40%) is indicative of the CNN method's "confusion". In particular, it appears that the CNN model detects the toaster appliance in operation, two times in the time interval between 7:45 and 8:00 AM, and in addition, the peak power values for the appliance in operation reach almost the 400 W, whereas the actual power values should be of about 900 W.

Fig. 9.8 shows the MAE metric results for noise in the aggregate signal in all cases (5%, 10%, 20%, 30%, 40%) between the proposed approach and the LSTM and CNN methods, per appliance. The quantitative results on AMPds and REFIT datasets, shown that the EnerGAN++ approach is robust to noise in the aggregate signal. Here, it is nevertheless necessary to remind that the smaller MAE error values provide a better performance. It is indicative that in most appliances, the rate of change of the MAE is lower than the increase in Gaussian error (for further details see Table II).

Fig. 9.9 compares the results of EnerGAN++ model with the corresponding results provided by the sequence to sub-sequence conditional GAN model [2], given noisy aggregate signal values, for the REFIT dataset. EnerGAN++ is a robust to noise model and achieves good performance, regardless of noise, in contrast to the SEQ2SUB model. SEQ2SUB model achieves good results in normal cases as shown in [2], however, given noise aggregate signal input, its performance is degraded (Fig. 9.9). Fig. 9.10 compares the result of EnerGAN model and WaveNILM [3] model given noisy aggregated inputs, for the AMPds dataset. The results using WaveNILM method are significantly degraded with the noisy inputs.

Table II shows the MAE error rate of change $\Delta r$ for the respective increase of Gaussian noise $((0-5)\%, (5-10)\%, (10-20)\%, (20-30)\%, (30-40)\%)$, as well as the average rate of change $\overline{\Delta r}$ for all the above mentioned cases.

$$\Delta r = \frac{MAE_{n_j\%} - MAE_{n_i\%}}{n_j - n_i}, \ \ n_j > n_i \tag{9.16}$$

$$\overline{\Delta r} = \frac{\sum_1^N |\Delta r|}{N} \tag{9.17}$$

As observed, the change of MAE by an additional one percentage point, referred as $\overline{\Delta r}$, is lower in our proposed case, rather than in the two other methods (BaBiLSTM and seq2seqCNN).

## 9.7    Conclusion

In this chapter, we attempt to incorporate the autoencoder architecture under a generative adversarial network, to achieve the inverse mapping and extract meaningful appliance power consumption signal from the generator. Furthermore, we propose a novel discriminator for sequence classification, that successfully distinguishes the generated appliance sequences from the real ones. A hybrid CNN-GRU model ensures that the discriminator creates proper features and exploits the temporal information of the timeseries. In the discriminator, the appliance power signal is paired with the aid of the aggregated signal, to assist the discriminator to properly separate the generated from the real values. Experimental results indicate the proposed method's superiority compared to the current state of the art.

# Chapter 10

# Conclusions

## 10.1 Summary and Contribution of the Dissertation

Deep neural networks have successfully been applied to address time series modelling problems. They have proved to be an effective solution given their capacity to automatically learn the temporal dependencies present in time series. However, selecting the most convenient type of deep neural network and its parametrisation is a complex task that requires considerable expertise. Numerous deep learning architectures have been developed to accommodate the diversity of time-series datasets across different domains. In this dissertation, we proposed architectures in both one-step-ahead and multi-horizon time-series prediction—describing how temporal information is incorporated into the model. Furthermore, we have proposed hybrid deep learning models, which combine different neural network components to improve pure methods in either category.

### 10.1.1 Contribution in the field of ionospheric TEC modeling

To establish high-precision forecast products and use them for positioning needs, it is necessary to introduce reliable regional observation data to characterize regional anomalies that affect the ionosphere. In this dissertation, we review a variety of deep learning models for time series ionospheric TEC prediction that have been developed to explicitly capture temporal relationships.

Recurrent Neural Networks and their variants have become competitive methods for time series TEC modeling. In chapter 3, a supervised unidirectional LSTM regression model that predicts vertical TEC values has been proposed. The LSTM structure models the temporal dependencies of the phenomenon and additionally the entrance to the model of external parameters related to solar and geomagnetic activity, enhances the models accuracy. Furthermore, non-linearity and sequence-to-sequence modelling is an additional advantage of the proposed LSTM method.

A Recurrent Neural Network is a type of neural network well-suited to time series data. Moving towards RNN models that have as basic structure recurrent layers, that are able to process a time series step-by-step, alternative architectures are proposed here, that adopt convolutional la yers as basic structure. The proposed network has its origins in the traditional CNN network that is popular for imaging applications. However, here is attempted to explore CNN correspondence in detecting temporal recurrent patterns in timeseries structures, through the implementation of various adjustments and proposing specialized convolutional architectures. A temporal convolutional neural network is proposed that consists of causal dilated 1D convolutions with autoregressive character to predict ionosphere variations above a ground station and extract individualized VTEC values per observed satellite. For ionosphere TEC modelling, the proposed network has statistically higher accuracy than the other compared models, due to the successful combination between dilated convolutions and very deep networks (augmented with residual layers). Thus, the temporal CNN model looks very far into the past to make a prediction, allowing temporal dependencies.

Then, a model for accurate TEC estimates from GNSS data, surpassing problems related to ionosphere irregular behavior at regional level, is proposed. The model creates robust regional TEC models, which can eventually be used to estimate ionosphere variations at different geographic locations (*spatial*), for different time periods (*temporal*) and under various solar and geomagnetic conditions. The proposed model allows different solar and geomagnetic parameters as features. As regards the spatial variability, depending on the station latitude, different weights to the model should be assigned. These weights are learnable during the training phase and the features contribute in a different way to the predictions of the model, depending on the station. In this step, the *convolutional neural networks (CNN)* properly adjust the weights to better capture the particularities of each station [36]. The proposed deep learning architecture is a combination between convolutional and recurrent layer structures, to enhance temporal variability. Our model leverages the strength of rapid progress in CNN, among with the need to apply the convolutional models for temporal sequences of data. The CNN layer is used for the optimal feature extraction and then, the stacked recurrent layers follow, to control and manage the temporal information flow.

### 10.1.2   Contribution in the field of energy disaggregation

In chapter 7, a contextually adaptive and Bayesian optimized bidirectional LSTM model for modeling appliances' consumption patterns in a NILM operational framework is proposed. The CoBiLSTM model is a generalized bidirectional LSTM network, adaptable to different contextual factors with an internal process of adaptation based on model performance error. The proposed approach based on progressive model updates can effectively adapt to a wider and more diverse range of instances and conditions, thus making a successful step towards scaled up NILM applicability. A bespoke, appliance-specific and specific to particular contextual configuration model would have better performance than our scalable context aware model, but it would only be applicable in a very con-strained set of settings. The proposed model provides an updating solution that is sufficiently general, while at the same time managing to capture individual particularities in appliance operation.

Then, a novel deep learning based on CNN structure is introduced for energy disaggregation. The proposed CNN deep-learning multi-input/multi-output regression model leverages the recurrent property to effectively model the temporal interdependencies of the power signals. Moreover, the incorporation of multiple channels, each for a different signal (active, reactive, apparent power and current), offers additional streams of information resulting in a more accurate model. Experimental results suggest higher performance and faster convergence times compared to state of the art approaches.

Then, the NILM problem is solved by incorporating an autoencoder architecture to a generative adversarial network, to achieve the inverse mapping and extract meaningful appliance power consumption signal from the generator. Furthermore, a novel discriminator for sequence classification is proposed. The discriminator successfully distinguishes the generated appliance sequences from the real ones. A hybrid CNN-GRU model ensures that the discriminator creates proper features and exploits the temporal information of the timeseries. In the discriminator, the appliance power signal is paired with the aid of the aggregated signal, to assist the discriminator to properly separate the generated from the real values. Experimental results indicate the proposed method's superiority compared to the current state of the art.

## 10.2 Future Prospects

There is, of course, room for additional work in each of the topics covered in this dissertation. Hence, in this final section, some ideas for future research are discussed. An issue that falls under the broader research area of this dissertation is the incorporation of the current trends in the literature such as ensemble models to enhance accuracy or transfer learning and to reduce the burden of high training times. As for time series prediction, providing explainable result is beneficial for measuring its reliability. Therefore, enhancing interpretability becomes a future topic attracting intensive attention. Acceleration for the learning process is another area in which deep learning needs to be improved over the next few years. In order to determine the weights of multiple layers in the network, deep learning-based approaches usually requires hours or even days for training, even using the latest GPU processors, which is unacceptable for time series prediction in some real-world application cases. For improving the computational efficiency, some studies have been conducted to propose algorithms and strategies for accelerating deep learning [20].

### 10.2.1 Ionospheric TEC modeling

The transmitted signals from Global Navigation Satellite Systems (GNSS) are directly affected by the ionospheric variations, causing delays [50]. These delays depend on the signal frequency and the electron density along the transmission path. Hence, ionospheric variability introduces an additional error source in GNSS positioning [51]. The use of multiple navigation signals of distinct center frequency transmitted from the same GNSS satellite allows direct estimation of these ionospheric delays. Exploiting the fact that different signal frequencies are affected differently by the ionosphere, an appropriate processing strategy of multiple-frequency GNSS signals, eliminates the ionospheric error [51]. Contrary to multi-frequency GNSS receivers, real-time (RT) single-frequency (SF) positioning with a low-cost receiver has received increasing attention in recent years due to its large amount of possible applications. However, in this case, one major challenge is the effective mitigation of these ionospheric delays [52]. RT-SF-SPP (Standard Point Positioning)/PPP (Precise Point Positioning) techniques use ionospheric vertical TEC (VTEC) products released by the International GNSS Service (IGS) real-time service [53], to eliminate the ionospheric error and apply corrections to the model as external parameters. However, these ionospheric VTEC products have global coverage. An interesting future prospect would be to extend the applicability of the proposed non-linear regional models for ionospheric variability modelling and to apply them as external ionosphere corrections in near-real time PPP-RTK processing. The outcomes of these models could be used as part of the integer ambiguity resolution (IAR)–enabled PPP owing to the use of predicted ionospheric delays in addition to other corrections.

As future work, it is important to assess the accuracy of the TECs computed by the model in the position domain, with precise coordinates of the IGS stations. Also, it is important to investigate model reliability, given as inputs GNSS observational data from a regional network of stations close to each other, to examine the extent to which such models capture the regional anomalies.

In addition, more complicated deep learning architectures such as semi-supervised learning [94], [95] and/or Generative Adversarial Networks (GAN) [90] can be examined to see if they can improve the results while retaining small computational cost.

### 10.2.2 Energy disaggregation

Efficient and careful utilisation of energy resources is necessary in order to conserve them. Energy disaggregation is an essential element in the conservation of energy since it elaborates the energy usage tendencies. The trends observed in the energy-usage patterns from a household can be used for security purposes. A detected anomaly behaviour might implies a sign of appliance failure or illegal use of supplied electricity. The appliance usage patterns can also be used to calculate and control the amount of carbon emissions. Also, investigating model's scalability and applicability to different contextual factors that constitute an obstacle to the solution of NILM problem, is another interesting prospect for future research.